%% file: main.tex
\PassOptionsToPackage{table,dvipsnames}{xcolor}
\documentclass{pkumelon}

\usepackage[utf8]{inputenc}
\input{math_commands.tex}

\usepackage{amssymb}
\usepackage{amsthm}
\usepackage{float}
\usepackage{url}
\usepackage{xspace}
\usepackage{arydshln}
\usepackage{wrapfig}
\usepackage{needspace}

\titlespacing*{\paragraph}{0pt}{1.5ex plus .3ex minus .2ex}{.8em}


\renewcommand{\titlefont}{\fontsize{17}{20}\selectfont}
\DeclareFontShape{T1}{bytesans}{b}{n}{<-> s * [1] pkumelon/bytesans}{}
\DeclareFontShape{T1}{bytesans}{bx}{n}{<-> s * [1] pkumelon/bytesans}{}
\renewcommand\affiliationformat[2][]{{\affiliationfont {\sffamily $^{#1}$#2}}}
\renewcommand\authorformat[2][]{%
  \authorfont {\sffamily
    \mbox{\ifstrempty{#1}{#2}{#2$^{#1}$}}%
  }%
}
\hypersetup{
  pdftitle={SDG: State-Dependent Gating at Chunk Boundaries for Linear Attention},
  pdfauthor={Renjia Deng, Yuxi Liu, Yutong He, Xue Wang, Kun Yuan}
}

\AtBeginDocument{%
  \expandafter\let\csname endtabular*\endcsname\endtabular}

\DeclareFontShape{OMX}{cmex}{m}{n}{<->cmex10}{}

\newtheorem{proposition}{Proposition}

\newtheorem{lemma}{Lemma}
\newtheorem{corollary}{Corollary}
\theoremstyle{definition}
\newtheorem{remark}{Remark}

\makeatletter
\def\sdg@boldseries{b}
\newcommand{\sdg}{%
  \ifx\f@series\sdg@boldseries SDG\else\textsc{SDG}\fi\xspace}
\makeatother
\newcommand{\src}{\mathrm{src}}

\definecolor{SDGLightPurple}{RGB}{243,239,249}
\definecolor{SDGQuestionBackground}{RGB}{248,246,252}
\definecolor{SDGPurple}{RGB}{126,96,164}
\newsavebox{\sdgquestionbox}
\newlength{\sdgsavedtextfloatsep}
\newcommand{\modelsep}{\addlinespace[2pt]}
\newlength{\modelcolwidth}
\newlength{\modelnamewidth}
\newlength{\metriccolwidth}
\newlength{\qacolwidth}
\newcolumntype{C}[1]{>{\centering\arraybackslash}p{#1}}
\newcounter{algorithm}
\newcounter{sdgalgorithmline}
\newcommand{\sdgalgline}[2][0pt]{%
  \par\noindent\stepcounter{sdgalgorithmline}%
  \makebox[1.7em][r]{\normalfont\footnotesize\arabic{sdgalgorithmline}:}%
  \hspace{0.5em}\hspace{#1}%
  \begin{minipage}[t]{\dimexpr\linewidth-2.2em-#1\relax}%
    \raggedright\strut#2\strut
  \end{minipage}\par}
\title{SDG: State-Dependent Gating at Chunk Boundaries for Linear Attention}

\author[1,2*]{Renjia Deng}
\author[1]{Yuxi Liu}
\author[1]{Yutong He}
\author[2]{Xue Wang}
\author[1\P]{Kun Yuan}
\affiliation[1]{Peking University}
\affiliation[2]{Meituan Inc.}

\abstract{
Recent gated linear attention models use state-independent gates, computed from token representations without directly reading the recurrent memory.
This design supports efficient chunkwise computation: sequences are divided into chunks of consecutive tokens, with parallel computation within each chunk.
However, the same token representation produces the same gate, even when the layer stores different information.
We introduce \emph{\underline{S}tate-\underline{D}ependent \underline{G}ating} (\sdg), which computes element-wise gates from the change in memory over a chunk and applies them at the chunk boundary.
This provides state feedback for forgetting while preserving parallel computation within each chunk.
We develop fused kernels for Gated DeltaNet, Kimi Delta Attention, Gated DeltaNet-2, and Mamba-3.
Our theoretical analysis quantifies the potential reconstruction benefits of state-dependent, element-wise gating under a signal--noise model.
We also show that boundary decay is an element-wise ridge proximal step when the gate is held fixed.
Empirical evaluation of 1.3B linear attention and hybrid models trained on 100B tokens shows improvements across language modeling, commonsense reasoning, synthetic and real-world retrieval, and long-context understanding. These results demonstrate that SDG is an effective approach to enhancing existing linear attention models.
}

\contribution[*]{Work done during the internship at Meituan Inc.}
\contribution[\P]{Corresponding author}

\checkdata[Code]{\url{https://github.com/lovekdl/SDG}}
\checkdata[Emails]{\href{mailto:2501210078@stu.pku.edu.cn}{\nolinkurl{2501210078@stu.pku.edu.cn}}, \href{mailto:yuxiliu666@stu.pku.edu.cn}{\nolinkurl{yuxiliu666@stu.pku.edu.cn}}, \href{mailto:yutonghe@pku.edu.cn}{\nolinkurl{yutonghe@pku.edu.cn}}, \href{mailto:kunyuan@pku.edu.cn}{\nolinkurl{kunyuan@pku.edu.cn}}}

\begin{document}

\maketitle
\enlargethispage{\baselineskip}
\suppressfloats[t]

\section{Introduction}
\label{sec:introduction}
\raggedbottom
\setlength{\sdgsavedtextfloatsep}{\textfloatsep}
\setlength{\textfloatsep}{10pt}

Linear attention stores context in a fixed-size recurrent state, enabling linear-time sequence modeling and constant-memory decoding~\citep{katharopoulos2020transformers,schlag2021linear,sun2023retnet}. Its limited capacity makes retention important: as new associations are written, they can interfere with those already stored~\citep{kohonen1972correlation,arora2023zoology}. Recent gated recurrent models control retention using coefficients predicted from token representations~\citep{gu2023mamba,yang2024gla,yang2024gdn,kimi2025linear,hatamizadeh2026gdn2}. These gates are data-dependent but state-independent: their coefficients do not directly read the recurrent memory of the same layer. State independence here refers to the gate predictor within a layer. Token representations may already contain context from earlier layers, but the predictor does not explicitly inspect the memory that its coefficients will modify.

\begin{figure}[t]
  \centering
  \setlength{\abovecaptionskip}{3pt}
  \includegraphics[width=\linewidth]{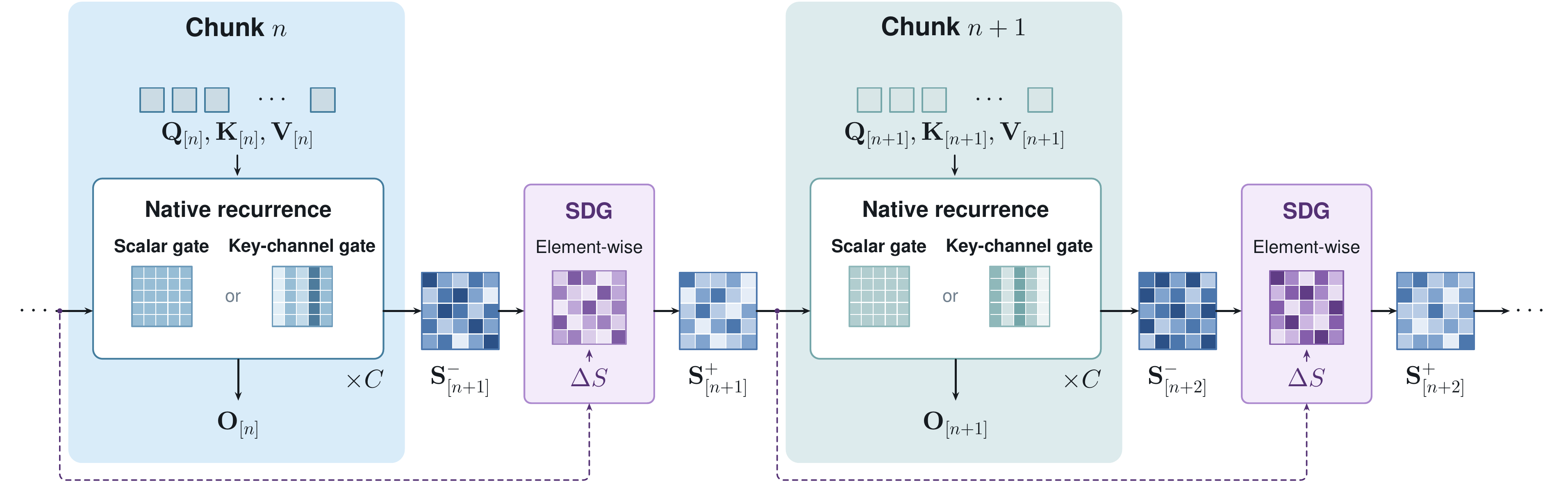}
  \caption{\textbf{State-Dependent Gating (SDG).} The base layer processes a chunk of $C$ tokens, taking the incoming state $\mathbf S_{[n]}^+$ to the updated state $\mathbf S_{[n+1]}^-$ and producing outputs $\mathbf O_{[n]}$. \sdg uses the difference between states to compute an element-wise gate and passes the new state $\mathbf S_{[n+1]}^+$ to the next chunk.}
  \label{fig:sdg-overview}
\end{figure}

This separation supports efficient chunkwise computation. A sequence is divided into chunks of consecutive tokens, with parallel matrix operations within each chunk and recurrent state propagation between chunks. The gate coefficients can be computed before the state updates, and their structure enables further reuse of computation. Gated DeltaNet (GDN) uses one decay coefficient per head, whereas Kimi Delta Attention (KDA) uses one per key channel. Both share decay coefficients across value channels, allowing their chunkwise algorithms to reuse transition factors~\citep{yang2024gdn,kimi2025linear}. A scalar gate cannot distinguish value rows or key channels, while a key-channel gate can distinguish columns but still treats all value rows alike.

The same design limits how gates use memory. For a fixed token representation, the gate is unchanged even if the layer stores different information. Although the delta update reads the state to correct stored associations, this read does not enter the gate predictor. Coefficient sharing also forces different state elements to use the same forgetting rate. Adding state-dependent, element-wise gates at every token, however, changes the structure used by existing chunkwise algorithms. Direct state feedback makes each gate depend on preceding state updates, while generic element-wise decay removes the transition sharing used by these kernels. The challenge is therefore to expose memory information to the retention mechanism at a point where it can be used without disrupting the base computation.

\begin{samepage}
These constraints motivate our central question:
\par\nopagebreak[4]

\begingroup
\setlength{\fboxsep}{4pt}
\sbox{\sdgquestionbox}{%
  \colorbox{SDGQuestionBackground}{%
    \parbox{\dimexpr\linewidth-2\fboxsep-1.5pt\relax}{%
      \raggedright\textbf{Question.}\enspace
      \textit{How can we introduce state-dependent, element-wise forgetting while preserving parallel computation within each chunk?}}}}
\noindent\textcolor{SDGPurple}{%
  \rule[-\dp\sdgquestionbox]{1.5pt}{%
    \dimexpr\ht\sdgquestionbox+\dp\sdgquestionbox\relax}}%
\usebox{\sdgquestionbox}
\par
\endgroup
\nopagebreak[4]

We address this question with \emph{State-Dependent Gating} (\sdg), which applies the additional gate only at chunk boundaries (Figure~\ref{fig:sdg-overview}). After the base layer processes a chunk, \sdg subtracts the incoming state from the updated state and uses this difference to compute a separate retention coefficient for each state element. For these gated recurrences, the difference reflects how the chunk modifies the incoming memory through writing, erasing, and forgetting. The gate then scales the updated state before it enters the next chunk. Because the current chunk's outputs have already been computed, its algorithm and parallel factorization remain unchanged. Both boundary states are available from the recurrence. The added computation evaluates and applies the boundary gate.
\end{samepage}

The two levels of gating play different roles. Native token gates control the updates within each chunk, while the boundary gate controls the memory passed to subsequent chunks. This separation allows the added gate to use the completed chunk's effect on memory. It also makes the feedback interval explicit: the gate observes a net state change once per chunk.

We support this design with theoretical analysis and empirical evaluation. Under a signal--noise model, our analysis compares fixed and shared gates with an ideal gate that can vary across boundary observations and state elements. This quantifies the potential reconstruction benefits of adaptive, element-wise decay. We assess the learned method across recurrent and hybrid architectures, including controls that replace state-change feedback with chunk information alone.

\Needspace{6\baselineskip}
Our contributions are summarized as follows:
\par\nopagebreak[4]
\begin{itemize}
  \interlinepenalty=10000
  \setlength{\itemsep}{2pt}
  \setlength{\parsep}{0pt}
  \item \textbf{Boundary gating and fused implementation.} We introduce element-wise decay from chunk-level state changes, applied at boundaries to preserve the base layer's within-chunk algorithm. We develop fused kernels for GDN, KDA, Gated DeltaNet-2, and Mamba-3.
  \item \textbf{Theoretical characterization.} Under a signal--noise model, we derive an exact decomposition of the excess reconstruction error of fixed and shared boundary gates. We also characterize boundary decay, with the realized gate held fixed, as an element-wise ridge proximal step.
  \item \textbf{Empirical evaluation.} At the 1.3B scale with 100B training tokens, \sdg improves WikiText perplexity and average real-world retrieval across four recurrent backbones. Retrieval gains extend to three sliding-window hybrids, while reasoning and broader long-context results are mixed. Ablations examine gate inputs, granularity, and projection, and runtime measurements quantify the cost of boundary gating.
\end{itemize}

\section{Related Work}
\label{sec:related-work}

\paragraph{Recurrent models and forgetting.}
Linear attention~\citep{katharopoulos2020transformers} admits recurrent computation. Random-feature variants approximate softmax attention~\citep{choromanski2021performers,peng2021rfa}. S4~\citep{gu2022s4} and Mamba~\citep{gu2023mamba} use structured and input-dependent state-space dynamics, respectively. Mamba-3~\citep{lahoti2026mamba3} adds exponential-trapezoidal discretization and complex-valued transitions. RetNet~\citep{sun2023retnet}, GLA~\citep{yang2024gla}, and Mamba-2~\citep{dao2024mamba2} support efficient chunkwise computation.

\Needspace{3\baselineskip}
Gated memory designs include outer-product updates in HGRN2~\citep{qin2024hgrn2}, matrix memory in xLSTM's mLSTM~\citep{beck2024xlstm}, and slot-based reading in Gated Slot Attention~\citep{zhang2024gsa}. DeltaNet~\citep{schlag2021linear,yang2024deltanet} corrects stored associations using the current state. GDN~\citep{yang2024gdn} and KDA~\citep{kimi2025linear} add per-head scalar and per-key-channel decay, respectively. GDN-2~\citep{hatamizadeh2026gdn2} decouples erase and write gates. Their gate predictors use token representations without directly reading the same layer's recurrent memory. RWKV-7~\citep{peng2025rwkv7} generalizes the delta rule with vector-valued gates and in-context learning rates. Forgetting Transformer~\citep{lin2025forgetting} introduces data-dependent decay into softmax attention.

\paragraph{State feedback and transition expressivity.}
Comba~\citep{hu2025comba} combines a closed-loop control interpretation, scalar-plus-low-rank transitions, state and output corrections, and efficient chunkwise computation. Its state correction reads memory, while its scalar forget gate is predicted from the token representation. DeltaProduct~\citep{siems2025deltaproduct} composes multiple generalized Householder transformations per token to increase state-tracking expressivity. Complex KDA~\citep{siems2026complexkda} extends the ranges of KDA's channel-wise gate and delta coefficient to represent rotations while retaining diagonal-plus-rank-one transitions. \sdg uses the change in the same layer's memory across a chunk to predict element-wise boundary decay, retaining the supported backbones' native token gates and within-chunk update rules. This separates the information used to choose the boundary gate from the structure of the token-level transition. State feedback enters after the chunk computation, so the native recurrence does not require a new gate at each token.

\paragraph{Memory capacity and retrieval.}
Empirical recall gaps~\citep{arora2023zoology} and finite-precision bounds for retrieval tasks~\citep{wen2025rnns} motivate better use of recurrent memory. Block-Recurrent Transformers~\citep{hutchins2022block} use attention and gated state updates across blocks, while Compressive Transformers~\citep{rae2019compressive} store older activations in a separate compressed memory. More recently, Sparse Delta Memory~\citep{cabannes2026sdm} expands GDN's state through sparse addressing, and HOLA~\citep{cui2026hola} augments a delta-rule state with a bounded exact key--value cache, using update residuals to select stored associations. These approaches increase or supplement memory capacity. \sdg changes retention within the native state matrix, keeping its dimensions fixed and adding no retrieval cache. This distinction concerns the memory representation passed between chunks. Cached decoding still requires an additional source-state matrix to compute the boundary difference.

\paragraph{Online learning and memory optimization.}
Matrix associative memories store outer-product superpositions~\citep{kohonen1972correlation}. Fast-weight programmers learn context-dependent weight changes~\citep{schmidhuber1992fastweights}, while decaying fast-weight memories retain recent activity~\citep{ba2016fastweights}. Delta-rule models, test-time training, and Longhorn connect recurrent updates to online learning~\citep{schlag2021linear,sun2024ttt,liu2024longhorn}. Test-time regression~\citep{wang2026ttr} unifies sequence models through regression weights, function classes, and optimization algorithms. Titans~\citep{behrouz2025titans} combines gradient-based memory updates with momentum and forgetting. MesaNet~\citep{vonoswald2026mesanet} uses a chunkwise implementation and a conjugate-gradient solver for an in-context regression objective. Preconditioned DeltaNet~\citep{tumma2026preconditioned} introduces diagonal curvature corrections to delta-rule models with efficient chunkwise algorithms. These methods refine memory learning, while \sdg adds a learned retention step after the base chunk update. Its gate parameters are learned during training, and evaluating the gate at a chunk boundary does not require an iterative optimizer or an additional test-time regression solve.

\paragraph{Retention as regularization.}
MIRAS~\citep{behrouz2026miras} separates associative-memory objectives, retention regularization, and online optimization, and derives alternative forgetting mechanisms. Consistent with this connection between forgetting and regularization, \sdg boundary decay is an element-wise ridge proximal step when the realized gate is held fixed~\citep{duchi2009forwardbackward,duchi2010composite,parikh2014proximal}. Its penalty coefficients depend on chunk-level state changes, and shrinkage is applied at chunk boundaries. Wiener shrinkage~\citep{wiener1949extrapolation} informs our oracle reconstruction analysis of adaptive and shared boundary gates. The oracle comparison isolates the effects of gate adaptation and coefficient sharing under a fixed state distribution. It characterizes potential reconstruction gains, without assuming that the learned gate observes the useful memory and interference separately.

\Needspace{11\baselineskip}
\section{Preliminaries}
\label{sec:preliminaries}

\subsection{Recurrent state and native gates}
\label{sec:prelim-memory-control}

For one head, queries and keys $\bm q_t,\bm k_t\in\mathbb R^{d_k}$ read and write a state $\mathbf S_t\in\mathbb R^{d_v\times d_k}$, with value channels as rows and key channels as columns. Values and outputs satisfy $\bm v_t,\bm o_t\in\mathbb R^{d_v}$. The unnormalized additive recurrence is~\citep{katharopoulos2020transformers,schlag2021linear}
\begin{equation}
  \mathbf S_t=\mathbf S_{t-1}+\bm v_t\bm k_t^\top,
  \qquad
  \bm o_t=\mathbf S_t\bm q_t.
  \label{eq:state-read}
\end{equation}
The read includes the current write. Sequences start from $\mathbf S_0=\mathbf 0$. The state size is independent of sequence length.

In GDN, KDA, and GDN-2, native gates depend on token representations, whereas delta corrections read the recurrent state. GDN uses scalar decay, KDA uses key-channel decay, and GDN-2 separates erase and write gates. For fixed inputs, these layers share a key-space transition across value rows. Their token-dependent factors can therefore be prepared before the state scan. Appendix~\ref{app:extended-preliminaries} gives their token updates. Mamba-3 instead uses a $d_v\times d_s$ state without delta correction (Appendix~\ref{app:mamba3-chunkwise}).

\subsection{Chunkwise computation}
\label{sec:chunkwise-preliminary}

Training and prefill scan states across chunks with parallel computation within each chunk. Decoding uses the native token recurrence. For the formulas below, assume $L=NC$ with $N$ complete chunks of $C$ tokens. The incoming state is $\mathbf S_{[n]}=\mathbf S_{(n-1)C}$. Additive and gated delta-rule chunks have a shared transition $\mathbf P_{[n]}\in\mathbb R^{d_k\times d_k}$ and a zero-memory response $\boldsymbol\Omega_{[n]}\in\mathbb R^{d_v\times d_k}$:
\begin{equation}
  \begin{aligned}
    \mathbf S_{[n+1]}
      &=\mathbf S_{[n]}\mathbf P_{[n]}+\boldsymbol\Omega_{[n]},\\
    \mathbf S_{[n+1]}-\mathbf S_{[n]}
      &=\mathbf S_{[n]}(\mathbf P_{[n]}-\mathbf I)+\boldsymbol\Omega_{[n]}.
  \end{aligned}
  \label{eq:preliminary-affine-chunk}
\end{equation}
For fixed token inputs and native gates, both factors are independent of incoming memory. Thus gated state changes include the incoming-memory response; additive updates have $\mathbf P_{[n]}=\mathbf I$. Appendix~\ref{app:extended-preliminaries} expands the additive computation, and Appendix~\ref{app:chunk-closed-form} derives the gated forms.

\setlength{\textfloatsep}{\sdgsavedtextfloatsep}
\raggedbottom

\section{State-Dependent Gating}
\label{sec:method}

\subsection{Element-wise decay at chunk boundaries}

\sdg applies element-wise decay after the native chunk recurrence $\mathcal F$, retaining its token gates and within-chunk computation. Let $\mathbf S_{[n]}^+$ be the incoming state, $\mathbf S_{[n+1]}^-$ the updated state, and $\mathbf S_{[n+1]}^+$ the gated state. $\mathcal X_{[n]}$ contains projected token inputs and native gates, and $\bm\xi_{[n]}$ denotes auxiliary state. The computation returns the chunk outputs $\mathbf O_{[n]}\in\mathbb R^{C\times d_v}$:
\begin{equation}
  \begin{aligned}
    (\mathbf O_{[n]},\mathbf S_{[n+1]}^{-},\bm\xi_{[n+1]})
    &=\mathcal F(\mathcal X_{[n]},\mathbf S_{[n]}^{+},\bm\xi_{[n]}),\\
    \boldsymbol\Lambda_{[n]}
    &=\mathcal G_\theta(\mathbf S_{[n]}^{+},\mathbf S_{[n+1]}^{-}),
    \qquad\boldsymbol\Lambda_{[n]}\in(0,1]^{d_v\times d_k},\\
    \mathbf S_{[n+1]}^{+}
    &=\boldsymbol\Lambda_{[n]}\odot\mathbf S_{[n+1]}^{-}.
  \end{aligned}
  \label{eq:boundary-update}
\end{equation}
Decay scales the entire updated state, including the current chunk's writes. Outputs are formed first, so the gate affects only subsequent chunks. Native auxiliary states advance unchanged.

Token-level state feedback prevents gate precomputation from token representations alone, while general element-wise decay removes the shared value-row transitions of Section~\ref{sec:preliminaries}. Boundary placement preserves native within-chunk factorization and confines the additional dependency to the inter-chunk scan (Appendix~\ref{app:operator-view}).

\subsection{Gate parameterization}
\label{sec:gate-function}

The gate uses the state change, which includes both the incoming-memory response and the chunk contribution: $\Delta\mathbf S_{[n]}=\mathbf S_{[n]}^+(\mathbf P_{[n]}-\mathbf I)+\boldsymbol\Omega_{[n]}$ by Eq.~(\ref{eq:preliminary-affine-chunk}). Each head learns an unconstrained projection $\mathbf W_S\in\mathbb R^{d_k\times d_k}$ across key channels within each value row and an unconstrained element-wise gain $\mathbf G\in\mathbb R^{d_v\times d_k}$. These predict the gate without directly projecting the stored state:
\begin{align}
  \Delta\mathbf S_{[n]}&=\mathbf S_{[n+1]}^--\mathbf S_{[n]}^+,
  \quad\mathbf X_{[n]}=\Delta\mathbf S_{[n]}\mathbf W_S^\top,
  \quad\mathbf Z_{[n]}=\mathbf G\odot\mathbf X_{[n]},
  \label{eq:projected-state-change}\\
  \boldsymbol\rho_{[n]}&=\mathbf Z_{[n]}+
    \sqrt{\mathbf Z_{[n]}\odot\mathbf Z_{[n]}+\rho_0^2\mathbf 1},
  \label{eq:boundary-rate}\\
  \boldsymbol\Lambda_{[n]}&=\exp(-\boldsymbol\rho_{[n]}),
  \qquad\mathbf S_{[n+1]}^+=\boldsymbol\Lambda_{[n]}\odot\mathbf S_{[n+1]}^-.
  \label{eq:canonical-sdg}
\end{align}
Equation~(\ref{eq:boundary-rate}) is twice Squareplus~\citep{barron2021squareplus}, applied element-wise (definition and derivatives in Appendix~\ref{app:kernel-details}). Positive scores increase decay, while negative scores reduce it. For finite scores, $0<\boldsymbol\Lambda_{[n]}<1$. We initialize $\mathbf G=\mathbf0$ and fix $\rho_0=-\log\lambda_0$, $\lambda_0=0.9999$, so all gates initially equal $\lambda_0\mathbf1$. Appendix~\ref{app:gate-parameters} gives the remaining initialization details.

Algorithm~\ref{alg:sdg-forward} shows the computation order, with the added gate steps in bold. The native recurrence $\mathcal F$ retains each backbone's state and output computations. The algorithm assumes complete chunks, $L=NC$. For arbitrary lengths and cached continuations, Appendix~\ref{app:reference-boundary-scan} specifies the scan. Incomplete final chunks do not trigger boundary decay.

\begin{minipage}{\linewidth}
  \vspace{0.2em}
  \refstepcounter{algorithm}
  \label{alg:sdg-forward}
  \small
  \hrule height 1.1pt
  \vspace{0.25em}
  \noindent\textbf{Algorithm \thealgorithm}\quad \textsc{SDG}: Boundary-Gated Chunk Scan
  \vspace{0.2em}
  \hrule
  \vspace{0.3em}

  \noindent\textbf{Require:} {Sequence length $L=NC$, and $N=L/C$ complete chunks}, native token quantities $\mathcal X$, initial state $\mathbf S_{[1]}^+$ and auxiliary state $\bm\xi_{[1]}$, chunk size $C$, gate parameters $\mathbf W_S,\mathbf G$, and fixed base rate $\rho_0$.

  \setcounter{sdgalgorithmline}{0}
  \sdgalgline{Partition $\mathcal X$ into $N$ complete chunks of $C$ tokens, $\{\mathcal X_{[n]}\}_{n=1}^N$.}
  \sdgalgline{\textbf{for} $n=1,\ldots,N$ \textbf{do}}
  \sdgalgline[1.2em]{\bfseries Retain the incoming state $\mathbf S^{\src}\gets\mathbf S_{[n]}^+$.}
  \sdgalgline[1.2em]{Compute $(\mathbf O_{[n]},\mathbf S_{[n+1]}^-,\bm\xi_{[n+1]})\gets\mathcal F(\mathcal X_{[n]},\mathbf S_{[n]}^+,\bm\xi_{[n]})$.}
  \sdgalgline[1.2em]{\bfseries Compute $\Delta\mathbf S_{[n]}\gets\mathbf S_{[n+1]}^--\mathbf S^{\src}$.}
  \sdgalgline[1.2em]{\bfseries Compute $\mathbf X_{[n]}\gets\Delta\mathbf S_{[n]}\mathbf W_S^\top$, $\mathbf Z_{[n]}\gets\mathbf G\odot\mathbf X_{[n]}$.}
  \sdgalgline[1.2em]{\bfseries Compute $\boldsymbol\rho_{[n]}\gets2\operatorname{squareplus}(\mathbf Z_{[n]};\rho_0^2)$, $\boldsymbol\Lambda_{[n]}\gets\exp(-\boldsymbol\rho_{[n]})$.}
  \sdgalgline[1.2em]{\bfseries Set $\mathbf S_{[n+1]}^+\gets\boldsymbol\Lambda_{[n]}\odot\mathbf S_{[n+1]}^-$.}
  \sdgalgline{\textbf{end for}}
  \sdgalgline{\textbf{return} $\operatorname{Concat}(\mathbf O_{[1]},\ldots,\mathbf O_{[N]})$, $\mathbf S_{[N+1]}^+$, and $\bm\xi_{[N+1]}$.}
  \vspace{0.3em}
  \hrule height 1.1pt
\end{minipage}

\subsection{Fused kernel implementation}
\label{sec:fused-kernel}

The boundary update is fused into the inter-chunk state scan using Triton~\citep{tillet2019triton}, while retaining native factor preparation and output kernels. Appendix~\ref{app:kernel-details} gives forward and backward kernel details, with reference algorithms for additive linear attention in Algorithms~\ref{alg:sdg-additive-forward} and~\ref{alg:sdg-backward}, respectively. Each head adds $O(d_vd_k^2)$ projection and $O(d_vd_k)$ element-wise work per boundary, amortized over $C$ tokens, plus backward computation (Appendix~\ref{app:flops}). For Mamba-3, replace $d_k$ by $d_s$. Cached decoding retains an additional source matrix and token counter. We use $C=64$ to balance recurrent overhead and on-chip memory usage, matching the chunk size of the adopted Flash Linear Attention kernels~\citep{yang2024fla}. We apply the same boundary rules during training, prefill, and decoding.

\section{Analysis}
\label{sec:analysis}

We characterize the realized boundary operation, then quantify potential gains from ideal reconstruction gates under a fixed upstream state distribution.

\subsection{Boundary decay as an adaptive element-wise ridge}
\label{sec:composite-online-learning}

Holding the realized gate fixed, boundary decay is an element-wise ridge proximal step.

\begin{remark}[Adaptive element-wise ridge]
\label{rem:ridge}
Fix $\boldsymbol\rho_{[n]}\in\mathbb R_{>0}^{d_v\times d_k}$, let $\boldsymbol\Lambda_{[n]}=\exp(-\boldsymbol\rho_{[n]})$, and set $\boldsymbol\kappa=\exp(\boldsymbol\rho_{[n]})-\mathbf 1$ element-wise. Then for any state $\mathbf S_{[n+1]}^-$ before boundary decay,
\begin{equation}
  \mathbf S_{[n+1]}^{+}
  =\boldsymbol\Lambda_{[n]}\odot\mathbf S_{[n+1]}^{-}
  =\argmin_{\mathbf M\in\mathbb R^{d_v\times d_k}}
   \left\{
     \tfrac12\bigl\|\mathbf M-\mathbf S_{[n+1]}^{-}\bigr\|_F^2
     +\tfrac12\sum_{a,b}\kappa_{ab}M_{ab}^2
   \right\}.
  \label{eq:ridge-equivalence}
\end{equation}
\end{remark}

The rates set separate ridge weights, with $\boldsymbol\kappa=\boldsymbol\rho_{[n]}+O(\boldsymbol\rho_{[n]}^{\odot2})$ at small rates. This identity applies to all four backbones and concerns only the boundary operation. Composing it with a chunk's token updates is not generally one forward--backward step. Appendix~\ref{app:ridge-proof} gives the proof, converse, and backbone-specific token-update interpretations.

\subsection{Optimal decay and the cost of sharing it}
\label{sec:oracle-decay}

At one boundary, keep the upstream distribution fixed and model the state as useful memory plus residual interference~\citep{kohonen1972correlation,schlag2021linear}:
\begin{equation}
  \mathbf S^-=\mathbf S^\star+\mathbf N,
  \label{eq:signal-noise-state}
\end{equation}
Here $\mathbf S^\star$ is useful to future queries and $\mathbf N$ is residual interference; neither is separately observed by the learned gate. For oracle information $\Phi$, a gate is \emph{admissible} if it is $\Phi$-measurable with finite reconstruction risk $\mathcal R(\boldsymbol\Lambda)=\mathbb E\|\boldsymbol\Lambda\odot\mathbf S^--\mathbf S^\star\|_F^2$. An independent isotropic query makes this risk proportional to readout error; other query models can change the optimum (Appendix~\ref{app:readout-derivation}).

\begin{lemma}[Conditional shrinkage and excess risk]
\label{lem:conditional-wiener}
Assume that $\mathbf S^\star$ and $\mathbf N$ have finite second moments and that $\mathbf m_-=\mathbb E[\mathbf S^-\odot\mathbf S^-\mid\Phi]>\mathbf 0$ almost surely, and define
\begin{equation}
  \boldsymbol\Lambda^\star
  =\mathbb E[\mathbf S^\star\odot\mathbf S^-\mid\Phi]\oslash\mathbf m_- .
  \label{eq:conditional-wiener-gate}
\end{equation}
Then every admissible gate satisfies
\begin{equation}
  \mathcal R(\boldsymbol\Lambda)
  =\mathcal R(\boldsymbol\Lambda^\star)
   +\mathbb E\bigl\|\mathbf m_-^{1/2}\odot
     (\boldsymbol\Lambda-\boldsymbol\Lambda^\star)\bigr\|_F^2 .
  \label{eq:conditional-excess}
\end{equation}
If in addition $\mathbb E[\mathbf S^\star\odot\mathbf N\mid\Phi]=\mathbf 0$, then $\boldsymbol\Lambda^\star=\mathbf m_\star\oslash(\mathbf m_\star+\mathbf m_N)\in[0,1]^{d_v\times d_k}$, with $\mathbf m_\star=\mathbb E[\mathbf S^\star\odot\mathbf S^\star\mid\Phi]$, $\mathbf m_N=\mathbb E[\mathbf N\odot\mathbf N\mid\Phi]$ and $\mathbf m_-=\mathbf m_\star+\mathbf m_N$. All operations are element-wise.
\end{lemma}

We compare fixed boundary coefficients with the adaptive oracle under the same state distribution. The fixed scalar and per-key families below describe coefficient sharing, not native GDN or KDA gates, which can vary with token inputs.

\begin{corollary}[Cost of fixing and of sharing the gate]
\label{cor:sharing-cost}
Let $\bar{\mathbf m}=\mathbb E[\mathbf m_-]$ be finite and strictly positive. Consider
\begin{equation*}
  \mathcal C_{\mathrm{scalar}}=\{\lambda\mathbf 1\}
  \subset
  \mathcal C_{\mathrm{key}}=\{\mathbf 1_{d_v}\bm\lambda^\top\}
  \subset
  \mathcal C_{\mathrm{elem}}=\mathbb R^{d_v\times d_k}
  \subset
  \mathcal C_{\mathrm{adapt}}=\{\text{admissible gates}\},
\end{equation*}
where the first three contain only constants. Define the optimal constant element-wise gate and its weighted averages by
\begin{equation*}
  \begin{aligned}
    \bar{\boldsymbol\Lambda}
    &=\mathbb E[\mathbf S^\star\odot\mathbf S^-]\oslash\bar{\mathbf m},
    &\bar m_{\cdot b}&=\sum_a\bar m_{ab},\\
    \mu_b&=\frac{\sum_a\bar m_{ab}\bar\Lambda_{ab}}{\bar m_{\cdot b}},
    &\mu&=\frac{\sum_b\bar m_{\cdot b}\mu_b}{\sum_b\bar m_{\cdot b}}.
  \end{aligned}
\end{equation*}
Then $\bar{\boldsymbol\Lambda}$, $\mathbf 1_{d_v}\bm\mu^\top$, and $\mu\mathbf 1$ minimize $\mathcal R$ over the constant element-wise, per-key, and scalar families, respectively, and
\begin{equation}
  \min_{\mathcal C_{\mathrm{scalar}}}\mathcal R
  -\min_{\mathcal C_{\mathrm{adapt}}}\mathcal R
  =\underbrace{\mathbb E\bigl\|\mathbf m_-^{1/2}\odot
      (\boldsymbol\Lambda^\star-\bar{\boldsymbol\Lambda})\bigr\|_F^2}
      _{\mathcal E_{\mathrm{state}}}
  +\underbrace{\sum_{a,b}\bar m_{ab}(\bar\Lambda_{ab}-\mu_b)^2}
      _{\mathcal E_{\mathrm{value}}}
  +\underbrace{\sum_{b}\bar m_{\cdot b}(\mu_b-\mu)^2}
      _{\mathcal E_{\mathrm{key}}}.
  \label{eq:sharing-cost}
\end{equation}
Dropping $\mathcal E_{\mathrm{key}}$ gives the per-key family's excess risk; dropping the last two gives that of the constant element-wise family. If $\boldsymbol\Lambda^\star\in[0,1]^{d_v\times d_k}$ almost surely, all four minimizers lie in this box. Constraining the families to the box then leaves Eq.~(\ref{eq:sharing-cost}) unchanged.
\end{corollary}

The three nonnegative costs respectively arise from fixing coefficients across observations, sharing those fixed coefficients across value rows, and further sharing across key channels. Each vanishes when the corresponding optimal coefficients already satisfy the restriction. Appendix~\ref{app:conditional-wiener-proof} gives the conditional-shrinkage and projection proofs, oracle-rate endpoints, and adaptive-sharing counterpart. Appendix~\ref{app:superposition-example} illustrates when costs are positive.

These are potential gains from adapting to $\Phi$. The first term alone does not measure information beyond token inputs. Constant scalar/per-key families are not native token-dependent GDN/KDA gates. The implemented gate uses only $\Delta\mathbf S_{[n]}$ through a restricted parameterization, which need not contain the full boundary-state pair's information. The analysis neither identifies the gate found by training nor predicts trained-model ordering. Moreover, the risk in Eq.~(\ref{eq:sharing-cost}) is defined on unnormalized states. Because a non-uniform gate can also change a readout's direction, the decomposition does not by itself characterize prediction error after RMS normalization.
Table~\ref{tab:ablation} evaluates the gate choices empirically; Appendix~\ref{app:proofs} provides the complete proofs and additional results.

\section{Experiments}
\label{sec:experiments}
\begingroup
\renewcommand{\modelsep}{\addlinespace[2pt]}
\setlength{\intextsep}{8pt plus 2pt minus 2pt}
\setlength{\floatsep}{8pt plus 2pt minus 2pt}
\setlength{\textfloatsep}{12pt plus 2pt minus 2pt}

We evaluate \sdg's effects on prediction quality, the contributions of its gate design, and its computational overhead. We also examine the local output changes caused by a single boundary update.

\subsection{Experimental setup}

We train paired GDN, KDA, Mamba-3 (SISO), and GDN-2 models on 100B FineWeb-Edu tokens~\citep{lozhkov2024finewebedu}, using sequences of 4096 tokens and chunk size $C=64$. Each model has approximately 1.3B non-embedding parameters. Its +\sdg counterpart adds a boundary gate to every recurrent layer. Table~\ref{tab:model-parameters} gives parameter counts. Each pair shares the data order, optimizer settings, and initial values of shared parameters. Training details are in Appendix~\ref{app:training-details}.

Evaluation covers language modeling, nine zero-shot accuracy tasks, four RULER needle-in-a-haystack tasks, six real-world retrieval tasks, and all 21 LongBench tasks. We also evaluate hybrid GDN, KDA, and GDN-2 models that alternate recurrent layers with 2K sliding-window attention.

\subsection{Performance across backbones}

\paragraph{Language modeling and zero-shot accuracy.}
\sdg lowers WikiText perplexity by 0.10--0.58 across the four recurrent backbones (Table~\ref{tab:lm-reasoning}). LAMBADA perplexity improves for three models and rises slightly for KDA. Average zero-shot accuracy increases for GDN, Mamba-3, and GDN-2, while KDA is unchanged at the reported precision. Task-level effects remain mixed, including lower BoolQ accuracy in all four pairs.

\begin{table*}[!htb]
  \setlength{\belowcaptionskip}{6pt}
  \caption{Language modeling and zero-shot accuracy. Avg. is the mean over the nine accuracy columns; acc$_n$ denotes length-normalized accuracy. Bold marks the better result within each pair.}
  \label{tab:lm-reasoning}
  \centering
  \footnotesize
  \setlength{\tabcolsep}{3pt}
  \renewcommand{\arraystretch}{1.14}
  \settowidth{\modelcolwidth}{\textbf{Mamba-3+SDG}}
  \addtolength{\modelcolwidth}{2pt}
  \setlength{\metriccolwidth}{\dimexpr(\linewidth-\modelcolwidth-2\tabcolsep-\arrayrulewidth-6pt)/12-2\tabcolsep\relax}
  \begin{tabular}{p{\modelcolwidth}*{2}{C{\metriccolwidth}}!{\vrule width \arrayrulewidth}*{4}{C{\metriccolwidth}}*{2}{C{\dimexpr\metriccolwidth+3pt\relax}}*{4}{C{\metriccolwidth}}}
    \toprule
    Model & \makebox[\linewidth][c]{Wiki.} & \makebox[\linewidth][c]{LMB.} & \makebox[\linewidth][c]{LMB.} & \makebox[\linewidth][c]{PIQA} & \makebox[\linewidth][c]{Hella.} & \makebox[\linewidth][c]{Wino.} & \makebox[\linewidth][c]{ARC-e} & \makebox[\linewidth][c]{ARC-c} & \makebox[\linewidth][c]{OBQA} & \makebox[\linewidth][c]{SIQA} & \makebox[\linewidth][c]{BoolQ} & \makebox[\linewidth][c]{Avg.} \\
    & ppl $\downarrow$ & ppl $\downarrow$ & acc $\uparrow$ & acc $\uparrow$ & \makebox[\linewidth][c]{acc$_n$ $\uparrow$} & acc $\uparrow$
      & acc $\uparrow$ & \makebox[\linewidth][c]{acc$_n$ $\uparrow$} & acc $\uparrow$ & acc $\uparrow$ & acc $\uparrow$ & acc $\uparrow$ \\
    \midrule
    GDN & 16.57 & 11.87 & 46.28 & 72.20 & 44.03 & \textbf{56.67} & 70.62 & \textbf{35.84} & 30.80 & 40.63 & \textbf{59.54} & 50.73 \\
    \rowcolor{SDGLightPurple}\textbf{GDN+\sdg} & \textbf{15.99} & \textbf{11.64} & \textbf{47.70} & \textbf{73.67} & \textbf{44.23} & 56.43 & \textbf{72.39} & 35.67 & \textbf{31.20} & \textbf{40.69} & 58.78 & \textbf{51.20} \\
    \modelsep
    KDA & 16.01 & \textbf{11.11} & 48.57 & \textbf{72.52} & 43.50 & \textbf{57.37} & 70.37 & 35.84 & \textbf{31.40} & 40.63 & \textbf{60.31} & \textbf{51.17} \\
    \rowcolor{SDGLightPurple}\textbf{KDA+\sdg} & \textbf{15.91} & 11.15 & \textbf{49.04} & 72.47 & \textbf{43.91} & 56.60 & \textbf{70.50} & \textbf{36.43} & \textbf{31.40} & \textbf{40.80} & 59.39 & \textbf{51.17} \\
    \modelsep
    Mamba-3 & 15.55 & 11.93 & 46.98 & 72.91 & 44.83 & 55.09 & 72.80 & 36.52 & \textbf{31.20} & 40.58 & \textbf{60.55} & 51.27 \\
    \rowcolor{SDGLightPurple}\textbf{Mamba-3+\sdg} & \textbf{15.34} & \textbf{11.37} & \textbf{48.09} & \textbf{73.18} & \textbf{44.93} & \textbf{57.30} & \textbf{72.98} & \textbf{37.37} & 30.80 & \textbf{40.94} & 59.91 & \textbf{51.72} \\
    \modelsep
    GDN-2 & 15.57 & 11.55 & 47.72 & \textbf{73.23} & 44.31 & 58.17 & 71.38 & 36.01 & \textbf{30.40} & 40.58 & \textbf{60.09} & 51.32 \\
    \rowcolor{SDGLightPurple}\textbf{GDN-2+\sdg} & \textbf{15.40} & \textbf{11.31} & \textbf{47.89} & 73.07 & \textbf{44.80} & \textbf{59.51} & \textbf{72.47} & \textbf{37.20} & \textbf{30.40} & \textbf{40.89} & 58.53 & \textbf{51.64} \\
    \bottomrule
  \end{tabular}
\end{table*}

\paragraph{In-context retrieval on synthetic data.}
On the evaluated RULER~\citep{hsieh2024ruler} tasks, multi-key retrieval improves across all four backbones at 2K and 4K (Table~\ref{tab:niah-results}). At 4K, MK-NIAH-1 gains are 0.2, 4.4, 13.6, and 9.8 percentage points for GDN, KDA, Mamba-3, and GDN-2, respectively. Single-needle results are less uniform: KDA falls from 36.4\% to 24.2\% on S-NIAH-3 at 4K. Thus, improvements vary even within the task length. The two tasks evaluated at 8K also give mixed results, so these gains do not establish better length extrapolation.

\begin{table*}[!htb]
  \setlength{\belowcaptionskip}{6pt}
  \caption{RULER retrieval accuracy (\%) for recurrent models trained at 4K. Columns give input lengths. Bold marks the better result within each baseline--\sdg pair.}
  \label{tab:niah-results}
  \centering
  \footnotesize
  \setlength{\tabcolsep}{2.5pt}
  \renewcommand{\arraystretch}{1.14}
  \settowidth{\modelcolwidth}{\textbf{Mamba-3+SDG}}
  \addtolength{\modelcolwidth}{2pt}
  \begin{tabular}{p{\modelcolwidth}*{14}{C{\dimexpr(\linewidth-\modelcolwidth-2\tabcolsep)/14-2\tabcolsep\relax}}}
    \toprule
    Model & \multicolumn{4}{c}{S-NIAH-1} & \multicolumn{4}{c}{S-NIAH-2}
      & \multicolumn{3}{c}{S-NIAH-3} & \multicolumn{3}{c}{MK-NIAH-1} \\
    \cmidrule(lr){2-5}\cmidrule(lr){6-9}\cmidrule(lr){10-12}\cmidrule(lr){13-15}
    & 1K & 2K & 4K & 8K & 1K & 2K & 4K & 8K & 1K & 2K & 4K & 1K & 2K & 4K \\
    \midrule
    GDN & 99.8 & \textbf{100} & \textbf{100} & \textbf{84.6} & 99.8 & \textbf{100} & 85.2 & 34.0 & 82.8 & 63.4 & 39.6 & 39.8 & 31.6 & 24.0 \\
    \rowcolor{SDGLightPurple}\textbf{GDN+\sdg} & \textbf{100} & \textbf{100} & \textbf{100} & 78.6 & \textbf{100} & \textbf{100} & \textbf{93.0} & \textbf{39.2} & \textbf{90.4} & \textbf{70.6} & \textbf{49.8} & \textbf{40.8} & \textbf{33.6} & \textbf{24.2} \\
    \modelsep
    KDA & \textbf{100} & \textbf{100} & 94.0 & 83.6 & \textbf{100} & \textbf{100} & 89.4 & 32.0 & 83.4 & 68.2 & \textbf{36.4} & 37.8 & 31.6 & 22.4 \\
    \rowcolor{SDGLightPurple}\textbf{KDA+\sdg} & \textbf{100} & \textbf{100} & \textbf{98.6} & \textbf{87.2} & \textbf{100} & 99.4 & \textbf{92.6} & \textbf{37.8} & \textbf{91.4} & \textbf{72.8} & 24.2 & \textbf{38.8} & \textbf{34.0} & \textbf{26.8} \\
    \modelsep
    Mamba-3 & 99.8 & 99.4 & 86.0 & \textbf{46.6} & 99.6 & 99.6 & 76.6 & \textbf{34.4} & 66.4 & 54.4 & 22.2 & \textbf{59.4} & 52.6 & 26.0 \\
    \rowcolor{SDGLightPurple}\textbf{Mamba-3+\sdg} & \textbf{100} & \textbf{100} & \textbf{92.8} & 39.6 & \textbf{100} & \textbf{99.8} & \textbf{78.8} & \textbf{34.4} & \textbf{75.2} & \textbf{66.2} & \textbf{25.6} & 58.0 & \textbf{54.8} & \textbf{39.6} \\
    \modelsep
    GDN-2 & \textbf{100} & \textbf{100} & \textbf{100} & \textbf{99.8} & \textbf{100} & 99.6 & \textbf{93.2} & 35.2 & \textbf{94.6} & 84.2 & 37.8 & 45.4 & 43.8 & 30.4 \\
    \rowcolor{SDGLightPurple}\textbf{GDN-2+\sdg} & \textbf{100} & \textbf{100} & \textbf{100} & 94.8 & \textbf{100} & \textbf{99.8} & 90.0 & \textbf{53.4} & \textbf{94.6} & \textbf{92.0} & \textbf{49.2} & \textbf{61.0} & \textbf{63.4} & \textbf{40.2} \\
    \bottomrule
  \end{tabular}
\end{table*}

\paragraph{In-context retrieval on real-world tasks.}
Following \citet{arora2024jrt}, we evaluate six real-world retrieval tasks (Table~\ref{tab:recall-results}). \sdg raises reported mean accuracy by 0.39--0.83 percentage points across the four backbones. 

\begin{table*}[!htbp]
  \setlength{\belowcaptionskip}{6pt}
  \caption{Real-world retrieval accuracy (\%) for recurrent models. Avg. is the mean over the six tasks. SQD, TQA, and NQ denote SQuAD, TriviaQA, and Natural Questions. Best results within each model family are bold (ties included).}
  \label{tab:recall-results}
  \centering
  \footnotesize
  \setlength{\tabcolsep}{5pt}
  \renewcommand{\arraystretch}{1.14}
  \settowidth{\modelcolwidth}{\textbf{Mamba-3+SDG}}
  \addtolength{\modelcolwidth}{2pt}
  \begin{tabular}{p{\modelcolwidth}*{7}{C{\dimexpr(\linewidth-\modelcolwidth-2\tabcolsep)/7-2\tabcolsep\relax}}}
    \toprule
    Model & SWDE & SQD & FDA & TQA & NQ & DROP & Avg. \\
    \midrule
    GDN & 24.37 & \textbf{37.92} & \textbf{23.07} & 60.19 & 21.16 & 20.99 & 31.28 \\
    \rowcolor{SDGLightPurple}\textbf{GDN+\sdg} & \textbf{26.80} & 37.15 & 22.42 & \textbf{60.84} & \textbf{21.82} & \textbf{22.40} & \textbf{31.91} \\
    \modelsep
    KDA & 26.34 & 38.26 & 23.62 & \textbf{62.68} & \textbf{23.12} & \textbf{21.18} & 32.53 \\
    \rowcolor{SDGLightPurple}\textbf{KDA+\sdg} & \textbf{27.74} & \textbf{38.97} & \textbf{24.89} & 62.62 & 22.71 & 20.60 & \textbf{32.92} \\
    \modelsep
    Mamba-3 & 27.93 & \textbf{37.05} & \textbf{26.25} & 61.73 & 21.03 & 21.47 & 32.58 \\
    \rowcolor{SDGLightPurple}\textbf{Mamba-3+\sdg} & \textbf{29.90} & 36.98 & 25.79 & \textbf{62.38} & \textbf{22.17} & \textbf{21.75} & \textbf{33.16} \\
    \modelsep
    GDN-2 & 27.93 & 37.76 & \textbf{25.52} & 61.85 & 21.95 & \textbf{21.80} & 32.80 \\
    \rowcolor{SDGLightPurple}\textbf{GDN-2+\sdg} & \textbf{31.02} & \textbf{37.79} & 25.40 & \textbf{63.09} & \textbf{22.65} & \textbf{21.80} & \textbf{33.63} \\
    \bottomrule
  \end{tabular}
\end{table*}

\paragraph{Long-context understanding.}
On LongBench~\citep{bai2023longbench}, SDG improves average performance for GDN, KDA, and Mamba-3. The Code category score increases across all four models. Appendix~\ref{app:longbench-results} reports the full results and analysis.

\paragraph{Hybrid models.}
Figure~\ref{fig:hybrid-results} shows results for hybrid GDN, KDA, and GDN-2. SDG lowers LAMBADA perplexity and improves average real-world retrieval accuracy across all three models. S-NIAH-3 accuracy also improves at 4K. LongBench scores increase for KDA and GDN-2. These results suggest that SDG can complement hybrid architectures. Appendix~\ref{app:hybrid-results} reports the full results.

\begin{figure}[!htb]
  \centering
  \setlength{\parskip}{0pt}
  \includegraphics[width=\linewidth,trim=0 8bp 0 0,clip]{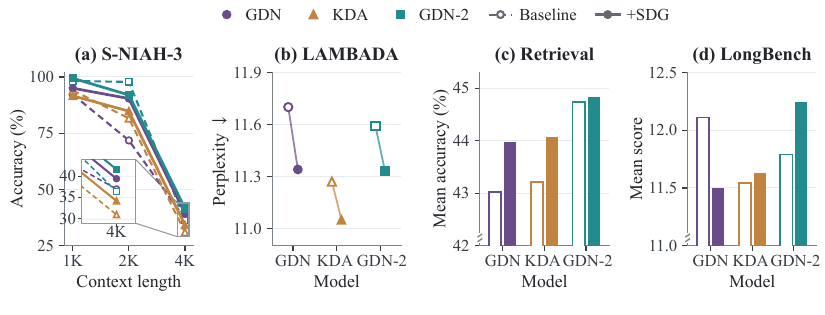}
  \setlength{\abovecaptionskip}{-10pt}
  \caption{S-NIAH-3 accuracy, LAMBADA perplexity, mean six-task retrieval accuracy, and mean 21-task LongBench score on hybrid models. Open and filled marks denote baseline and +\sdg.}
  \label{fig:hybrid-results}
\end{figure}

\subsection{Ablations and boundary diagnostics}

\begin{table}[!htb]
  \centering
  \small
  \setlength{\abovecaptionskip}{0pt}
  \setlength{\belowcaptionskip}{6pt}
  \caption{Ablations of SDG on GDN. NIAH@4K averages S-NIAH-2, S-NIAH-3, and MK-NIAH-1 at 4K. Bold marks the highest score in each metric.}
  \label{tab:ablation}
  \renewcommand{\modelsep}{\noalign{\vskip 1pt}\hdashline\noalign{\vskip 1pt}}
  \begin{minipage}{0.88\textwidth}
  \centering
  \footnotesize
  \setlength{\tabcolsep}{1.5pt}
  \renewcommand{\arraystretch}{1.14}
  \settowidth{\modelnamewidth}{GDN+\sdg{}}
  \settowidth{\modelcolwidth}{\textbf{GDN+\sdg{}}}
  \ifdim\modelcolwidth>\modelnamewidth
    \setlength{\modelnamewidth}{\modelcolwidth}
  \fi
  \settowidth{\modelcolwidth}{\makebox[\modelnamewidth][l]{GDN+\sdg{}}\,{\scriptsize(Channel-wise)}}
  \addtolength{\modelcolwidth}{4pt}
  \setlength{\metriccolwidth}{\dimexpr(\linewidth-\modelcolwidth-10\tabcolsep-\arrayrulewidth)/4\relax}
  \begin{tabular}{p{\modelcolwidth}!{\vrule width \arrayrulewidth}*{4}{C{\metriccolwidth}}}
    \toprule
    \multirow{2}{*}{\textbf{Variant}} & \textbf{Commonsense} & \textbf{NIAH}
      & \textbf{Recall} & \textbf{LongBench} \\
    & avg $\uparrow$ & @4K $\uparrow$ & avg $\uparrow$ & avg $\uparrow$ \\
    \midrule
    \multicolumn{5}{l}{\textit{Gate granularity}} \\
    GDN
      & 50.73 & 49.60 & 31.28 & 10.37 \\
    \mbox{\makebox[\modelnamewidth][l]{GDN+\sdg{}}\,
      {\scriptsize(Channel-wise)}}
      & 51.06 & 54.20 & 31.69 & 10.71 \\
    \rowcolor{SDGLightPurple}
    \mbox{\makebox[\modelnamewidth][l]{\textbf{GDN+\sdg{}}}\,
      {\scriptsize(Element-wise)}}
      & \textbf{51.20} & \textbf{55.67} & \textbf{31.91} & 10.95 \\
    \modelsep
    \multicolumn{5}{l}{\textit{Gate input}} \\
    Fixed boundary decay
      & 50.89 & 42.87 & 28.87 & 9.79 \\
    Learned static gate
      & 50.76 & 48.73 & 29.32 & 10.03 \\
    Gate input $\boldsymbol\Omega$
      & 51.15 & 55.00 & 31.89 & \textbf{11.13} \\
    Gate input $\mathbf S_{[n]}^+$
      & 50.94 & 52.60 &30.84 & 10.91\\
    \modelsep
    \multicolumn{5}{l}{\textit{Gate projection}} \\
    Fixed $\mathbf W_S=\mathbf I$
      & 50.99 & 51.13 & 31.82 & 10.86 \\
    \bottomrule
  \end{tabular}
  \end{minipage}
\end{table}

\paragraph{Gate inputs.}
Table~\ref{tab:ablation} compares gate inputs, coefficient sharing, and projection on GDN. Fixed boundary decay and a learned static gate score below \sdg on all four metrics. To examine within-layer state feedback, the chunk-input control replaces $\Delta\mathbf S_{[n]}$ with $\boldsymbol\Omega_{[n]}$, the native chunk contribution from a zero incoming state. For fixed chunk inputs, $\boldsymbol\Omega_{[n]}$ excludes the response of that layer's incoming memory, while token representations can still carry context from earlier layers. The dynamic inputs use the same gate parameterization. Constructing $\boldsymbol\Omega_{[n]}$ requires an additional matrix product, whereas $\Delta\mathbf S_{[n]}$ uses a subtraction of two available states. Appendix~\ref{app:ablation-details} compares their GDN operator latencies in forward and backward computation.

\Needspace{3\baselineskip}
The $\boldsymbol\Omega$ control recovers most of \sdg's gains. Using $\Delta\mathbf S_{[n]}$ instead raises NIAH@4K by 0.67 points and real-world retrieval by 0.02 points, but lowers LongBench by 0.18 points. The difference between these two inputs is modest and task-dependent. Using the incoming state $\mathbf S_{[n]}^+$ gives lower scores than the state-change input on all four metrics. The reported scores are point estimates without uncertainty estimates; small gaps do not establish a statistically reliable advantage.

\paragraph{Sharing and projection.}
Element-wise gating scores above channel-wise gating on all four metrics, including a 1.47-point gain on NIAH@4K. The channel-wise variant shares one retention coefficient per key channel across all value rows and retains the same parameter count as element-wise gating. Learning $\mathbf W_S$ also improves all four scores relative to fixing $\mathbf W_S=\mathbf I$, with a 4.54-point gain on NIAH@4K. The static and identity-projection variants use fewer parameters. Appendix~\ref{app:ablation-details} details these controls and costs.

\begin{figure}[!htbp]
  \centering
  \setlength{\parskip}{0pt}
  \includegraphics[width=\linewidth]{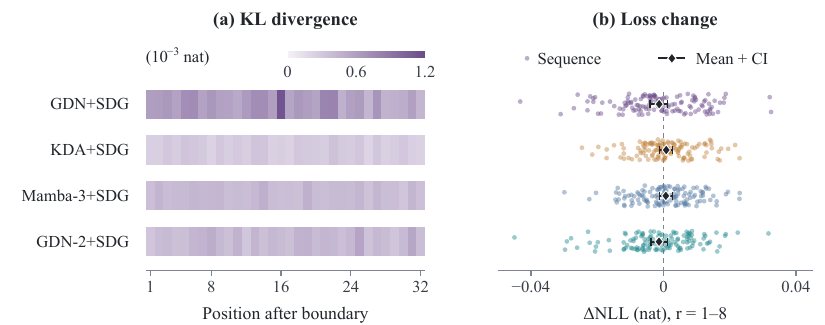}
  \setlength{\abovecaptionskip}{3pt}
  \caption{Effects of one SDG boundary update, with 128 observations per model. \textbf{(a)} The heatmap shows mean KL divergence from skip to normal at each of the first 32 affected positions. \textbf{(b)} Colored points show sequence means of $\Delta$NLL over positions 1--8, defined as normal minus skip. Black diamonds and bars show means and bootstrap intervals adjusted across four models for approximately 95\% simultaneous coverage.}
  \label{fig:boundary-stability}
\end{figure}
\paragraph{Boundary output stability.}
We compare normal execution with skipping one boundary update across all recurrent layers of the same trained model. The two branches start from identical caches and consume the same subsequent ground-truth tokens. Over the first eight affected positions, top-1 predictions agree at 98.34--99.12\%, and mean Kullback--Leibler (KL) divergence is below $6.5\times10^{-4}$. The four-model-adjusted intervals for mean negative log-likelihood (NLL) change all contain zero, with upper bounds below 0.003 nat per token (Figure~\ref{fig:boundary-stability}).

\Needspace{3\baselineskip}
These small averages coexist with individual loss increases and decreases. They do not bound per-token changes or characterize free-running generation. Mamba-3 contributes summary statistics. Sample matching and intervention controls were checked for GDN, KDA, and GDN-2. We also use valid filler tokens to place answer generation before and after chunk boundaries across benchmarks, with absolute changes in accuracy of at most 0.5 percentage points on commonsense reasoning and 1 percentage point on retrieval. Appendix~\ref{app:boundary-stability} gives the protocol and results.

\subsection{Efficiency}

\paragraph{Training throughput.}
We measure end-to-end training throughput for recurrent-only models on a single NVIDIA H200, with 16,384 input tokens per optimizer step. Across the tested 2K--16K sequence lengths, \sdg retains 85.0--89.4\% of baseline throughput for GDN, KDA, and GDN-2, and 72.5--75.1\% for Mamba-3. Preserving within-chunk computation therefore still incurs a measurable training cost. GDN and GDN-2 with \sdg show a smaller throughput decline across the tested lengths than the full-attention reference (Figure~\ref{fig:inference-latency}). See Appendix~\ref{app:efficiency} for details.

\paragraph{Inference latency.}
Inference measurements use hybrids with alternating recurrent and 2K SWA layers, on a single NVIDIA H200 at batch size one. For GDN and GDN-2 with \sdg, time to first token (TTFT) grows approximately linearly at long contexts. Time per output token (TPOT) stays nearly constant for prompts from 512 to 512K tokens (Figure~\ref{fig:inference-latency}). At 256K, \sdg adds 0.23 and 0.25 seconds to their TTFT, yielding 1.75 and 1.91 seconds, compared with 18.34 seconds for the full-attention reference. KDA incurs prefill and decoding overhead, whereas Mamba-3 has higher TTFT but lower TPOT (Appendix~\ref{app:efficiency-tables}). The models retain favorable long-context scaling, with overhead depending on the backbone.

\begin{figure}[!htbp]
  \centering
  \setlength{\parskip}{0pt}
  \includegraphics[width=\linewidth,trim=0 8bp 0 0,clip]{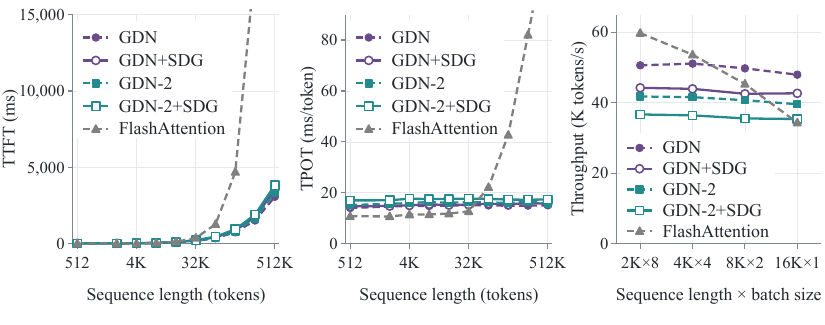}
  \par\vspace{-2pt}
  \setlength{\abovecaptionskip}{2pt}
  \caption{End-to-end efficiency on a single NVIDIA H200. Left and middle: hybrid-model TTFT and TPOT at batch size one. Right: recurrent-only training throughput, with 16,384 input tokens per optimizer step. FlashAttention uses full attention in both comparisons.}
  \label{fig:inference-latency}
\end{figure}

\endgroup
\FloatBarrier
\section{Conclusion and Limitation}
\enlargethispage{\baselineskip}
\label{sec:limitations}
\label{sec:conclusion}

We introduced State-Dependent Gating (\sdg), which derives element-wise boundary decay from state changes while preserving native within-chunk computation. Fused implementations support the evaluated backbones. With the realized gate fixed, boundary decay is an element-wise ridge proximal step. Reconstruction analysis quantifies potential gains for ideal gate families under a signal--noise model. In 1.3B models trained on 100B tokens, \sdg improves WikiText perplexity and average real-world retrieval across four recurrent backbones. Average retrieval gains extend to three sliding-window hybrids, while zero-shot accuracy and broader long-context results remain mixed. 

The gate uses only the net state change and cannot distinguish write--erase histories with the same resulting difference. We evaluate only a 64-token boundary interval, so the effect of other intervals on memory control and parallel efficiency remains untested. Runtime overhead varies across backbones, and cached decoding requires an additional source-state matrix.

\clearpage
\begingroup
\setlength{\bibsep}{2.5pt plus .3pt minus .3pt}
\interlinepenalty=10000
\bibliography{sdg}
\bibliographystyle{icml2026}
\endgroup

\clearpage
\appendix
\raggedbottom

\section{Notation}
\label{app:notation}

\begin{table}[H]
  \setlength{\belowcaptionskip}{6pt}
  \caption{Symbols used throughout the paper. Shapes are per head unless stated otherwise.}
  \label{tab:notation}
  \centering
  \small
  \renewcommand{\arraystretch}{1.12}
  \begin{tabular}{lll}
    \toprule
    Symbol & Shape or type & Meaning \\
    \midrule
    $t,n$ & indices & token and chunk indices \\
    $H$ & scalar & number of recurrent heads \\
    $d_k,d_v$ & scalars & key/query and value/output dimensions \\
    $d_s$ & scalar & Mamba-3 SSM state width per head \\
    $C$ & scalar & chunk size \\
    $\mathbf Q_{[n]},\mathbf K_{[n]},\mathbf V_{[n]}$ & chunk tensors
      & query, key, and value blocks of chunk $n$ \\
    $\mathbf S_{[n]}^{+}$ & $d_v\times d_k$ & state entering chunk $n$ \\
    $\mathbf S_{[n+1]}^{-}$ & $d_v\times d_k$ & raw state after chunk $n$ \\
    $\mathbf S_{[n+1]}^{+}$ & $d_v\times d_k$ & state after \sdg \\
    $\Delta\mathbf S_{[n]}$ & $d_v\times d_k$ & state change over chunk $n$ \\
    $\mathbf W_S$ & $d_k\times d_k$ & learned key-channel projection \\
    $\mathbf X_{[n]}$ & $d_v\times d_k$ & projected state change \\
    $\mathbf G$ & $d_v\times d_k$ & learned element-wise gain \\
    $\mathbf Z_{[n]}$ & $d_v\times d_k$ & signed decay score \\
    $\rho_0$ & scalar & fixed positive base rate \\
    $\boldsymbol\rho_{[n]}$ & $d_v\times d_k$ & non-negative boundary rate \\
    $\boldsymbol\Lambda_{[n]}$ & $d_v\times d_k$
        & element-wise decay gate; $(0,1]$ in Eq.~(\ref{eq:boundary-update}), $(0,1)$ under Eq.~(\ref{eq:canonical-sdg}) \\
    $\mathcal F$ & operator & original layer's chunk recurrence \\
       $\mathcal G_\theta$ & operator & \sdg gate function with learned parameters $\theta$ \\
    \midrule
    \multicolumn{3}{@{}l}{\textit{Analysis (Section~\ref{sec:analysis})}} \\
    $\boldsymbol\kappa$ & $d_v\times d_k$
      & ridge penalty weights, $\boldsymbol\kappa=e^{\boldsymbol\rho}-\mathbf 1$ \\
    $\mathbf S^\star,\mathbf N$ & $d_v\times d_k$
      & target state and residual interference, $\mathbf S^-=\mathbf S^\star+\mathbf N$ \\
    $\Phi$ & random element
      & oracle conditioning information, e.g.\ $(\mathbf S_{[n]}^+,\mathbf S_{[n+1]}^-)$ \\
    $\mathbf m_\star,\mathbf m_N,\mathbf m_-$ & $d_v\times d_k$
      & element-wise second moments given $\Phi$ (Lemma~\ref{lem:conditional-wiener}) \\
    $\bar{\mathbf m}$ & $d_v\times d_k$
      & unconditional second moment $\mathbb E[\mathbf m_-]$ \\
    $\mathcal R$ & functional
      & reconstruction risk $\mathbb E\|\boldsymbol\Lambda\odot\mathbf S^--\mathbf S^\star\|_F^2$ \\
    $\boldsymbol\Lambda^\star,\bar{\boldsymbol\Lambda}$ & $d_v\times d_k$
      & optimal $\Phi$-measurable gate and optimal constant gate \\
    $\mathcal E_{\mathrm{state}},\mathcal E_{\mathrm{value}},\mathcal E_{\mathrm{key}}$ & scalars
      & excess risks in Eq.~(\ref{eq:sharing-cost}) \\
    \bottomrule
  \end{tabular}
\end{table}

\section{Gate parameters and initialization}
\label{app:gate-parameters}

The gate function in Section~\ref{sec:gate-function} learns a projection $\mathbf W_S$ and an element-wise gain $\mathbf G$ for each head. This adds
\begin{equation}
  H(d_k^2+d_vd_k)
  \label{eq:parameter-count}
\end{equation}
parameters to each \sdg-enabled linear attention layer. We initialize
\begin{equation}
  \mathbf G=\mathbf 0,
  \qquad
  \rho_0=-\log\lambda_0,
  \qquad
  \lambda_0=0.9999.
  \label{eq:initialization}
\end{equation}
This gives $\boldsymbol\Lambda_{[n]}=\lambda_0\mathbf 1$ at every boundary before training, independently of the state and projection. Each head's $\mathbf W_S$ uses Xavier uniform initialization with bounds $\pm\sqrt{3/d_k}$. Both $\mathbf W_S$ and $\mathbf G$ are stored in FP32. They are trainable, while $\rho_0$ is fixed throughout training. At initialization, the zero gain makes the gradient of $\mathbf W_S$ zero. The projection begins to learn after $\mathbf G$ changes.

\section{Experimental settings}
\label{app:training-details}

\subsection{Models}

Table~\ref{tab:model-parameters} reports parameter counts. The counts include normalization parameters and all additional \sdg parameters. The models have 18 layers and hidden width 2304. GDN, GDN-2, and KDA use head dimension 128. The Mamba-3 experiments use the SISO variant. Hybrid models replace every other recurrent layer with sliding-window attention and use a window of 2048 tokens.

\begin{table}[h!]
  \setlength{\belowcaptionskip}{6pt}
  \caption{Non-embedding parameter counts for recurrent models, excluding the input embedding and language-model head. SDG is added without reducing the width of the corresponding base model.}
  \label{tab:model-parameters}
  \centering
  \small
  \renewcommand{\arraystretch}{1.12}
  \begin{tabular}{lrrr}
    \toprule
    Model & Base & With \sdg & Added \\
    \midrule
    GDN & 1,302,419,520 & 1,311,856,704 & 9,437,184 \\
    KDA & 1,300,649,760 & 1,310,086,944 & 9,437,184 \\
    Mamba-3 & 1,303,308,576 & 1,335,159,072 & 31,850,496 \\
    GDN-2 & 1,302,640,416 & 1,312,077,600 & 9,437,184 \\
    \bottomrule
  \end{tabular}
\end{table}

\subsection{Training}

The models are trained for 100B tokens using sequences of 4096 tokens and a global batch of 512 sequences. We train with AdamW, $\beta_1=0.9$, $\beta_2=0.95$, weight decay 0.1, and gradient clipping at 1.0. The peak learning rate is $4\times10^{-4}$. The learning rate follows a linear warm-up and cosine decay, with a warm-up over 1B tokens and a final rate of one tenth of the peak rate. Training uses bfloat16.

Each baseline and its +\sdg counterpart use the same data order and optimizer settings. The initial boundary decay factor is $\lambda_0=0.9999$, corresponding to the base rate $\rho_0=-\log\lambda_0$ in Eq.~(\ref{eq:initialization}).

\subsection{Evaluation}

We evaluate language modeling and zero-shot reasoning using the Language Model Evaluation Harness~\citep{eval-harness}. Language modeling performance is measured by perplexity on WikiText~\citep{merity2017pointer} and LAMBADA~\citep{paperno2016lambada}. The zero-shot suite comprises LAMBADA, PIQA~\citep{bisk2020piqa}, HellaSwag~\citep{zellers2019hellaswag}, WinoGrande~\citep{sakaguchi2020winogrande}, ARC-Easy, ARC-Challenge~\citep{clark2018arc}, OpenBookQA~\citep{mihaylov2018openbookqa}, SocialIQA~\citep{sap2019socialiqa}, and BoolQ~\citep{clark2019boolq}. We report accuracy on these tasks, with length normalization for HellaSwag and ARC-Challenge.

For synthetic retrieval, we evaluate S-NIAH-1 and S-NIAH-2 at 1K, 2K, 4K, and 8K tokens, and S-NIAH-3 and MK-NIAH-1 at 1K, 2K, and 4K tokens. Real-world retrieval is evaluated on SWDE~\citep{hao2011swde,lockard2019openceres}, FDA~\citep{arora2023datalakes}, SQuAD~\citep{rajpurkar2018squad}, DROP~\citep{dua2019drop}, Natural Questions~\citep{kwiatkowski2019naturalquestions}, and TriviaQA~\citep{joshi2017triviaqa}. We assess long-context understanding on all 21 LongBench~\citep{bai2023longbench} tasks, reporting category means and the unweighted average over tasks in Table~\ref{tab:longbench-results}.

\subsection{Efficiency protocol}
\label{app:efficiency}

\paragraph{Training throughput.}
We measure end-to-end training throughput for the recurrent-only models on a single NVIDIA H200. Each step processes 16,384 input tokens, using sequence-length $\times$ batch-size configurations of 2K$\times$8, 4K$\times$4, 8K$\times$2, and 16K$\times$1. We report median throughput after warm-up in thousands of input tokens per second (K tokens/s). Each timed step includes gradient reset, the forward pass and loss computation, backpropagation, gradient clipping, and the AdamW update. The optimizer updates FP32 master parameters and copies the updated weights to the model. Models receive identical inputs. We use neither distributed training nor gradient accumulation.

\paragraph{Inference latency.}
We measure end-to-end inference latency on a single NVIDIA H200 with batch size one. After warm-up, we run five TTFT trials and three TPOT trials. TTFT is the synchronized elapsed time for prompt processing and first-token prediction. TPOT is the average time per token over 256 subsequent cached greedy decoding steps. We report the median across trials for each metric.

\FloatBarrier
\subsection{Detailed efficiency results}
\label{app:efficiency-tables}

Tables~\ref{tab:efficiency-ttft} and~\ref{tab:efficiency-tpot} give the complete median inference latencies for the four backbones and the FlashAttention reference, including the GDN and GDN-2 curves in Figure~\ref{fig:inference-latency}. These inference runs use hybrid models with alternating recurrent and sliding-window attention layers (window size 2048), whereas the training runs in Table~\ref{tab:efficiency-training} use recurrent-only models. FlashAttention uses full attention in both comparisons. All runs use a single NVIDIA H200. Sequence-length labels use $1\mathrm{K}=1024$ tokens.


\begin{table}[!htbp]
  \setlength{\belowcaptionskip}{6pt}
  \caption{Time to first token (TTFT, ms). Values are medians over five trials at batch size one. Base and +\sdg denote the original and gated hybrid models, respectively. FlashAttn. is the full-attention reference. The - indicates OOM.}
  \label{tab:efficiency-ttft}
  \centering
  \footnotesize
  \setlength{\tabcolsep}{3pt}
  \renewcommand{\arraystretch}{1.16}
  \begin{tabular*}{\linewidth}{@{\extracolsep{\fill}}l*{9}{r}@{}}
    \toprule
    Length & \multicolumn{2}{c}{GDN} & \multicolumn{2}{c}{KDA} & \multicolumn{2}{c}{Mamba-3} & \multicolumn{2}{c}{GDN-2} & FlashAttn. \\
    \cmidrule(lr){2-3}\cmidrule(lr){4-5}\cmidrule(lr){6-7}\cmidrule(lr){8-9}
     & Base & +\sdg & Base & +\sdg & Base & +\sdg & Base & +\sdg &  \\
    \midrule
    512 & 22.53 & 24.38 & 23.84 & 24.82 & 16.77 & 17.88 & 24.19 & 26.13 & 11.76 \\
    2K & 23.66 & 24.41 & 24.00 & 25.57 & 18.98 & 31.44 & 24.79 & 25.97 & 12.13 \\
    4K & 25.46 & 29.01 & 27.04 & 31.05 & 34.17 & 58.66 & 27.60 & 31.66 & 22.68 \\
    8K & 46.88 & 53.40 & 50.24 & 59.47 & 65.52 & 114.29 & 51.14 & 59.09 & 50.21 \\
    16K & 91.60 & 105.59 & 97.46 & 114.30 & 130.93 & 227.73 & 100.32 & 117.67 & 131.11 \\
    32K & 184.57 & 210.21 & 196.31 & 227.36 & 259.83 & 452.57 & 200.47 & 230.83 & 380.85 \\
    64K & 367.49 & 424.45 & 398.35 & 466.09 & 534.25 & 920.80 & 400.97 & 466.97 & 1275.78 \\
    128K & 755.20 & 868.70 & 806.35 & 948.42 & 1086.01 & 1888.16 & 810.34 & 945.03 & 4694.28 \\
    256K & 1523.78 & 1750.00 & 1658.86 & 1909.11 & \textemdash & \textemdash & 1664.77 & 1912.12 & 18335.50 \\
    512K & 3075.18 & 3529.47 & 3345.86 & 3841.77 & \textemdash & \textemdash & 3356.33 & 3829.97 & \textemdash \\
    \bottomrule
  \end{tabular*}
\end{table}

\begin{table}[!htbp]
  \setlength{\belowcaptionskip}{6pt}
  \caption{Time per output token (TPOT, ms/token). Each trial averages 256 cached decoding steps. Entries are medians over three trials at batch size one. Model configurations follow Table~\ref{tab:efficiency-ttft}.}
  \label{tab:efficiency-tpot}
  \centering
  \footnotesize
  \setlength{\tabcolsep}{3pt}
  \renewcommand{\arraystretch}{1.16}
  \begin{tabular*}{\linewidth}{@{\extracolsep{\fill}}l*{9}{r}@{}}
    \toprule
    Length & \multicolumn{2}{c}{GDN} & \multicolumn{2}{c}{KDA} & \multicolumn{2}{c}{Mamba-3} & \multicolumn{2}{c}{GDN-2} & FlashAttn. \\
    \cmidrule(lr){2-3}\cmidrule(lr){4-5}\cmidrule(lr){6-7}\cmidrule(lr){8-9}
     & Base & +\sdg & Base & +\sdg & Base & +\sdg & Base & +\sdg &  \\
    \midrule
    512 & 14.03 & 14.65 & 14.53 & 15.91 & 17.99 & 15.94 & 15.19 & 17.03 & 10.78 \\
    2K & 14.63 & 14.84 & 14.62 & 16.28 & 18.49 & 16.22 & 15.68 & 17.05 & 10.74 \\
    4K & 14.84 & 15.07 & 15.31 & 16.43 & 18.76 & 16.40 & 15.97 & 17.72 & 11.44 \\
    8K & 14.83 & 15.47 & 15.20 & 16.60 & 18.80 & 17.03 & 16.38 & 17.54 & 11.50 \\
    16K & 14.97 & 15.21 & 14.84 & 16.78 & 18.67 & 16.42 & 16.07 & 17.65 & 11.86 \\
    32K & 15.18 & 15.31 & 15.01 & 16.58 & 18.91 & 16.57 & 16.32 & 17.69 & 12.71 \\
    64K & 15.06 & 15.77 & 15.47 & 16.77 & 18.80 & 16.40 & 16.30 & 17.61 & 22.18 \\
    128K & 14.94 & 15.36 & 15.07 & 16.59 & 19.15 & 16.70 & 15.95 & 17.35 & 42.71 \\
    256K & 14.98 & 16.11 & 15.56 & 16.89 & \textemdash & \textemdash & 16.65 & 17.30 & 82.00 \\
    512K & 15.09 & 15.62 & 14.87 & 16.99 & \textemdash & \textemdash & 15.72 & 17.37 & \textemdash \\
    \bottomrule
  \end{tabular*}
\end{table}

\begin{table}[!htbp]
  \setlength{\belowcaptionskip}{6pt}
  \caption{End-to-end training throughput in thousands of input tokens/s (higher is better). Each column specifies sequence length $\times$ microbatch size and processes 16,384 input tokens per optimizer step. Parentheses give +\sdg throughput as a percentage of its corresponding base model. These are the values used for the training curves in Figure~\ref{fig:inference-latency}, with KDA and Mamba-3 included for completeness.}
  \label{tab:efficiency-training}
  \centering
  \small
  \setlength{\tabcolsep}{4pt}
  \renewcommand{\arraystretch}{1.18}
  \begin{tabular}{p{\dimexpr0.24\linewidth-2\tabcolsep\relax}*{4}{>{\raggedleft\arraybackslash}p{\dimexpr0.19\linewidth-2\tabcolsep\relax}}}
    \toprule
    Model & 2K$\times$8 & 4K$\times$4 & 8K$\times$2 & 16K$\times$1 \\
    \midrule
    GDN & 50.609 & 51.089 & 49.751 & 47.951 \\
    \rowcolor{SDGLightPurple}GDN+\sdg & 44.233 (87.4\%) & 43.936 (86.0\%) & 42.537 (85.5\%) & 42.677 (89.0\%) \\
    \addlinespace[3pt]
    KDA & 42.403 & 42.536 & 41.797 & 40.571 \\
    \rowcolor{SDGLightPurple}KDA+\sdg & 36.721 (86.6\%) & 36.155 (85.0\%) & 35.820 (85.7\%) & 36.148 (89.1\%) \\
    \addlinespace[3pt]
    Mamba-3 & 29.496 & 28.153 & 25.600 & 21.976 \\
    \rowcolor{SDGLightPurple}Mamba-3+\sdg & 21.650 (73.4\%) & 20.664 (73.4\%) & 19.226 (75.1\%) & 15.932 (72.5\%) \\
    \addlinespace[3pt]
    GDN-2 & 41.768 & 41.576 & 40.707 & 39.579 \\
    \rowcolor{SDGLightPurple}GDN-2+\sdg & 36.672 (87.8\%) & 36.421 (87.6\%) & 35.537 (87.3\%) & 35.384 (89.4\%) \\
    \midrule
    FlashAttention & 59.760 & 53.617 & 45.328 & 34.343 \\
    \bottomrule
  \end{tabular}
\end{table}
\FloatBarrier

\section{Additional experimental results}
\label{app:full-experiments}

\subsection{LongBench results for recurrent models}
\label{app:longbench-results}

On LongBench~\citep{bai2023longbench}, \sdg raises the Code category score by 1.81--2.62 points across all four backbones (Table~\ref{tab:longbench-results}). Average scores improve for GDN, KDA, and Mamba-3, but fall from 11.65 to 11.15 for GDN-2. Its few-shot score falls from 26.94 to 20.85, outweighing gains in code and question answering. Retrieval improvements thus do not translate uniformly to broader long-context performance.

\begin{table*}[!htb]
  \setlength{\belowcaptionskip}{6pt}
  \caption{LongBench scores. Columns give category means. Avg. is the unweighted mean over all 21 tasks.}
  \label{tab:longbench-results}
  \centering
  \footnotesize
  \setlength{\tabcolsep}{2pt}
  \renewcommand{\arraystretch}{1.14}
  \settowidth{\modelcolwidth}{\textbf{Mamba-3+SDG}}
  \addtolength{\modelcolwidth}{2pt}
  \settowidth{\qacolwidth}{Single-doc QA}
  \addtolength{\qacolwidth}{2pt}
  \setlength{\metriccolwidth}{\dimexpr(\linewidth-\modelcolwidth-2\qacolwidth-16\tabcolsep)/5\relax}
  \begin{tabular}{p{\modelcolwidth}*{2}{C{\metriccolwidth}}*{2}{C{\qacolwidth}}*{3}{C{\metriccolwidth}}}
    \toprule
    Model & Code & \mbox{Few-shot} & \mbox{Multi-doc QA}
      & \mbox{Single-doc QA} & Summ. & Synthetic & Avg. \\
    \midrule
    GDN & 23.19 & 17.20 & \textbf{6.22} & 6.89 & 10.39 & 2.86 & 10.37 \\
    \rowcolor{SDGLightPurple}\textbf{GDN+\sdg} & \textbf{25.00} & \textbf{19.10} & 5.62 & \textbf{7.24} & \textbf{10.86} & \textbf{2.87} & \textbf{10.95} \\
    \modelsep
    KDA & 20.05 & \textbf{23.83} & 5.27 & 5.60 & 9.74 & \textbf{1.88} & 10.64 \\
    \rowcolor{SDGLightPurple}\textbf{KDA+\sdg} & \textbf{22.59} & 23.10 & \textbf{6.03} & \textbf{6.37} & \textbf{10.34} & 1.55 & \textbf{11.10} \\
    \modelsep
    Mamba-3 & 20.37 & 18.69 & \textbf{6.46} & \textbf{7.06} & 8.62 & \textbf{2.18} & 10.03 \\
    \rowcolor{SDGLightPurple}\textbf{Mamba-3+\sdg} & \textbf{22.22} & \textbf{20.08} & 5.94 & 6.82 & \textbf{11.63} & 1.28 & \textbf{10.77} \\
    \modelsep
    GDN-2 & 22.38 & \textbf{26.94} & 5.19 & 6.65 & \textbf{10.09} & 1.46 & \textbf{11.65} \\
    \rowcolor{SDGLightPurple}\textbf{GDN-2+\sdg} & \textbf{25.00} & 20.85 & \textbf{6.41} & \textbf{7.05} & 9.83 & \textbf{2.53} & 11.15 \\
    \bottomrule
  \end{tabular}
\end{table*}
\FloatBarrier

\subsection{Hybrid models}
\label{app:hybrid-results}

Tables~\ref{tab:lm-reasoning-hybrid}--\ref{tab:longbench-results-hybrid} report results for hybrid models that alternate recurrent layers with sliding-window attention. The evaluation follows the recurrent-model comparisons in Section~\ref{sec:experiments}. Bold marks the better result within each baseline--\sdg pair, including ties.

\paragraph{Language modeling and reasoning.}
Table~\ref{tab:lm-reasoning-hybrid} shows that \sdg improves LAMBADA perplexity across all three hybrid models. GDN also improves on WikiText and average commonsense reasoning accuracy, while KDA and GDN-2 show small declines on these measures. The language modeling and reasoning benefits are thus less consistent than in the recurrent-only models.

\begin{table*}[!htbp]
  \setlength{\belowcaptionskip}{6pt}
  \caption{Language modeling and zero-shot reasoning for hybrid models. Metrics follow Table~\ref{tab:lm-reasoning}. Avg. is the mean over the nine accuracy columns.}
  \label{tab:lm-reasoning-hybrid}
  \centering
  \footnotesize
  \setlength{\tabcolsep}{3pt}
  \renewcommand{\arraystretch}{1.12}
  \settowidth{\modelcolwidth}{\textbf{Mamba-3+SDG}}
  \addtolength{\modelcolwidth}{2pt}
  \setlength{\metriccolwidth}{\dimexpr(\linewidth-\modelcolwidth-2\tabcolsep-\arrayrulewidth-6pt)/12-2\tabcolsep\relax}
  \begin{tabular}{p{\modelcolwidth}*{2}{C{\metriccolwidth}}!{\vrule width \arrayrulewidth}*{4}{C{\metriccolwidth}}*{2}{C{\dimexpr\metriccolwidth+3pt\relax}}*{4}{C{\metriccolwidth}}}
    \toprule
    Model & \makebox[\linewidth][c]{Wiki.} & \makebox[\linewidth][c]{LMB.} & \makebox[\linewidth][c]{LMB.} & \makebox[\linewidth][c]{PIQA} & \makebox[\linewidth][c]{Hella.} & \makebox[\linewidth][c]{Wino.} & \makebox[\linewidth][c]{ARC-e} & \makebox[\linewidth][c]{ARC-c} & \makebox[\linewidth][c]{OBQA} & \makebox[\linewidth][c]{SIQA} & \makebox[\linewidth][c]{BoolQ} & \makebox[\linewidth][c]{Avg.} \\
    & ppl $\downarrow$ & ppl $\downarrow$ & acc $\uparrow$ & acc $\uparrow$ & \makebox[\linewidth][c]{acc$_n$ $\uparrow$} & acc $\uparrow$
      & acc $\uparrow$ & \makebox[\linewidth][c]{acc$_n$ $\uparrow$} & acc $\uparrow$ & acc $\uparrow$ & acc $\uparrow$ & acc $\uparrow$ \\
    \midrule
    GDN & 16.01 & 11.70 & 48.03 & 72.25 & 43.62 & \textbf{57.77} & 70.24 & \textbf{35.24} & 30.40 & \textbf{40.89} & 59.63 & 50.90 \\
    \rowcolor{SDGLightPurple}\textbf{GDN+\sdg} & \textbf{15.59} & \textbf{11.34} & \textbf{49.04} & \textbf{73.29} & \textbf{44.42} & 56.91 & \textbf{72.14} & 34.39 & \textbf{31.00} & 40.35 & \textbf{63.85} & \textbf{51.71} \\
    \modelsep
    KDA & \textbf{15.65} & 11.27 & 48.92 & \textbf{72.85} & \textbf{44.72} & 56.83 & \textbf{71.68} & \textbf{37.29} & \textbf{32.00} & 41.20 & \textbf{62.60} & \textbf{52.01} \\
    \rowcolor{SDGLightPurple}\textbf{KDA+\sdg} & 15.66 & \textbf{11.05} & \textbf{49.76} & 72.69 & 44.48 & \textbf{57.70} & 71.59 & 35.58 & 31.60 & \textbf{41.56} & 62.51 & 51.94 \\
    \modelsep
    GDN-2 & \textbf{15.44} & 11.59 & 48.92 & \textbf{72.31} & \textbf{44.55} & \textbf{58.33} & 69.78 & \textbf{36.52} & \textbf{32.60} & 41.15 & 62.02 & \textbf{51.80} \\
    \rowcolor{SDGLightPurple}\textbf{GDN-2+\sdg} & 15.60 & \textbf{11.33} & \textbf{49.17} & 72.09 & 43.70 & 58.01 & \textbf{70.24} & 33.96 & 31.60 & \textbf{41.20} & \textbf{63.96} & 51.55 \\
    \bottomrule
  \end{tabular}
\end{table*}

\paragraph{Synthetic retrieval.}
Table~\ref{tab:niah-results-hybrid} shows that the effect of \sdg varies across tasks and input lengths. GDN-2 benefits most broadly, improving multi-key retrieval across all evaluated lengths and single-needle retrieval in several longer-context settings. GDN shows clear gains on S-NIAH-3, while KDA has a more mixed response. As in the recurrent-only models, improvements on synthetic retrieval do not imply uniformly better length extrapolation.

\begin{table*}[!htbp]
  \setlength{\belowcaptionskip}{6pt}
  \caption{RULER retrieval accuracy (\%) for hybrid models.}
  \label{tab:niah-results-hybrid}
  \centering
  \footnotesize
  \setlength{\tabcolsep}{1.5pt}
  \renewcommand{\arraystretch}{1.12}
  \settowidth{\modelcolwidth}{\textbf{Mamba-3+SDG}}
  \addtolength{\modelcolwidth}{2pt}
  \begin{tabular}{p{\modelcolwidth}*{14}{C{\dimexpr(\linewidth-\modelcolwidth-2\tabcolsep)/14-2\tabcolsep\relax}}}
    \toprule
    Model & \multicolumn{4}{c}{S-NIAH-1} & \multicolumn{4}{c}{S-NIAH-2}
      & \multicolumn{3}{c}{S-NIAH-3} & \multicolumn{3}{c}{MK-NIAH-1} \\
    \cmidrule(lr){2-5}\cmidrule(lr){6-9}\cmidrule(lr){10-12}\cmidrule(lr){13-15}
    & 1K & 2K & 4K & 8K & 1K & 2K & 4K & 8K & 1K & 2K & 4K & 1K & 2K & 4K \\
    \midrule
    GDN & \textbf{100.0} & 96.4 & \textbf{51.4} & \textbf{24.4} & \textbf{100.0} & \textbf{100.0} & \textbf{56.4} & \textbf{26.6} & 92.2 & 71.8 & 37.0 & 77.8 & 63.4 & \textbf{44.8} \\
    \rowcolor{SDGLightPurple}\textbf{GDN+\sdg} & \textbf{100.0} & \textbf{98.8} & 47.2 & 23.8 & \textbf{100.0} & 99.6 & 52.6 & 25.8 & \textbf{95.0} & \textbf{90.4} & \textbf{39.4} & \textbf{84.6} & \textbf{75.0} & 42.2 \\
    \modelsep
    KDA & \textbf{100.0} & \textbf{100.0} & \textbf{54.6} & \textbf{26.2} & \textbf{100.0} & \textbf{100.0} & 54.8 & 29.6 & \textbf{94.0} & 81.6 & 31.0 & 83.6 & \textbf{73.2} & 39.4 \\
    \rowcolor{SDGLightPurple}\textbf{KDA+\sdg} & \textbf{100.0} & 99.4 & 53.8 & \textbf{26.2} & \textbf{100.0} & \textbf{100.0} & \textbf{56.4} & \textbf{29.8} & 91.6 & \textbf{84.8} & \textbf{34.2} & \textbf{84.2} & 69.4 & \textbf{40.4} \\
    \modelsep
    GDN-2 & \textbf{100.0} & \textbf{100.0} & 54.2 & 22.0 & \textbf{100.0} & \textbf{100.0} & 44.6 & 20.8 & 98.2 & \textbf{97.8} & 36.4 & 82.0 & 78.4 & 38.6 \\
    \rowcolor{SDGLightPurple}\textbf{GDN-2+\sdg} & \textbf{100.0} & \textbf{100.0} & \textbf{57.8} & \textbf{26.2} & \textbf{100.0} & 99.8 & \textbf{55.8} & \textbf{28.0} & \textbf{99.4} & 92.0 & \textbf{41.6} & \textbf{86.6} & \textbf{81.4} & \textbf{42.0} \\
    \bottomrule
  \end{tabular}
\end{table*}

\paragraph{Real-world retrieval.}
Table~\ref{tab:recall-results-hybrid} shows higher average retrieval accuracy for all three hybrid models, extending the trend observed in the recurrent-only setting. GDN and KDA show clearer average gains, while GDN-2 changes little overall because improvements and declines across tasks largely offset one another. These results suggest that \sdg can complement sliding-window attention in hybrid models.

\begin{table*}[!htbp]
  \setlength{\belowcaptionskip}{6pt}
  \caption{Real-world retrieval accuracy (\%) for hybrid models. Avg. is the mean over the six tasks.}
  \label{tab:recall-results-hybrid}
  \centering
  \footnotesize
  \setlength{\tabcolsep}{4.2pt}
  \renewcommand{\arraystretch}{1.12}
  \settowidth{\modelcolwidth}{\textbf{Mamba-3+SDG}}
  \addtolength{\modelcolwidth}{2pt}
  \begin{tabular}{p{\modelcolwidth}*{7}{C{\dimexpr(\linewidth-\modelcolwidth-2\tabcolsep)/7-2\tabcolsep\relax}}}
    \toprule
    Model & SWDE & SQD & FDA & TQA & NQ & DROP & Avg. \\
    \midrule
    GDN & 46.39 & 41.82 & 59.85 & 61.20 & \textbf{25.66} & \textbf{23.24} & 43.03 \\
    \rowcolor{SDGLightPurple}\textbf{GDN+\sdg} & \textbf{48.17} & \textbf{44.71} & \textbf{60.13} & \textbf{62.44} & 25.28 & 23.10 & \textbf{43.97} \\
    \modelsep
    KDA & 43.96 & \textbf{43.37} & 58.86 & 62.03 & 26.39 & \textbf{24.68} & 43.22 \\
    \rowcolor{SDGLightPurple}\textbf{KDA+\sdg} & \textbf{45.08} & 43.23 & \textbf{61.94} & \textbf{64.22} & \textbf{27.02} & 22.95 & \textbf{44.07} \\
    \modelsep
    GDN-2 & \textbf{49.86} & \textbf{43.67} & 62.03 & \textbf{63.45} & 25.15 & 24.25 & 44.74 \\
    \rowcolor{SDGLightPurple}\textbf{GDN-2+\sdg} & 48.92 & 42.29 & \textbf{64.49} & 62.03 & \textbf{26.51} & \textbf{24.77} & \textbf{44.84} \\
    \bottomrule
  \end{tabular}
\end{table*}

\FloatBarrier
\paragraph{Long-context understanding.}
Table~\ref{tab:longbench-results-hybrid} shows consistent gains in code and few-shot learning across all three hybrid models. Average performance improves for KDA and GDN-2 but declines for GDN, whose gains are outweighed by weaker question answering and summarization. As in the recurrent-only setting, retrieval improvements do not translate uniformly to broader long-context understanding.

\begin{table*}[!htbp]
  \setlength{\belowcaptionskip}{6pt}
  \caption{LongBench scores (\%) for hybrid models. Columns give category means. Avg. is the unweighted mean over all 21 tasks. Higher is better.}
  \label{tab:longbench-results-hybrid}
  \centering
  \footnotesize
  \setlength{\tabcolsep}{4pt}
  \renewcommand{\arraystretch}{1.12}
  \settowidth{\modelcolwidth}{\textbf{Mamba-3+SDG}}
  \addtolength{\modelcolwidth}{2pt}
  \begin{tabular}{p{\modelcolwidth}*{7}{C{\dimexpr(\linewidth-\modelcolwidth-2\tabcolsep)/7-2\tabcolsep\relax}}}
    \toprule
    Model & Code & Few-shot & \shortstack{Multi-doc\\QA}
      & \makebox[\linewidth][c]{\shortstack{Single-doc\\QA}} & Summ. & Synthetic & Avg. \\
    \midrule
    GDN & 19.62 & 25.64 & \textbf{7.18} & \textbf{7.52} & \textbf{11.85} & \textbf{2.11} & \textbf{12.11} \\
    \rowcolor{SDGLightPurple}\textbf{GDN+\sdg} & \textbf{21.26} & \textbf{27.86} & 5.70 & 5.32 & 9.37 & 2.02 & 11.50 \\
    \modelsep
    KDA & 20.10 & 26.23 & \textbf{7.18} & \textbf{7.05} & 9.08 & 1.36 & 11.54 \\
    \rowcolor{SDGLightPurple}\textbf{KDA+\sdg} & \textbf{21.27} & \textbf{26.63} & 5.85 & 6.53 & \textbf{10.07} & \textbf{1.77} & \textbf{11.63} \\
    \modelsep
    GDN-2 & 21.39 & 25.27 & 6.85 & \textbf{7.63} & \textbf{9.99} & \textbf{1.95} & 11.79 \\
    \rowcolor{SDGLightPurple}\textbf{GDN-2+\sdg} & \textbf{28.43} & \textbf{25.74} & \textbf{6.87} & 6.74 & 9.47 & 1.65 & \textbf{12.24} \\
    \bottomrule
  \end{tabular}
\end{table*}
\FloatBarrier

\subsection{Ablation settings}
\label{app:ablation-details}

\paragraph{Gate inputs.}
The fixed boundary control uses $\boldsymbol\Lambda_{[n]}=0.9999\mathbf 1$. The learned static control sets $\mathbf Z_{[n]}=\mathbf G$ in Eq.~(\ref{eq:boundary-rate}), with zero-initialized trainable $\mathbf G$. These controls examine whether the boundary gate benefits from adapting to its input. The chunk-input control further tests whether directly using the layer's state changes provides additional value. It uses $\boldsymbol\Omega_{[n]}$ from the affine chunk transition in Eq.~(\ref{eq:affine-chunk-transition}) of Appendix~\ref{app:operator-view}. For fixed chunk token representations and native data-dependent gates, this is the state produced by the native chunk update from a zero incoming state. In GDN, it is obtained by setting $\mathbf S_{[n]}^+=\mathbf 0$ in Eq.~(\ref{eq:gdn-chunk-state}). It retains the chunk's writes and their subsequent within-chunk updates, while excluding the response of the incoming memory.

The control replaces $\Delta\mathbf S_{[n]}$ with $\boldsymbol\Omega_{[n]}$ in Eq.~(\ref{eq:projected-state-change}) and retains the rate and boundary update in Eqs.~(\ref{eq:boundary-rate})--(\ref{eq:canonical-sdg}). The main recurrence still starts from the actual $\mathbf S_{[n]}^+$, and the gate scales the entire updated state $\mathbf S_{[n+1]}^-$. Only the gate input changes. The independence of $\boldsymbol\Omega_{[n]}$ from the incoming state is within the current layer, since its token representations can contain context from earlier layers.

The state-input control replaces $\Delta\mathbf S_{[n]}$ with $\mathbf S_{[n]}^+$ in Eq.~(\ref{eq:projected-state-change}). It reads the memory entering the chunk and retains the same gate parameterization and boundary update, allowing a comparison between the incoming state and its change across the chunk.

\paragraph{Channel-wise parameterization.}
The channel-wise variant retains the per-head projection $\mathbf W_S\in\mathbb R^{d_k\times d_k}$ and gain $\mathbf G\in\mathbb R^{d_v\times d_k}$ of element-wise \sdg. It computes $\mathbf Z_{[n]}=\mathbf G\odot(\Delta\mathbf S_{[n]}\mathbf W_S^\top)$ as in Eq.~(\ref{eq:projected-state-change}) and averages these scores over value rows before applying the nonlinear rate map:
\begin{align}
  \bar z_{[n],b}
    &=\frac{1}{d_v}\sum_{a=1}^{d_v}Z_{[n],ab},\\
  \lambda_{[n],b}
    &=\exp\!\left(-\bar z_{[n],b}
      -\sqrt{\bar z_{[n],b}^{\,2}+\rho_0^2}\right).
  \label{eq:channel-wise-gate}
\end{align}
The resulting coefficient for each key channel is broadcast across all value rows, giving $\Lambda_{[n],ab}=\lambda_{[n],b}$ and $\boldsymbol\Lambda_{[n]}=\mathbf1_{d_v}\boldsymbol\lambda_{[n]}^\top$. This gate scales the entire updated state through Eq.~(\ref{eq:canonical-sdg}). The variant uses the same zero initialization for $\mathbf G$ and the same fixed $\rho_0=-\log\lambda_0$, so all retention coefficients initially equal $\lambda_0=0.9999$. It adds $H(d_k^2+d_vd_k)$ parameters per layer, matching element-wise \sdg.

\paragraph{Paired training.}
Both gate-input controls and \sdg use identical initial values for the backbone and $\mathbf W_S$, with $\mathbf G=\mathbf 0$ and $\lambda_0=0.9999$. They therefore start from the same model function. The comparison holds the data order, training token budget, optimizer, learning-rate schedule, chunk size, gated layers, and evaluation settings fixed. Each variant uses its own gate input throughout training and evaluation. A comparison at a reduced training budget requires training all compared variants at that budget.

\paragraph{Parameter and computation costs.}
All three dynamic inputs use $H(d_k^2+d_vd_k)$ additional parameters per layer and the same gate projection. Constructing $\boldsymbol\Omega_{[n]}$ explicitly requires a matrix multiplication costing approximately $2Cd_vd_k$ forward FLOPs per head and boundary, whereas constructing $\Delta\mathbf S_{[n]}$ requires a state subtraction. The state-input control uses the already available $\mathbf S_{[n]}^+$. Thus, matching the parameter count and gate function does not imply equal total computation. The identity-projection control retains the state-difference input and element-wise gain while removing the learned projection and its matrix multiplication. The gate-input comparison therefore matches parameter count while allowing different input-construction costs.

\paragraph{Scope of the comparisons.}
The $\boldsymbol\Omega$-input and incoming-state controls test the input choice with the gate parameterization fixed. The granularity comparison changes coefficient sharing while retaining the same parameter count. The identity-projection control reduces the parameter count by removing $\mathbf W_S$. The reported scores are point estimates without uncertainty estimates, so small differences between variants do not establish a statistically reliable advantage.

\paragraph{Gate-input runtime.}
Table~\ref{tab:gate-input-runtime} compares the complete GDN operator with $\Delta\mathbf S$- and $\boldsymbol\Omega$-input SDG. Timing includes query/key normalization, WY preparation, the fused state scan, and output kernels. Backward includes boundary recomputation and gradients for all inputs and gate parameters. The timings are measured independently, with backward-only runs reusing a fixed forward graph, so their medians need not be additive. At 16,384 tokens per batch, $\boldsymbol\Omega$ increases forward latency by 1.30--6.07\%, backward latency by 46.17--49.29\%, and combined latency by 35.89--39.17\%. At batch size one, the differences are more shape-dependent, with a small forward decrease at 512 tokens. These measurements favor $\Delta\mathbf S$ for training-time operator efficiency but do not measure full-model runtime.

\begin{table}[!htbp]
  \setlength{\belowcaptionskip}{6pt}
  \caption{Median GDN operator latency (ms) on an RTX 5090. $\Delta t$ is the percentage change for $\boldsymbol\Omega$ relative to $\Delta\mathbf S$. The upper block fixes $LB=16{,}384$ tokens per batch, and the lower block uses $B=1$.}
  \label{tab:gate-input-runtime}
  \centering
  \small
  \setlength{\tabcolsep}{2pt}
  \renewcommand{\arraystretch}{1.1}
  \begin{tabular*}{\linewidth}{@{\extracolsep{\fill}}l*{9}{r}@{}}
    \toprule
    & \multicolumn{3}{c}{Forward} & \multicolumn{3}{c}{Backward}
      & \multicolumn{3}{c}{Forward + backward} \\
    \cmidrule(lr){2-4}\cmidrule(lr){5-7}\cmidrule(l){8-10}
    $L\times B$ & $\Delta\mathbf S$ & $\boldsymbol\Omega$ & $\Delta t$ (\%)
      & $\Delta\mathbf S$ & $\boldsymbol\Omega$ & $\Delta t$ (\%)
      & $\Delta\mathbf S$ & $\boldsymbol\Omega$ & $\Delta t$ (\%) \\
    \midrule
    $2048\times8$  & 5.164 & 5.451 & $+5.56$ & 12.342 & 18.426 & $+49.29$ & 15.765 & 21.939 & $+39.17$ \\
    $4096\times4$  & 5.112 & 5.422 & $+6.07$ & 12.867 & 19.030 & $+47.90$ & 16.637 & 22.875 & $+37.50$ \\
    $8192\times2$  & 4.800 & 4.912 & $+2.34$ & 13.252 & 19.513 & $+47.25$ & 16.850 & 23.353 & $+38.60$ \\
    $16384\times1$ & 5.705 & 5.779 & $+1.30$ & 12.686 & 18.542 & $+46.17$ & 16.624 & 22.589 & $+35.89$ \\
    \midrule
    $512\times1$   & 0.570 & 0.539 & $-5.53$ & 2.013 & 2.298 & $+14.13$ & 2.180 & 2.193 & $+0.59$ \\
    $2048\times1$  & 0.850 & 0.891 & $+4.80$ & 1.635 & 2.061 & $+26.09$ & 2.571 & 2.946 & $+14.59$ \\
    $4096\times1$  & 1.378 & 1.443 & $+4.73$ & 3.321 & 4.398 & $+32.42$ & 4.309 & 5.623 & $+30.51$ \\
    $8192\times1$  & 3.026 & 3.057 & $+1.04$ & 6.455 & 9.567 & $+48.20$ & 8.800 & 11.757 & $+33.60$ \\
    \bottomrule
  \end{tabular*}
\end{table}
\FloatBarrier

\subsection{Boundary gate distributions}
\label{app:gate-distributions}

\begin{figure}[!htbp]
  \centering
  \includegraphics[width=0.9\linewidth]{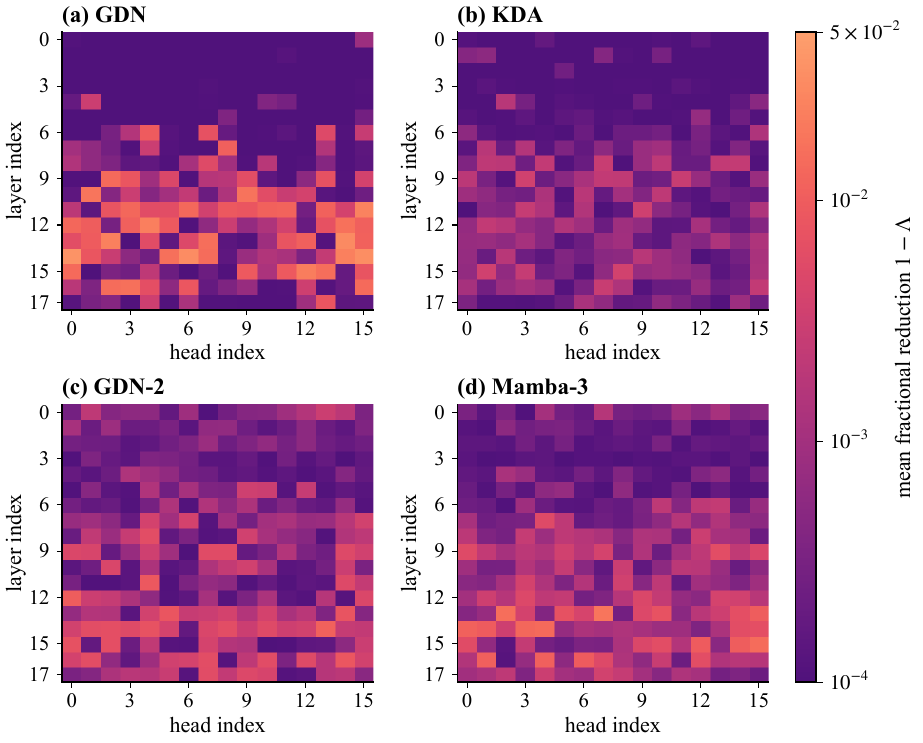}
  \setlength{\abovecaptionskip}{3pt}
  \caption{Mean fractional reduction $1-\Lambda$ across layers and heads on a shared logarithmic color scale. Mamba-3 statistics use the first 16 heads.}
  \label{fig:gate-distributions}
\end{figure}

We examine the distribution of the element-wise retention coefficient $\Lambda$ at observed chunk boundaries. We report the fractional reduction $d=1-\Lambda$, so larger values indicate stronger decay. Figure~\ref{fig:gate-distributions} shows the mean reduction across layers and heads. Table~\ref{tab:gate-distributions} reports statistics from the original model archives.

\paragraph{Observations and aggregation.}
Each model archive contains 18 layers, 32 sample IDs, and eight boundary positions per sample: 64, 640, 1216, 1792, 2368, 2944, 3520, and 4096 tokens consumed. This gives 256 boundary observations per layer. GDN, KDA, and GDN-2 each contribute 67,108,864 gate-cell observations per layer, across 16 heads. For Mamba-3, gate statistics are computed from the first 16 heads. Each measured heatmap cell averages the reduction over all observed samples, boundaries, and state elements for one layer and head. Means and threshold fractions use all observed cells; the reported medians use a uniform reservoir of 65,536 cells per layer. Repeated observations are not treated as independent experimental replicates. The reduction is computed stably as $-\operatorname{expm1}(-\rho)$, retaining small values even when FP32 retention rounds to one.

\paragraph{Distributions in the delta-rule models.}
For GDN, KDA, and GDN-2, every layer's sampled median reduction is approximately $10^{-4}$, corresponding to retention near the initialization value $0.9999$. The means are higher and vary across layers. Averaged over all observed cells, the reduction is 0.3104\%, 0.0538\%, and 0.1496\%, respectively (Table~\ref{tab:gate-distributions}). GDN and GDN-2 reach their largest layer means at layer 14 (1.0982\% and 0.4693\%); KDA peaks at layer 8 (0.1161\%). Only 0.69\%, 0.06\%, and 0.25\% of cells, respectively, have retention below 0.9. Thus, large reductions are concentrated in a small fraction of observations even though the median remains close to its initial value.

\begin{table}[!htbp]
  \setlength{\belowcaptionskip}{6pt}
  \caption{Measured gate statistics pooled across layers with equal weight per observed cell. All entries are percentages; threshold fractions use the archived FP32 comparisons.}
  \label{tab:gate-distributions}
  \centering
  \small
  \setlength{\tabcolsep}{10pt}
  \renewcommand{\arraystretch}{1.16}
  \begin{tabular}{@{}lrrr@{}}
    \toprule
    Model & Mean $1-\Lambda$ & $\Pr(\Lambda<0.99)$ & $\Pr(\Lambda<0.9)$ \\
    \midrule
    GDN+\sdg & 0.3104 & 4.2875 & 0.6915 \\
    KDA+\sdg & 0.0538 & 0.8487 & 0.0587 \\
    GDN-2+\sdg & 0.1496 & 2.5786 & 0.2541 \\
    Mamba-3 & 2.8041 & 99.6959 & $1.25\times10^{-6}$ \\
    \bottomrule
  \end{tabular}
\end{table}

\FloatBarrier

\subsection{Boundary output stability}
\label{app:boundary-stability}

\begin{figure}[!htbp]
  \centering
  \includegraphics[width=\linewidth]{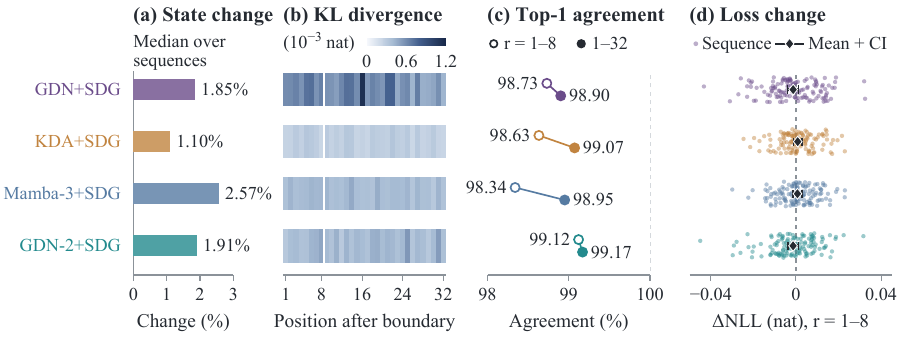}
  \setlength{\abovecaptionskip}{3pt}
  \caption{Effects of one SDG boundary update under identical inputs. \textbf{(a)} Median relative state change. \textbf{(b)} KL divergence from skip to normal. \textbf{(c)} Top-1 agreement. \textbf{(d)} Sequence-level mean $\Delta$NLL over positions 1--8. Black diamonds and bars denote means and four-model-adjusted bootstrap intervals.}
  \label{fig:boundary-stability-full}
\end{figure}

\paragraph{Paired boundary intervention.}
We compare normal execution with skipping one \sdg update in the same trained model. At the selected boundary, we clone the complete cache before applying the gate. The normal branch applies \sdg in all recurrent layers, while the skip branch passes the unmodified pre-gate states to the next chunk. All other operations are unchanged. Both branches then consume the same subsequent ground-truth tokens through cached decoding. This comparison isolates the effect of the selected update without changing the input or comparing separately trained models.

\paragraph{Prediction positions and metrics.}
Let $b$ be the input position after which the selected boundary update occurs, and let $r$ count subsequent input positions. The logits at position $b+r$ predict the ground-truth token $y_{b+r+1}$. The output at $r=0$ is computed before the boundary update and is therefore unaffected. We exclude this position from all window averages. Top-1 agreement is the fraction of affected positions at which both branches assign the highest probability to the same token. We report this fraction over $r=1,\ldots,8$ and $r=1,\ldots,32$, with equal weight for every sequence and position. At each position, we also measure KL divergence from the skip distribution to the normal distribution. This measures a change in the full output distribution rather than only its most probable token. All 32 positions precede the next boundary update.

\paragraph{Prediction loss.}
Agreement alone does not measure prediction quality, since both branches can predict the same incorrect token. We therefore also measure the negative log-likelihood (NLL) of the ground-truth next token. Given the common token prefix $y_{\leq b+r}$, it is
\begin{equation*}
  \mathrm{NLL}(r)
  =-\log p\!\left(y_{b+r+1}\mid y_{\leq b+r}\right),
\end{equation*}
where $p$ is the predictive distribution of the branch under consideration. We use the natural logarithm, so losses are measured in nats. Lower NLL means that the model assigns greater probability to the ground-truth token. We define the loss change as
\begin{equation}
  \Delta\mathrm{NLL}(r)
  =\mathrm{NLL}_{\mathrm{normal}}(r)
   -\mathrm{NLL}_{\mathrm{skip}}(r).
  \label{eq:boundary-loss-change}
\end{equation}
Positive values indicate that applying the update increases the target loss, while negative values indicate a decrease. We average $r=1,\ldots,8$ within each sequence before averaging across sequences, and inspect the loss changes at all 32 affected positions.

\paragraph{Samples and controls.}
The diagnostic includes 128 observations per model. GDN, KDA, and GDN-2 use the same 128 evaluation sequences at training step 47,683, with one selected boundary per sequence. Boundaries at input positions 1024, 2048, and 3072 account for 43, 43, and 42 sequences, respectively. Each intervention covers all 18 recurrent layers, with 16 heads per layer. For these three models, two unmodified executions produce identical logits, and the logit difference at $r=0$ is zero. Each layer receives exactly one boundary intervention. Mamba-3 contributes position-wise and sequence-level summaries; cross-model sample matching and intervention controls are assessed for GDN, KDA, and GDN-2.

\paragraph{Prediction agreement.}
Figure~\ref{fig:boundary-stability-full} shows that the two branches predict the same token at 98.34--99.12\% of the first eight affected positions. Over 32 positions, agreement is 98.90\%, 99.07\%, 98.95\%, and 99.17\% for GDN, KDA, Mamba-3, and GDN-2, respectively. These are position-level rates, not the fraction of entire sequences with identical predictions. For GDN, KDA, and GDN-2, all 32 top-1 predictions agree in 70.31\%, 71.88\%, and 75.00\% of the evaluation sequences, respectively. The remaining sequences contain at least one changed prediction. The small KL values complement the agreement rates by showing that the average change in the full distribution is also small.

\begin{figure}[!htbp]
  \centering
  \includegraphics[width=\linewidth]{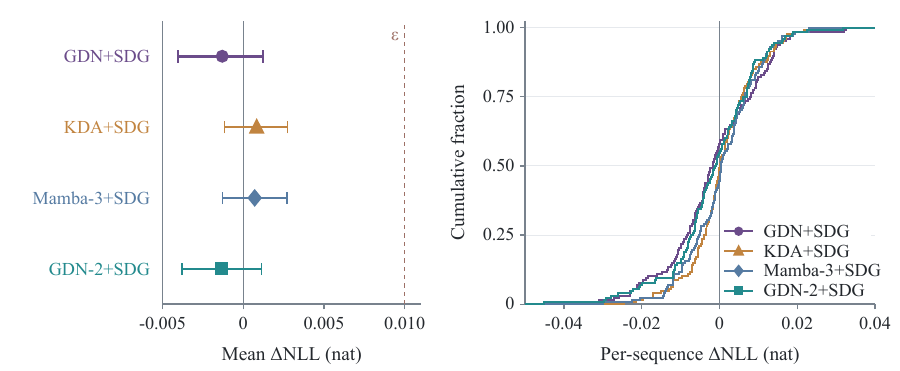}
  \setlength{\abovecaptionskip}{3pt}
  \caption{Prediction loss after a single SDG boundary update relative to skipping that update. The left panel reports mean $\Delta$NLL over the first eight affected positions with bootstrap intervals over sequences adjusted across four models for approximately 95\% simultaneous coverage. The dotted line marks the reference tolerance $\epsilon=0.01$ nat per token. The right panel shows the empirical distribution of sequence-level mean changes. Positive values indicate increased loss.}
  \label{fig:boundary-effects}
\end{figure}

\begin{figure}[!htbp]
  \centering
  \includegraphics[width=\linewidth]{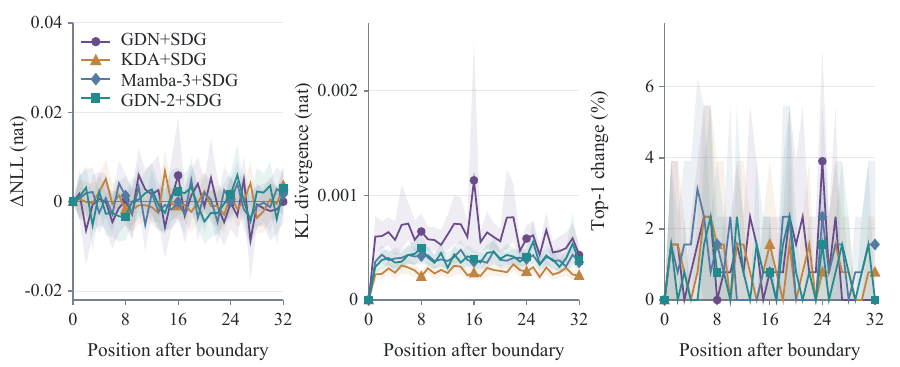}
  \setlength{\abovecaptionskip}{3pt}
  \caption{Prediction changes following a single SDG boundary update relative to skipping it. The panels show mean $\Delta$NLL, KL divergence from the skip distribution to the normal distribution, and the fraction of changed top-1 predictions. Shaded regions are pointwise 95\% bootstrap intervals over sequences. Position zero is unaffected because its output precedes the update.}
  \label{fig:boundary-response}
\end{figure}

\paragraph{Uncertainty estimates.}
We resample whole sequences 5,000 times. Percentile-bootstrap intervals for mean loss changes use Bonferroni correction across four models, with quantiles 0.00625 and 0.99375, targeting approximately 95\% simultaneous coverage. The $\epsilon=0.01$ line in Figure~\ref{fig:boundary-effects} is a descriptive reference for the population mean loss increase, not a per-sequence or per-token bound; our conclusions use the measured confidence bounds. Figure~\ref{fig:boundary-response} shows pointwise 95\% intervals, while the KL heatmap in Figure~\ref{fig:boundary-stability} shows the corresponding means. Window-level agreement percentages are descriptive averages without confidence intervals.

\paragraph{Average loss changes.}
Table~\ref{tab:boundary-summary} summarizes loss and output changes over the first eight affected positions. All four corrected intervals contain zero, and their upper bounds are below 0.003 nat per token. The small upper bounds constrain the average loss increase under the sampling protocol. The average KL divergence is below 0.00065 nat, while top-1 predictions differ at 0.88--1.66\% of the first eight affected positions. Figure~\ref{fig:boundary-response} retains the full position-wise variation rather than smoothing the curves. 

\begin{table}[!htbp]
  \setlength{\belowcaptionskip}{6pt}
  \caption{Prediction changes over the first eight affected positions. Loss and KL values are in nats. Confidence intervals are adjusted across four models.}
  \label{tab:boundary-summary}
  \centering
  \small
  \setlength{\tabcolsep}{5pt}
  \renewcommand{\arraystretch}{1.16}
  \begin{tabular}{@{}lrc rr@{}}
    \toprule
    Model & Mean $\Delta$NLL & Adjusted interval & Mean KL & Top-1 change \\
    \midrule
    GDN+\sdg & $-0.001301$ & $[-0.004024,\ 0.001216]$ & 0.000640 & 1.27\% \\
    KDA+\sdg & $0.000834$ & $[-0.001152,\ 0.002742]$ & 0.000276 & 1.37\% \\
    Mamba-3+\sdg & $0.000710$ & $[-0.001282,\ 0.002695]$ & 0.000406 & 1.66\% \\
    GDN-2+\sdg & $-0.001342$ & $[-0.003787,\ 0.001129]$ & 0.000399 & 0.88\% \\
    \bottomrule
  \end{tabular}
\end{table}

\begin{figure}[!htbp]
  \centering
  \includegraphics[width=\linewidth]{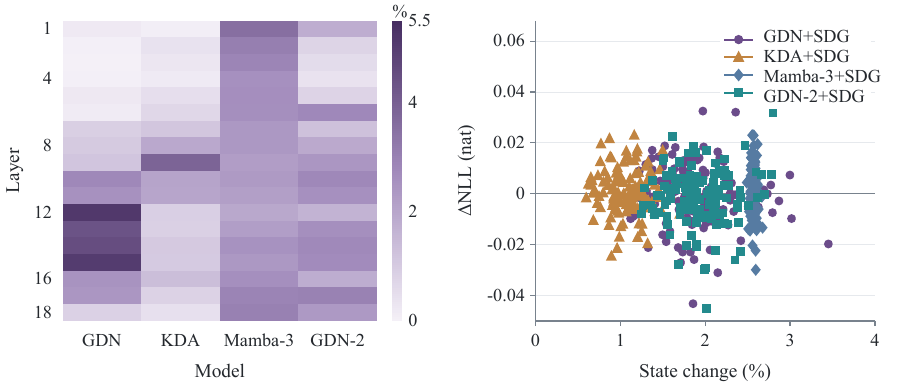}
  \setlength{\abovecaptionskip}{3pt}
  \caption{State changes at chunk boundaries. Left: relative Frobenius change per layer, averaged over heads and sequences, on a common scale. Right: sequence-level mean state change versus mean $\Delta$NLL over the first eight affected positions.}
  \label{fig:boundary-states}
\end{figure}

\paragraph{Changes to the state.}
For each head, we measure the relative change caused by the boundary gate as
\begin{equation}
  \frac{
    \left\|\mathbf S_{[n+1]}^{+}-\mathbf S_{[n+1]}^{-}\right\|_F
  }{
    \left\|\mathbf S_{[n+1]}^{-}\right\|_F
  }.
  \label{eq:boundary-relative-change}
\end{equation}
This measures the gate's effect on the state, not the change $\Delta\mathbf S_{[n]}$ across the chunk used to compute the gate. Figure~\ref{fig:boundary-states} averages Eq.~(\ref{eq:boundary-relative-change}) over heads and sequences for each layer. Its scatter plot relates each sequence's mean change across layers and heads to its mean loss change over the first eight affected positions. The median state changes across sequences, shown in Figure~\ref{fig:boundary-stability-full}, are approximately 1.85\%, 1.10\%, 2.57\%, and 1.91\% for GDN, KDA, Mamba-3, and GDN-2, respectively.

\paragraph{Individual loss increases.}
The 95th percentiles of sequence-level mean loss changes range from 0.0147 to 0.0161 nat across the four models. Individual-token analysis covers GDN, KDA, and GDN-2. Among GDN's 4,096 affected positions, one has a loss increase of 0.656645 nat, occurring 16 positions after the boundary at input position 3072. The normal and skip losses are 1.864409 and 1.207765, respectively, so the normal branch assigns the target about 51.9\% of the probability assigned by the skip branch. Both top-1 predictions are the same and incorrect. This event contributes to the response at position 16 in Figure~\ref{fig:boundary-response}, outside the primary eight-position window. Its sequence-level mean state change is about 1.97\%, so an unusually large average state change does not identify it. KDA and GDN-2 have no positions with a loss increase above 0.5 nat in this sample.

\paragraph{Boundary-offset diagnostic.}
We insert valid filler tokens into benchmark inputs to place the start of answer generation before and after chunk boundaries. As specified in Appendix~\ref{app:reference-boundary-scan}, these tokens enter the native recurrence and advance the within-chunk token counter. They therefore shift answer generation relative to the $C$-token boundary schedule. Across the tested benchmarks, absolute changes in accuracy are at most 0.5 percentage points on commonsense reasoning and 1 percentage point on retrieval. These are observed changes in benchmark accuracy, measured in percentage points rather than relative percentages. Because the filler tokens also change the input sequence, this diagnostic measures sensitivity to the combined change in input and boundary alignment. It complements the paired boundary intervention above, which holds the token sequence fixed and skips one boundary update.

\FloatBarrier
\section{State updates and implementation}
\label{app:backbone-details}

\subsection{Extended preliminaries}
\label{app:extended-preliminaries}

\paragraph{State convention.}
Equation~(\ref{eq:state-read}) uses value channels as state rows and key channels as columns. The output reads the state after the current write, so it includes the current token. At a fresh sequence start, $\mathbf S_0=\mathbf 0$. The state size is independent of sequence length $L$.

\paragraph{Native token updates.}
\textbf{DeltaNet}~\citep{schlag2021linear,yang2024deltanet} corrects the value stored at the current key using a gradient step~\citep{widrow1960adaptive} on the local regression loss
\begin{equation}
  \ell_t(\mathbf S)
  =
  \frac12\|\mathbf S\bm k_t-\bm v_t\|_2^2.
  \label{eq:delta-loss-preliminary}
\end{equation}
Starting from $\mathbf S_{t-1}$, a gradient step with data-dependent step size $\beta_t\in[0,1]$ gives
\begin{equation}
  \mathbf S_t
  =
  \mathbf S_{t-1}
  +\beta_t(\bm v_t-\mathbf S_{t-1}\bm k_t)\bm k_t^\top
  =
  \mathbf S_{t-1}(\mathbf I-\beta_t\bm k_t\bm k_t^\top)
  +\beta_t\bm v_t\bm k_t^\top.
  \label{eq:delta-preliminary}
\end{equation}
For unit-norm keys, $\beta_t$ controls how much of the stored value is replaced by $\bm v_t$.

\textbf{Gated DeltaNet (GDN)}~\citep{yang2024gdn} applies a data-dependent decay gate $\alpha_t\in(0,1]$ before the delta update:
\begin{equation}
  \mathbf S_t
  =
  \alpha_t\mathbf S_{t-1}
  (\mathbf I-\beta_t\bm k_t\bm k_t^\top)
  +\beta_t\bm v_t\bm k_t^\top.
  \label{eq:gdn-preliminary}
\end{equation}
The scalar $\alpha_t$ sets one retention rate for all state elements within a head.

\textbf{Kimi Delta Attention (KDA)}~\citep{kimi2025linear} assigns a decay coefficient to each key channel. Let $\mathbf D_t=\operatorname{Diag}(\bm\alpha_t)$ with $\bm\alpha_t\in(0,1]^{d_k}$. The state update becomes
\begin{equation}
  \mathbf S_t
  =
  \mathbf S_{t-1}\mathbf D_t
  (\mathbf I-\beta_t\bm k_t\bm k_t^\top)
  +\beta_t\bm v_t\bm k_t^\top.
  \label{eq:kda-preliminary}
\end{equation}
Each decay coefficient is shared across value channels, while $\beta_t$ controls both erasing and writing.

\textbf{Gated DeltaNet-2 (GDN-2)}~\citep{hatamizadeh2026gdn2} replaces the shared delta step size $\beta_t$ with separate erase and write gates, retaining the decay matrix $\mathbf D_t$. Let $\bm e_t=\bm b_t\odot\bm k_t$ and $\bm z_t=\bm w_t\odot\bm v_t$, where $\bm b_t\in[0,1]^{d_k}$ and $\bm w_t\in[0,1]^{d_v}$ are the erase and write gates. Its update is
\begin{equation}
  \mathbf S_t
  =
  \mathbf S_{t-1}\mathbf D_t
  (\mathbf I-\bm e_t\bm k_t^\top)
  +\bm z_t\bm k_t^\top.
  \label{eq:gdn2-preliminary}
\end{equation}
The erase gate weights the key channels used to read stored content, and the write gate scales the incoming value channels. Setting $\bm b_t=\beta_t\mathbf 1_{d_k}$ and $\bm w_t=\beta_t\mathbf 1_{d_v}$ recovers KDA. Further setting $\mathbf D_t=\alpha_t\mathbf I$ recovers GDN.

In these layers, the gate coefficients are predicted from token representations without directly reading the recurrent state; the delta correction itself reads that state. For fixed token inputs, every value row is propagated through the same transition in key space. This shared structure supports the chunkwise computation in Section~\ref{sec:chunkwise-preliminary}.

The Mamba-3 SISO backbone~\citep{lahoti2026mamba3} has a matrix state and no delta read-and-correct term. Its state-space recurrence and our boundary gating are detailed in Appendix~\ref{app:mamba3-chunkwise}.

\paragraph{Additive chunk computation.}
We use the complete-chunk convention stated in Section~\ref{sec:chunkwise-preliminary}.
The chunkwise kernels considered here combine a recurrent state scan across chunks with parallel matrix operations within each chunk~\citep{yang2024gla,qin2024lightning,yang2024deltanet,yang2024gdn,kimi2025linear,hatamizadeh2026gdn2}. Chunk $n\in\{1,\ldots,N\}$ contains tokens $(n-1)C+1,\ldots,nC$, with incoming state $\mathbf S_{[n]}=\mathbf S_{(n-1)C}$. Let $\mathbf Q_{[n]},\mathbf K_{[n]}\in\mathbb R^{C\times d_k}$ and $\mathbf V_{[n]},\mathbf O_{[n]}\in\mathbb R^{C\times d_v}$ stack the corresponding token vectors as rows. In the additive case of Eq.~(\ref{eq:state-read}),
\begin{align}
  \mathbf S_{[n+1]}
  &=
  \mathbf S_{[n]}+\mathbf V_{[n]}^\top\mathbf K_{[n]},
  \label{eq:additive-chunk-state}\\
  \mathbf O_{[n]}
  &=
  \mathbf Q_{[n]}\mathbf S_{[n]}^\top
  +
  \left(
    \mathbf Q_{[n]}\mathbf K_{[n]}^\top\odot\mathbf M_C
  \right)\mathbf V_{[n]},
  \label{eq:additive-chunk-output}
\end{align}
where $\mathbf M_C$ is a binary lower-triangular mask including the diagonal. The two output terms read the incoming memory and combine causal interactions within the chunk. For fixed chunk size and head dimensions, the arithmetic work scales linearly with sequence length.

\paragraph{Shared factors and the incoming-memory response.}
Gated delta-rule layers combine cumulative decay with compact WY representations of the accumulated rank-one updates~\citep{bischof1987wy,yang2024deltanet,yang2024gdn,kimi2025linear,hatamizadeh2026gdn2}. Their factors are shared across value rows and can be computed from token representations before the state scan. Let $\mathbf P_{[n]}\in\mathbb R^{d_k\times d_k}$ denote the shared chunk transition and $\boldsymbol\Omega_{[n]}\in\mathbb R^{d_v\times d_k}$ the state produced from zero incoming memory. The resulting affine update and state difference are given in Eq.~(\ref{eq:preliminary-affine-chunk}).
For fixed token representations and native gates, $\mathbf P_{[n]}$ and $\boldsymbol\Omega_{[n]}$ do not depend on this layer's incoming state. The additive case has $\mathbf P_{[n]}=\mathbf I$, so its state change contains only new writes. In the gated layers, the change also reflects how the chunk transforms the existing memory. Appendix~\ref{app:chunk-closed-form} gives the exact chunk formulas.

Training and prefill use chunkwise computation; decoding applies the native recurrence one token at a time. Section~\ref{sec:method} uses the incoming and updated chunk states to compute an additional boundary gate.

\subsection{Chunkwise notation and boundary gating}
\label{app:chunkwise-notation}

We present each backbone in the WY/UT matrix form used by GDN~\citep{yang2024gdn}, KDA~\citep{kimi2025linear}, and GDN-2~\citep{hatamizadeh2026gdn2}. We retain the state orientation of this paper, $\mathbf S\in\mathbb R^{d_v\times d_k}$, so the KDA and GDN-2 state equations are transposed relative to their original presentations. For one head and complete chunk $n$, $\mathbf Q,\mathbf K\in\mathbb R^{C\times d_k}$ and $\mathbf V\in\mathbb R^{C\times d_v}$ stack the native token vectors. We suppress $[n]$ on these chunk-local quantities and their WY factors, but retain it on the input state, output block, and state passed to the next chunk. Let $\operatorname{tril}(\cdot)$ include the diagonal and $\operatorname{tril}(\cdot,-1)$ exclude it; $\oslash$ denotes element-wise division.

In every model, $\mathbf S_{[n]}^+$ enters the original chunk computation, which produces $\mathbf O_{[n]}$ and the raw state $\mathbf S_{[n+1]}^-$. SDG then computes
\begin{equation}
  \begin{aligned}
    \Delta\mathbf S_{[n]}&=\mathbf S_{[n+1]}^- -\mathbf S_{[n]}^+,\\
    \mathbf Z_{[n]}&=\mathbf G\odot
      (\Delta\mathbf S_{[n]}\mathbf W_S^\top),\\
    \boldsymbol\Lambda_{[n]}&=\exp\!\left[-\mathbf Z_{[n]}-
      \sqrt{\mathbf Z_{[n]}\odot\mathbf Z_{[n]}+\rho_0^2\mathbf 1}\right].
  \end{aligned}
  \label{eq:backbone-sdg-insertion}
\end{equation}
We display the final update $\mathbf S_{[n+1]}^+=\boldsymbol\Lambda_{[n]}\odot\mathbf S_{[n+1]}^-$ alongside each backbone's chunk equations below. In all four cases \sdg changes only the matrix carried to the next chunk: the gate acts on the entire raw state including the chunk's writes, whereas the WY factors, the pseudo-value term, the intra-chunk product, the output block, and the native query scaling, normalization, output gates, and output projections are untouched. The WY factors are computed from token-local inputs, while $\boldsymbol\Lambda_{[n]}$ is evaluated inside the state scan. We do not repeat this per backbone below.
\subsection{Derivation of the chunk closed forms}
\label{app:chunk-closed-form}

The closed forms in Appendices~\ref{app:gdn-chunkwise}--\ref{app:gdn2-chunkwise} are three specializations of one lemma. Within a complete chunk, stack $\bm e_i^\top$ into $\mathbf E\in\mathbb R^{C\times d_k}$ and $\bm z_i^\top$ into $\mathbf Z\in\mathbb R^{C\times d_v}$ (chunk-local matrices; this $\mathbf Z$ is not the \sdg score $\mathbf Z_{[n]}$), and let $\bm\gamma_i=\bigodot_{j\le i}\bm\alpha_j$ with the convention $\bm\gamma_0=\mathbf 1$. Define the cumulative and remaining decay matrices $\boldsymbol\Gamma^{\mathrm{pre}},\boldsymbol\Gamma^{\mathrm{tail}}\in\mathbb R^{C\times d_k}$ row-wise by
\begin{equation}
  \boldsymbol\Gamma^{\mathrm{pre}}_{i,:}=\bm\gamma_i^\top,
  \qquad
  \boldsymbol\Gamma^{\mathrm{tail}}_{i,:}=(\bm\gamma_C\oslash\bm\gamma_i)^\top,
  \label{eq:unified-decay}
\end{equation}
together with
\begin{equation}
  \bar{\mathbf K}=\mathbf K\oslash\boldsymbol\Gamma^{\mathrm{pre}},
  \quad
  \bar{\mathbf E}=\boldsymbol\Gamma^{\mathrm{pre}}\odot\mathbf E,
  \quad
  \mathbf A=\bigl[\mathbf I_C+\operatorname{tril}(\bar{\mathbf E}\bar{\mathbf K}^\top,-1)\bigr]^{-1},
  \quad
  \mathbf Y=\mathbf A\bar{\mathbf E},
  \quad
  \mathbf U=\mathbf A\mathbf Z .
  \label{eq:unified-wy}
\end{equation}

\begin{lemma}[Unified chunk closed form]
\label{lem:chunk-closed-form}
Let $\mathbf S_i=\mathbf S_{i-1}\mathbf D_i(\mathbf I-\bm e_i\bm k_i^\top)+\bm z_i\bm k_i^\top$ with $\mathbf D_i=\operatorname{Diag}(\bm\alpha_i)$ and $\mathbf S_0=\mathbf S_{[n]}^+$. Then
\begin{equation}
  \mathbf S_i
  =\mathbf S_0\operatorname{Diag}(\bm\gamma_i)
   +\sum_{j\le i}\bm\delta_j
     \bigl(\bm k_j\odot\bm\gamma_i\oslash\bm\gamma_j\bigr)^\top,
  \qquad
  \boldsymbol\Delta=\mathbf U-\mathbf Y\mathbf S_0^\top,
  \label{eq:unified-state-expansion}
\end{equation}
where $\boldsymbol\Delta\in\mathbb R^{C\times d_v}$ stacks the rows $\bm\delta_j^\top$. Consequently
\begin{equation}
  \mathbf S_{[n+1]}^-
  =\mathbf S_0\operatorname{Diag}(\bm\gamma_C)
   +\boldsymbol\Delta^\top(\boldsymbol\Gamma^{\mathrm{tail}}\odot\mathbf K),
  \qquad
  \mathbf O_{[n]}
  =\mathbf Q_\gamma\mathbf S_0^\top
   +\operatorname{tril}\bigl(\mathbf Q_\gamma\bar{\mathbf K}^\top\bigr)
    \boldsymbol\Delta,
  \label{eq:unified-state-output}
\end{equation}
with $\mathbf Q_\gamma=\boldsymbol\Gamma^{\mathrm{pre}}\odot\mathbf Q$.
\end{lemma}

\begin{proof}
Induct on $i$. The claim holds at $i=0$. Assume it at $i-1$. Since $\operatorname{Diag}(\bm\gamma_{i-1})\mathbf D_i=\operatorname{Diag}(\bm\gamma_i)$ and $(\bm k_j\odot\bm\gamma_{i-1}\oslash\bm\gamma_j)^\top\mathbf D_i=(\bm k_j\odot\bm\gamma_i\oslash\bm\gamma_j)^\top$,
\begin{equation*}
  \mathbf S_{i-1}\mathbf D_i
  =\mathbf S_0\operatorname{Diag}(\bm\gamma_i)
   +\sum_{j<i}\bm\delta_j
    \bigl(\bm k_j\odot\bm\gamma_i\oslash\bm\gamma_j\bigr)^\top .
\end{equation*}
Writing $\bm\delta_i=\bm z_i-\mathbf S_{i-1}\mathbf D_i\bm e_i$ gives $\mathbf S_i=\mathbf S_{i-1}\mathbf D_i+\bm\delta_i\bm k_i^\top$, which is Eq.~(\ref{eq:unified-state-expansion}) at $i$ because $\bm\gamma_i\oslash\bm\gamma_i=\mathbf 1$. Using $\operatorname{Diag}(\bm\gamma_i)\bm e_i=\bar{\mathbf E}_{i,:}^\top$ and $(\bm k_j\odot\bm\gamma_i\oslash\bm\gamma_j)^\top\bm e_i=\bar{\mathbf K}_{j,:}\bar{\mathbf E}_{i,:}^\top$,
\begin{equation*}
  \bm\delta_i
  +\sum_{j<i}\bigl(\bar{\mathbf E}\bar{\mathbf K}^\top\bigr)_{ij}\bm\delta_j
  =\bm z_i-\mathbf S_0\bar{\mathbf E}_{i,:}^\top ,
\end{equation*}
that is $\bigl[\mathbf I_C+\operatorname{tril}(\bar{\mathbf E}\bar{\mathbf K}^\top,-1)\bigr]\boldsymbol\Delta=\mathbf Z-\bar{\mathbf E}\mathbf S_0^\top$, which is the stated $\boldsymbol\Delta$. Evaluating Eq.~(\ref{eq:unified-state-expansion}) at $i=C$ gives the state in Eq.~(\ref{eq:unified-state-output}), and contracting row $i$ with $\bm q_i$ gives the output; the mask includes the diagonal because $j\le i$.
\end{proof}

Specializations. GDN-2 takes $\mathbf E=\mathbf B\odot\mathbf K$ and $\mathbf Z=\mathbf Z^{\mathrm{write}}$, so Eq.~(\ref{eq:unified-wy}) coincides with Eq.~(\ref{eq:gdn2-wy-factors}) and Eq.~(\ref{eq:unified-state-output}) reproduces Eqs.~(\ref{eq:gdn2-chunk-state})--(\ref{eq:gdn2-chunk-output}). KDA takes $\bm e_i=\beta_i\bm k_i$ and $\bm z_i=\beta_i\bm v_i$, so $\bar{\mathbf E}=\mathbf B_\beta(\boldsymbol\Gamma^{\mathrm{pre}}\odot\mathbf K)$ and $\mathbf Z=\mathbf B_\beta\mathbf V$. Its transform in Eq.~(\ref{eq:kda-ut}) equals $\mathbf A\mathbf B_\beta$ for the $\mathbf A$ of Eq.~(\ref{eq:unified-wy}), so the lemma's factors coincide with Eq.~(\ref{eq:kda-wy-factors}): $\mathbf Y=\mathbf A\mathbf B_\beta(\boldsymbol\Gamma^{\mathrm{pre}}\odot\mathbf K)=\mathbf W$ and $\mathbf U=\mathbf A\mathbf B_\beta\mathbf V$. GDN further sets $\bm\alpha_i=\alpha_i\mathbf 1_{d_k}$ with a scalar $\alpha_i$, so $\bar{\mathbf E}\bar{\mathbf K}^\top=\operatorname{Diag}(\bm\gamma)\mathbf B_\beta\mathbf K\mathbf K^\top\operatorname{Diag}(\bm\gamma)^{-1}$. Diagonal similarity preserves strict lower-triangular structure entrywise, so
\begin{equation}
  \mathbf A
  =\operatorname{Diag}(\bm\gamma)
   \bigl[\mathbf I_C+\operatorname{tril}(\mathbf B_\beta\mathbf K\mathbf K^\top,-1)\bigr]^{-1}
   \operatorname{Diag}(\bm\gamma)^{-1}
  =\bigl[\mathbf I_C+\operatorname{tril}
     \bigl(\mathbf B_\beta(\mathbf K\mathbf K^\top\odot\boldsymbol\Gamma),-1\bigr)\bigr]^{-1},
  \label{eq:gdn-similarity}
\end{equation}
whence $\mathbf Y=\operatorname{Diag}(\bm\gamma)\mathbf W$ with $\mathbf W$ from Eq.~(\ref{eq:gdn-ut-w}), and $\mathbf U=\mathbf A\mathbf B_\beta\mathbf V$ coincides with Eq.~(\ref{eq:gdn-ut-u}). Equations~(\ref{eq:gdn-chunk-state})--(\ref{eq:gdn-chunk-output}) then follow from Eq.~(\ref{eq:unified-state-output}). Numerically, $\bar{\mathbf K}=\mathbf K\oslash\boldsymbol\Gamma^{\mathrm{pre}}$ is never formed explicitly; the kernels evaluate the required ratios through differences of cumulative log-decays, as noted at the end of Appendix~\ref{app:gdn2-chunkwise}.

\subsection{Derivation of the Mamba-3 chunk form}
\label{app:mamba3-closed-form}

For Eq.~(\ref{eq:mamba3-trapezoid-recurrence}), we claim
\begin{equation}
  \mathbf S_i=\gamma_i\mathbf H_{[n]}
  +\sum_{j\le i}w^{(i)}_j\mathbf J_j,
  \qquad
  w^{(i)}_j=
  \begin{cases}
    (\gamma_i/\gamma_j)\eta_j, & j<i,\\
    c_i, & j=i,
  \end{cases}
  \label{eq:mamba3-expansion}
\end{equation}
with $\mathbf H_{[n]}$ and $\eta_j$ from Eq.~(\ref{eq:mamba3-chunk-carry}). At $i=1$, $\mathbf S_1=\alpha_1(\mathbf S_0+p_1\mathbf J_0)+c_1\mathbf J_1=\gamma_1\mathbf H_{[n]}+c_1\mathbf J_1$. Assume Eq.~(\ref{eq:mamba3-expansion}) at $i-1$. Then $\alpha_i\gamma_{i-1}=\gamma_i$; for $j<i-1$, $\alpha_iw^{(i-1)}_j=(\gamma_i/\gamma_j)\eta_j$; for $j=i-1$ the decayed diagonal term $\alpha_ic_{i-1}$ combines with the explicit $\alpha_ip_i\mathbf J_{i-1}$ to give $\alpha_i(c_{i-1}+p_i)=(\gamma_i/\gamma_{i-1})\eta_{i-1}$; and $c_i\mathbf J_i$ supplies the new diagonal term. This is Eq.~(\ref{eq:mamba3-expansion}) at $i$ and matches $\boldsymbol\Gamma^{\mathrm{trap}}$ in Eq.~(\ref{eq:mamba3-chunk-mask}). Setting $i=C$ gives $w^{(C)}_j=\omega_j$ and $\sum_j\omega_j\bm v_j\bm k_j^\top=\mathbf V^\top\operatorname{Diag}(\bm\omega)\mathbf K$, which is Eq.~(\ref{eq:mamba3-chunk-state}); contracting Eq.~(\ref{eq:mamba3-expansion}) with $\bm q_i$ gives Eq.~(\ref{eq:mamba3-chunk-output}). The weight $\eta_C=c_C$ excludes the next token's trapezoidal term, which is why that term is added only after the boundary gate, as in Eq.~(\ref{eq:mamba3-sdg-carry}).
\subsection{Gated DeltaNet}
\label{app:gdn-chunkwise}

\paragraph{Cumulative decay and WY factors.}
For GDN~\citep{yang2024gdn}, define $\gamma_i=\prod_{j=1}^i\alpha_j$, $\bm\gamma=(\gamma_1,\ldots,\gamma_C)^\top$, and $\mathbf B_\beta=\operatorname{Diag}(\beta_1,\ldots,\beta_C)$. Its causal decay mask is
\begin{equation}
  \Gamma_{ij}=
  \begin{cases}\gamma_i/\gamma_j,&i\geq j,\\0,&i<j.\end{cases}
  \label{eq:gdn-chunk-decay}
\end{equation}
Using the original GDN convention, the rows $\bm w_i^\top,\bm u_i^\top$ of the WY factors satisfy
\begin{align}
  \bm w_i&=\beta_i\left(\bm k_i-
    \sum_{j<i}(\bm k_j^\top\bm k_i)\bm w_j\right),
    \label{eq:gdn-wy-row-w}\\
  \bm u_i&=\beta_i\left(\bm v_i-
    \sum_{j<i}\frac{\gamma_i}{\gamma_j}
      (\bm k_j^\top\bm k_i)\bm u_j\right).
    \label{eq:gdn-wy-row-u}
\end{align}
The UT form writes these factors as matrix products with unit lower-triangular inverses:
\begin{align}
  \mathbf W&=
    \left[\mathbf I_C+
      \operatorname{tril}(\mathbf B_\beta\mathbf K\mathbf K^\top,-1)\right]^{-1}
      \mathbf B_\beta\mathbf K,
      \label{eq:gdn-ut-w}\\
  \mathbf U&=
    \left[\mathbf I_C+
      \operatorname{tril}\!\left(
        \mathbf B_\beta(\mathbf K\mathbf K^\top\odot\boldsymbol\Gamma),-1
      \right)\right]^{-1}\mathbf B_\beta\mathbf V.
      \label{eq:gdn-ut-u}
\end{align}
Here $\mathbf W\in\mathbb R^{C\times d_k}$ is a WY factor, distinct from the learned $\mathbf W_S$. Scalar cumulative decay remains outside $\mathbf W$ in the following equations.

\paragraph{Chunkwise computation with SDG.}
The state and output have the GDN chunkwise form, followed by our boundary update:
\begin{align}
  \mathbf S_{[n+1]}^-
    &=\gamma_C\mathbf S_{[n]}^+
      +\left(\mathbf U-\operatorname{Diag}(\bm\gamma)\mathbf W
        (\mathbf S_{[n]}^+)^\top\right)^\top
        \operatorname{Diag}(\gamma_C/\bm\gamma)\mathbf K,
      \label{eq:gdn-chunk-state}\\
  \mathbf O_{[n]}
    &=\operatorname{Diag}(\bm\gamma)\mathbf Q(\mathbf S_{[n]}^+)^\top
      +(\mathbf Q\mathbf K^\top\odot\boldsymbol\Gamma)
        \left(\mathbf U-\operatorname{Diag}(\bm\gamma)\mathbf W
          (\mathbf S_{[n]}^+)^\top\right),
      \label{eq:gdn-chunk-output}\\
  \mathbf S_{[n+1]}^+
    &=\underbrace{\boldsymbol\Lambda_{[n]}\odot\mathbf S_{[n+1]}^-}
        _{\text{SDG boundary update}}.
      \label{eq:gdn-chunk-sdg}
\end{align}
The pseudo-value term $\mathbf U-\operatorname{Diag}(\bm\gamma)\mathbf W(\mathbf S_{[n]}^+)^\top$ is shared by the state and output computations. SDG leaves this term and the WY preparation unchanged, and replaces the raw state carried to the next chunk with the final line above.

\subsection{Kimi Delta Attention}
\label{app:kda-chunkwise}

\paragraph{Channel-wise decay and the UT transform.}
For KDA~\citep{kimi2025linear}, let $\bm\gamma_i=\bigodot_{j=1}^i\bm\alpha_j$ and recall Eq.~(\ref{eq:unified-decay}),
\begin{equation}
  \boldsymbol\Gamma^{\mathrm{pre}}_{i,:}=\bm\gamma_i^\top,
  \qquad
  \boldsymbol\Gamma^{\mathrm{tail}}_{i,:}
    =(\bm\gamma_C\oslash\bm\gamma_i)^\top.
  \label{eq:kda-chunk-decay}
\end{equation}
These are the cumulative decay from the chunk start and the remaining decay to its end, respectively. With $\mathbf B_\beta=\operatorname{Diag}(\beta_1,\ldots,\beta_C)$, KDA's UT transform and WY factors are
\begin{align}
  \mathbf A
    &=\left[\mathbf I_C+
      \operatorname{tril}\!\left(
        \mathbf B_\beta(\boldsymbol\Gamma^{\mathrm{pre}}\odot\mathbf K)
        (\mathbf K\oslash\boldsymbol\Gamma^{\mathrm{pre}})^\top,-1
      \right)\right]^{-1}\mathbf B_\beta,
      \label{eq:kda-ut}\\
  \mathbf W&=\mathbf A(\boldsymbol\Gamma^{\mathrm{pre}}\odot\mathbf K),
    \qquad \mathbf U=\mathbf A\mathbf V.
      \label{eq:kda-wy-factors}
\end{align}
Here $\mathbf A$ denotes the UT transform, reserving $\mathbf M_C$ for the causal mask. In this convention, cumulative channel-wise decay is already included in $\mathbf W$.

\paragraph{Chunkwise computation with SDG.}
The resulting state, output, and boundary update are
\begin{align}
  \mathbf S_{[n+1]}^-
    &=\mathbf S_{[n]}^+\operatorname{Diag}(\bm\gamma_C)
      +\left(\mathbf U-\mathbf W(\mathbf S_{[n]}^+)^\top\right)^\top
        (\boldsymbol\Gamma^{\mathrm{tail}}\odot\mathbf K),
      \label{eq:kda-chunk-state}\\
  \mathbf O_{[n]}
    &=(\boldsymbol\Gamma^{\mathrm{pre}}\odot\mathbf Q)
        (\mathbf S_{[n]}^+)^\top\nonumber\\
    &\quad+\operatorname{tril}\!\left(
      (\boldsymbol\Gamma^{\mathrm{pre}}\odot\mathbf Q)
      (\mathbf K\oslash\boldsymbol\Gamma^{\mathrm{pre}})^\top
      \right)
      \left(\mathbf U-\mathbf W(\mathbf S_{[n]}^+)^\top\right),
      \label{eq:kda-chunk-output}\\
  \mathbf S_{[n+1]}^+
    &=\underbrace{\boldsymbol\Lambda_{[n]}\odot\mathbf S_{[n+1]}^-}
        _{\text{SDG boundary update}}.
      \label{eq:kda-chunk-sdg}
\end{align}
The two output terms are the inter-chunk read and intra-chunk masked matrix product. The shared pseudo-values remain $\mathbf U-\mathbf W(\mathbf S_{[n]}^+)^\top$. KDA's diagonal token decay is retained; SDG adds element-wise decay only after the raw chunk state has been formed.

\subsection{Gated DeltaNet-2}
\label{app:gdn2-chunkwise}

\paragraph{Asymmetric erase and write factors.}
GDN-2~\citep{hatamizadeh2026gdn2} uses the same $\boldsymbol\Gamma^{\mathrm{pre}}$ and $\boldsymbol\Gamma^{\mathrm{tail}}$ as KDA. Let $\mathbf B\in\mathbb R^{C\times d_k}$ and $\mathbf W^{\mathrm{write}}\in\mathbb R^{C\times d_v}$ stack its erase gates $\bm b_i$ and write gates $\bm w_i$. Following its asymmetric WY form, define
\begin{equation}
  \bar{\mathbf K}=\mathbf K\oslash\boldsymbol\Gamma^{\mathrm{pre}},
  \qquad
  \bar{\mathbf E}=\boldsymbol\Gamma^{\mathrm{pre}}\odot(\mathbf B\odot\mathbf K),
  \qquad
  \mathbf Z^{\mathrm{write}}=\mathbf W^{\mathrm{write}}\odot\mathbf V.
  \label{eq:gdn2-chunk-factors}
\end{equation}
The superscript $\mathrm{write}$ distinguishes these token-write quantities from the SDG score and WY factors. GDN-2's triangular inverse and erase/write auxiliaries are
\begin{equation}
  \mathbf T=\operatorname{tril}(\bar{\mathbf E}\bar{\mathbf K}^\top,-1),
  \qquad \mathbf A=(\mathbf I_C+\mathbf T)^{-1},
  \qquad \mathbf Y=\mathbf A\bar{\mathbf E},
  \qquad \mathbf U=\mathbf A\mathbf Z^{\mathrm{write}}.
  \label{eq:gdn2-wy-factors}
\end{equation}
The notation $\mathbf Y$ follows the erase-side auxiliary in GDN-2. Define its output and tail factors as
\begin{equation}
  \mathbf Q_\gamma=\boldsymbol\Gamma^{\mathrm{pre}}\odot\mathbf Q,
  \qquad \mathbf K_{\mathrm{tail}}=\boldsymbol\Gamma^{\mathrm{tail}}\odot\mathbf K,
  \qquad \mathbf A_{qk}=\operatorname{tril}(\mathbf Q_\gamma\bar{\mathbf K}^\top).
  \label{eq:gdn2-output-factors}
\end{equation}

\paragraph{Chunkwise computation with SDG.}
In the state orientation used here, the GDN-2 equations and our insertion are
\begin{align}
  \mathbf S_{[n+1]}^-
    &=\mathbf S_{[n]}^+\operatorname{Diag}(\bm\gamma_C)
      +\left(\mathbf U-\mathbf Y(\mathbf S_{[n]}^+)^\top\right)^\top
        \mathbf K_{\mathrm{tail}},
      \label{eq:gdn2-chunk-state}\\
  \mathbf O_{[n]}
    &=\mathbf Q_\gamma(\mathbf S_{[n]}^+)^\top
      +\mathbf A_{qk}\left(\mathbf U-\mathbf Y(\mathbf S_{[n]}^+)^\top\right),
      \label{eq:gdn2-chunk-output}\\
  \mathbf S_{[n+1]}^+
    &=\underbrace{\boldsymbol\Lambda_{[n]}\odot\mathbf S_{[n+1]}^-}
        _{\text{SDG boundary update}}.
      \label{eq:gdn2-chunk-sdg}
\end{align}
The erase and write gates enter through $\bar{\mathbf E}$ and $\mathbf Z^{\mathrm{write}}$, respectively. SDG changes only the matrix carried between chunks; $\mathbf A_{qk}$, $\mathbf K_{\mathrm{tail}}$, and the native WY preparation are retained. Tying $\bm b_i$ and $\bm w_i$ to $\beta_i$ recovers KDA, with $\mathbf Y=\mathbf W$. Under scalar decay, this equals $\operatorname{Diag}(\bm\gamma)\mathbf W$ in the GDN convention above.

All triangular inverses above denote solves with unit lower-triangular matrices. Decay ratios are evaluated through cumulative log-decays and their differences in the kernels. The mathematical notation does not require explicitly storing inverse cumulative decays.

\subsection{Mamba-3 SISO}
\label{app:mamba3-chunkwise}

\paragraph{Trapezoidal write factors.}
For Mamba-3~\citep{lahoti2026mamba3}, $\mathbf Q,\mathbf K\in\mathbb R^{C\times d_s}$ contain the native normalized, biased, and rotated queries and keys. Let $\delta_i>0$ be the step size, $\tau_i\in[0,1]$ the trapezoidal mixing coefficient, and $\alpha_i$ the scalar decay. Define
\begin{equation}
  p_i=(1-\tau_i)\delta_i,
  \qquad c_i=\tau_i\delta_i,
  \qquad \mathbf J_i=\bm v_i\bm k_i^\top.
  \label{eq:mamba3-trapezoid-coefficients}
\end{equation}
Its native physical-state recurrence, with $\mathbf S_0=\mathbf S_{[n]}^+$, is
\begin{equation}
  \mathbf S_i=\alpha_i\mathbf S_{i-1}
    +\alpha_i p_i\mathbf J_{i-1}+c_i\mathbf J_i,
  \qquad \bm o_i=\mathbf S_i\bm q_i.
  \label{eq:mamba3-trapezoid-recurrence}
\end{equation}
To give the same inter-chunk plus intra-chunk matrix presentation, set $\gamma_i=\prod_{j=1}^i\alpha_j$, $\bm\gamma=(\gamma_1,\ldots,\gamma_C)^\top$, and
\begin{equation}
  \mathbf H_{[n]}=\mathbf S_{[n]}^++p_1\mathbf J_0,
  \qquad
  \eta_j=\begin{cases}c_j+p_{j+1},&j<C,\\c_C,&j=C.\end{cases}
  \label{eq:mamba3-chunk-carry}
\end{equation}
Here $\mathbf J_0$ uses the preceding token's cached key and value, or is zero at a fresh sequence start; the carry $\mathbf H_{[n]}$ is unrelated to the head count $H$.. The trapezoidal mask and tail weights are
\begin{equation}
  \Gamma^{\mathrm{trap}}_{ij}=
  \begin{cases}
    (\gamma_i/\gamma_j)\eta_j,&i>j,\\
    c_i,&i=j,\\
    0,&i<j,
  \end{cases}
  \qquad \omega_j=(\gamma_C/\gamma_j)\eta_j.
  \label{eq:mamba3-chunk-mask}
\end{equation}

\paragraph{Chunkwise computation with SDG.}
The SISO chunk computation can then be written as
\begin{align}
  \mathbf S_{[n+1]}^-
    &=\gamma_C\mathbf H_{[n]}
      +\mathbf V^\top\operatorname{Diag}(\bm\omega)\mathbf K,
      \label{eq:mamba3-chunk-state}\\
  \mathbf O_{[n]}
    &=\operatorname{Diag}(\bm\gamma)\mathbf Q\mathbf H_{[n]}^\top
      +(\mathbf Q\mathbf K^\top\odot\boldsymbol\Gamma^{\mathrm{trap}})\mathbf V,
      \label{eq:mamba3-chunk-output}\\
  \mathbf S_{[n+1]}^+
    &=\underbrace{\boldsymbol\Lambda_{[n]}\odot\mathbf S_{[n+1]}^-}
        _{\text{SDG boundary update}}.
      \label{eq:mamba3-chunk-sdg}
\end{align}
This has the same separation into an incoming-state read and a causal within-chunk matrix product. Mamba-3 has no delta-rule erase term, so its value factor is $\mathbf V$ rather than a WY pseudo-value. Its native skip connection and output gate are applied without modification.

\paragraph{Boundary placement and the next chunk.}
Equation~(\ref{eq:backbone-sdg-insertion}) uses the physical states $\mathbf S_{[n]}^+$ and $\mathbf S_{[n+1]}^-$, not the corrected carry $\mathbf H_{[n]}$. The last write weight $\eta_C=c_C$ excludes the next token's trapezoidal contribution. Only after SDG has updated the physical state is that contribution added to the next carry:
\begin{equation}
  \mathbf H_{[n+1]}
    =\mathbf S_{[n+1]}^+
      +p_{[n+1],1}\mathbf J_{[n],C}.
  \label{eq:mamba3-sdg-carry}
\end{equation}
The saved SDG source is $\mathbf S_{[n+1]}^+$, before this correction. Thus the gate does not scale the next token's previous-key/value contribution. The rotary phase and previous-key/value caches continue with their native updates.

\subsection{Reference boundary scan}
\label{app:reference-boundary-scan}

The following procedure describes how the state is updated during training, prefill, and decoding. Its cache contains the current state $\mathbf S$, the state $\mathbf S^{\src}$ at the preceding completed boundary, native auxiliary state $\bm\xi$, and a valid-token counter $r\in\{0,\ldots,C-1\}$. The procedure below makes the order of reads, writes, and boundary decay explicit. Let $\mathcal X$ denote the projected token inputs and native token gates. For a segment of the sequence, the input and output states satisfy
\begin{equation*}
  (\mathbf O^{\mathrm{seg}},\mathbf S^{\mathrm{out}},\bm\xi')
  =
  \mathcal F
  (\mathcal X^{\mathrm{seg}},\mathbf S^{\mathrm{in}},\bm\xi).
\end{equation*}
At an aligned chunk start, $\mathbf S^{\mathrm{in}}=\mathbf S^{\src}=\mathbf S_{[n]}^{+}$. During continuation of an incomplete chunk, $\mathbf S^{\mathrm{in}}$ is instead the current cached state and generally differs from the retained $\mathbf S^{\src}$ used by the eventual gate function.

\begin{enumerate}
    \setlength{\itemsep}{0.05em}
    \item \textbf{Input.} Native token-local quantities $\mathcal X$ for $L$ valid tokens, chunk size $C$, and cache $(\mathbf S,\mathbf S^{\src},\bm\xi,r)$. Set $p\gets1$ and $\mathbf O\gets[\,]$.
    \item While $p\le L$, set $m\gets\min\{C-r,L-p+1\}$ and run the original recurrence on the next contiguous segment.
      \begin{equation*}
        (\mathbf O^{\mathrm{seg}},\mathbf S^{\mathrm{out}},\bm\xi')
        \gets
        \mathcal F
        (\mathcal X_{p:p+m-1},\mathbf S,\bm\xi).
      \end{equation*}
    \item Append $\mathbf O^{\mathrm{seg}}$ to $\mathbf O$ and set $\mathbf S\gets\mathbf S^{\mathrm{out}}$, $\bm\xi\gets\bm\xi'$, $r\gets r+m$, and $p\gets p+m$.
    \item If $r=C$, compute $\boldsymbol\Lambda\gets \mathcal G_\theta(\mathbf S^{\src},\mathbf S^{\mathrm{out}})$ and then set $\mathbf S\gets\boldsymbol\Lambda\odot\mathbf S^{\mathrm{out}}$, $\mathbf S^{\src}\gets\mathbf S$, and $r\gets0$. The segment outputs and native auxiliary update have already been formed. The boundary changes neither.
    \item Repeat Steps 2--4 and return $\mathbf O,(\mathbf S,\mathbf S^{\src},\bm\xi,r)$.
\end{enumerate}
\begin{proposition}[Segmentation invariance]
\label{prop:segmentation}
Assume the base recurrence is segmentation consistent, that is, for every split of a token stream, $\mathcal F$ applied to the concatenation equals the sequential composition of $\mathcal F$ on the two parts with the intermediate state and auxiliary state passed through. Then for any partition of a stream into consecutive calls, the procedure above returns the same outputs and the same final cache as a single call on the whole stream, and both agree with the chunkwise definition of Eq.~(\ref{eq:boundary-update}) with boundaries at absolute positions that are multiples of $C$.
\end{proposition}

\begin{proof}
Induct on the number of calls, starting from $r=0$ at a fresh sequence. Step~2 chooses $m=\min\{C-r,L-p+1\}$, so no segment crosses a boundary and the hypothesis applies to each segment. The counter $r$ therefore records the absolute position modulo $C$, and the set of positions at which Step~4 fires is determined by absolute positions alone, independently of the partition. Between two consecutive firings, $\mathbf S$ and $\bm\xi$ evolve only through $\mathcal F$, whose composition is split invariant by assumption, and $\mathbf S^{\src}$ is written only in Step~4. Hence at the boundary after chunk $n$ the two arguments of $\mathcal G_\theta$ are exactly $\mathbf S_{[n]}^+$ and $\mathbf S_{[n+1]}^-$, as in Eq.~(\ref{eq:boundary-update}), and the outputs of each segment are formed before the gate is applied. For Mamba-3 the auxiliary state carries the previous key and value, and the gate acts on the physical state, so Eq.~(\ref{eq:mamba3-sdg-carry}) composes in the same way.
\end{proof}

The identity holds in exact arithmetic; in finite precision the two execution paths group the same products differently, so our correctness tests compare them with a tolerance rather than bitwise. Consequently training with the chunkwise form, prefill in one call, and decoding one token per call define the same function of the token stream.

The remaining conventions are implementation choices. For a fresh sequence $\mathbf S=\mathbf S^{\src}$ is the supplied initial state or zero and $\bm\xi$ uses the base model's initialization; each nonempty packed example owns an independent cache tuple, and empty packed examples are rejected. Masked padding neither invokes the recurrence nor advances $r$ in fresh, prefill, and packed execution, whereas the GDN-2, KDA, and Mamba-3 cached one-token paths assume that every cache row supplies one valid token. In the boundary-offset diagnostic, the inserted filler tokens are valid inputs that enter the recurrence and advance $r$, thereby shifting answer generation relative to chunk boundaries. A cache is retained only while continuing the same sequence and is reset when it ends. Step~4 also shows why $\mathbf S^{\src}$ must not be overwritten while an incomplete tail accumulates.

\subsection{Model implementations}

\paragraph{GDN, GDN-2, and KDA.}
All three adapters return current-chunk outputs before applying \sdg. Their native token gates, delta residuals, WY constructions, and output gates remain unchanged. The optimized implementations replace or extend only the persistent inter-chunk state scan and its vector--Jacobian product.

\paragraph{Mamba-3.}
For the bundled SISO recurrence, the native recurrent tuple is
\begin{equation}
  \left(
    \bm s^{\mathrm{angle}},
    \mathbf S^{\mathrm{ssm}},
    \bm k^{\mathrm{prev}},
    \bm v^{\mathrm{prev}}
  \right).
  \label{eq:mamba3-state-tuple}
\end{equation}
\sdg additionally stores the state $\mathbf S^{\src,\mathrm{ssm}}$ and tail counter $r$. Only $\mathbf S^{\mathrm{ssm}}$ is passed to $\mathcal G_\theta$ and multiplied by its decay gate. All other tuple components come directly from the native transition.

\paragraph{Supported configurations.}
The GDN adapter supports key and value head dimensions up to 128. The GDN-2 and KDA adapters currently require equal numbers of value and query/key heads. Their fused paths support $d_k,d_v\leq128$, with wider states using a fallback path. The current projected state-change configurations use boundary recomputation. The Mamba-3 adapter targets the bundled SISO implementation and exposes only successive one-token cached decoding calls. These are implementation limits, not requirements of the method.

\section{Complete proofs}
\label{app:proofs}

\subsection{Adaptive element-wise ridge}
\label{app:ridge-proof}

Remark~\ref{rem:ridge} is stated in Section~\ref{sec:composite-online-learning}. The following argument establishes the identity and its converse, followed by the backbone-specific token-update interpretations~\citep{duchi2009forwardbackward,duchi2010composite,parikh2014proximal}.

\paragraph{Proof of Remark~\ref{rem:ridge}.}
Write $\mathbf S=\mathbf S_{[n+1]}^-$, $\boldsymbol\rho=\boldsymbol\rho_{[n]}$, and $\boldsymbol\kappa=e^{\boldsymbol\rho}-\mathbf 1$. The objective
\begin{equation*}
  J(\mathbf M)=\tfrac12\|\mathbf M-\mathbf S\|_F^2
  +\tfrac12\sum_{a,b}\kappa_{ab}M_{ab}^2
\end{equation*}
is separable across entries and strictly convex because $1+\kappa_{ab}=e^{\rho_{ab}}>0$. From $\partial J/\partial M_{ab}=e^{\rho_{ab}}M_{ab}-S_{ab}$, the unique minimizer is $M_{ab}=e^{-\rho_{ab}}S_{ab}=\Lambda_{ab}S_{ab}$, which is Eq.~(\ref{eq:ridge-equivalence}). For the converse, we show that the quadratic is the only separable differentiable penalty with this proximal operator, up to an additive constant. Let $g(\mathbf M)=\sum_{a,b}g_{ab}(M_{ab})$ be separable and differentiable with $\operatorname{prox}_g(\mathbf S)=\boldsymbol\Lambda\odot\mathbf S$ for every $\mathbf S$. Stationarity at $M_{ab}=\Lambda_{ab}S_{ab}$ reads $\Lambda_{ab}S_{ab}-S_{ab}+g_{ab}'(\Lambda_{ab}S_{ab})=0$ for all $S_{ab}$, hence $g_{ab}'(u)=(\Lambda_{ab}^{-1}-1)u$ and $g_{ab}(u)=\tfrac12\kappa_{ab}u^2+\mathrm{const}$ with $\kappa_{ab}=\Lambda_{ab}^{-1}-1=e^{\rho_{ab}}-1$. Boundary decay is thus the prox of a uniquely determined ridge penalty, and $\boldsymbol\kappa=\boldsymbol\rho+O(\boldsymbol\rho^{\odot2})$ as $\boldsymbol\rho\to\mathbf 0$. \hfill$\square$

\paragraph{When the token update is a gradient step.}
Let $\widetilde{\mathbf S}=\mathbf S_{t-1}\mathbf D_t$ be the state after token decay, so that Eqs.~(\ref{eq:delta-preliminary})--(\ref{eq:gdn2-preliminary}) all take the form
\begin{equation}
  \mathbf S_t=\widetilde{\mathbf S}
    -\bigl(\widetilde{\mathbf S}\bm e_t-\bm z_t\bigr)\bm k_t^\top .
  \label{eq:generic-token-update}
\end{equation}

\begin{lemma}
\label{lem:gradient-step}
Fix $\bm k_t\ne\bm 0$, $\bm e_t$, and $\bm z_t$. The field $\mathbf F(\widetilde{\mathbf S})=(\widetilde{\mathbf S}\bm e_t-\bm z_t)\bm k_t^\top$ is the gradient of a scalar function of $\widetilde{\mathbf S}$ with respect to the Frobenius inner product if and only if $\bm e_t=c\,\bm k_t$ for some $c\in\mathbb R$.
\end{lemma}

\begin{proof}
$\mathbf F$ is affine with constant Jacobian $\mathcal A(\mathbf H)=(\mathbf H\bm e_t)\bm k_t^\top$, and
\begin{equation*}
  \langle\mathcal A(\mathbf H),\mathbf G\rangle_F
  =\bm e_t^\top\mathbf H^\top\mathbf G\bm k_t
  =\bigl\langle\mathbf H,(\mathbf G\bm k_t)\bm e_t^\top\bigr\rangle_F,
\end{equation*}
so $\mathcal A^*(\mathbf G)=(\mathbf G\bm k_t)\bm e_t^\top$. On a convex domain a $C^1$ field is a gradient field if and only if its Jacobian is self-adjoint, here if and only if $\mathbf H(\bm e_t\bm k_t^\top-\bm k_t\bm e_t^\top)=\mathbf 0$ for all $\mathbf H$, that is $\bm e_t\bm k_t^\top=\bm k_t\bm e_t^\top$. Right-multiplying by $\bm k_t$ gives $\bm e_t=\bm k_t(\bm e_t^\top\bm k_t)/\|\bm k_t\|_2^2$. Conversely $\bm e_t=c\bm k_t$ with $c\ne0$ gives $\mathbf F=\nabla_{\widetilde{\mathbf S}}\bigl[\tfrac c2\|\widetilde{\mathbf S}\bm k_t-\bm z_t/c\|_2^2\bigr]$, and $c=0$ gives a constant field.
\end{proof}

\begin{remark}[GDN-2 as a preconditioned gradient step]
\label{rem:gdn2-preconditioned}
Let $\bm b_t>\bm 0$, $\tilde{\bm k}_t=\bm b_t\odot\bm k_t$, and $\ell_t(\mathbf S)=\tfrac12\|\mathbf S\tilde{\bm k}_t-\bm z_t\|_2^2$, so $\nabla\ell_t(\widetilde{\mathbf S})=(\widetilde{\mathbf S}\tilde{\bm k}_t-\bm z_t)\tilde{\bm k}_t^\top$. Since $\tilde{\bm k}_t^\top\operatorname{Diag}(\bm b_t)^{-1}=\bm k_t^\top$,
\begin{equation}
  \nabla\ell_t(\widetilde{\mathbf S})\operatorname{Diag}(\bm b_t)^{-1}
  =(\widetilde{\mathbf S}\tilde{\bm k}_t-\bm z_t)\bm k_t^\top
  =\mathbf F(\widetilde{\mathbf S}),
  \label{eq:gdn2-preconditioned}
\end{equation}
so $\mathbf F$ is the gradient of $\ell_t$ in the key-weighted inner product $\langle\mathbf A,\mathbf B\rangle=\operatorname{tr}(\mathbf A\operatorname{Diag}(\bm b_t)\mathbf B^\top)$: GDN-2's token update is a right-preconditioned gradient step. After the chunk's token updates, the boundary decay is the separate proximal step characterized by Remark~\ref{rem:ridge}.
\end{remark}

DeltaNet, GDN, and KDA have $\bm e_t=\beta_t\bm k_t$ and $\bm z_t=\beta_t\bm v_t$, so Eq.~(\ref{eq:generic-token-update}) is $\widetilde{\mathbf S}-\beta_t\nabla\ell_t(\widetilde{\mathbf S})$ with $\ell_t$ from Eq.~(\ref{eq:delta-loss-preliminary}): an exact gradient step at each token. A chunk applies these local updates sequentially before the boundary proximal step; their composition is not generally one gradient step on a single loss. GDN-2 has $\bm e_t=\bm b_t\odot\bm k_t$ and meets the condition of Lemma~\ref{lem:gradient-step} only when $b_{t,j}$ is constant on $\{j:k_{t,j}\ne0\}$. When $\bm b_t>\bm 0$, Remark~\ref{rem:gdn2-preconditioned} instead gives a preconditioned-gradient interpretation. If an erase-gate entry is zero, $\operatorname{Diag}(\bm b_t)$ is singular and this positive-definite metric interpretation does not apply. Mamba-3's recurrence contains no read-and-correct term.

\subsection{Conditional shrinkage and excess risk}
\label{app:conditional-wiener-proof}

Section~\ref{sec:oracle-decay} states Lemma~\ref{lem:conditional-wiener} and Corollary~\ref{cor:sharing-cost}. This appendix gives their proofs, the readout connection, oracle-rate interpretation, and detailed conditional and projection calculations.

\paragraph{Isotropic readout risk.}
For the reconstruction risk defined in Section~\ref{sec:oracle-decay}, let $\widehat{\mathbf S}=\boldsymbol\Lambda\odot\mathbf S^-$ and take $\bm q$ independent of the reconstruction error, with $\mathbb E[\bm q\bm q^\top]=d_k^{-1}\mathbf I$. Then
\begin{equation}
  \mathbb E\bigl[\|(\widehat{\mathbf S}-\mathbf S^\star)\bm q\|_2^2\bigr]
  =\frac1{d_k}\,
   \mathbb E\bigl[\|\widehat{\mathbf S}-\mathbf S^\star\|_F^2\bigr].
  \label{eq:isotropic-query-risk}
\end{equation}
Other query models can favor different shrinkage; Appendix~\ref{app:readout-derivation} compares this criterion with querying a stored key.

Let $\Phi$ denote the information available to an oracle gate, and call a gate \emph{admissible} if it is a function of $\Phi$ with finite risk. The full boundary-state pair $\Phi=(\mathbf S_{[n]}^+,\mathbf S_{[n+1]}^-)$ provides one oracle reference. The implemented gate uses only $\Delta\mathbf S_{[n]}$ through Section~\ref{sec:gate-function}'s parameterization, so it belongs to a restricted subclass. The difference need not contain all the information in the pair.

\paragraph{Oracle rate under conditional uncorrelatedness.}
Under the additional condition $\mathbb E[\mathbf S^\star\odot\mathbf N\mid\Phi]=\mathbf 0$ in Lemma~\ref{lem:conditional-wiener}, the oracle retention on entries with $m_{\star,ab}>0$ and $m_{N,ab}>0$ satisfies
\begin{equation}
  \boldsymbol\kappa^\star=\mathbf m_N\oslash\mathbf m_\star,
  \qquad
  \boldsymbol\rho^\star
  =\log\!\left(\mathbf 1+\mathbf m_N\oslash\mathbf m_\star\right).
  \label{eq:optimal-penalty}
\end{equation}
Thus both the optimal ridge weight and decay rate are finite and strictly positive on these entries. If $m_{N,ab}=0$, the oracle is the identity gate with $\rho^\star_{ab}=\kappa^\star_{ab}=0$; if $m_{\star,ab}=0$, it is the zero gate, obtained only as $\rho_{ab},\kappa_{ab}\to\infty$. Because finite scores yield retention strictly between zero and one, these endpoint gates occur only as limits. This is an oracle identity for moments conditioned on $\Phi$, not a claim that the learned gate explicitly estimates those moments.

\paragraph{Proof of Lemma~\ref{lem:conditional-wiener}.}
Fix an entry $(a,b)$ and abbreviate $x=S^\star_{ab}$, $y=S^-_{ab}$, $\lambda=\Lambda_{ab}$, $m=\mathbb E[y^2\mid\Phi]>0$, and $c=\mathbb E[xy\mid\Phi]$. Because $\lambda$ is $\Phi$-measurable it factors out of the conditional expectation, so
\begin{equation}
  \mathbb E\bigl[(\lambda y-x)^2\mid\Phi\bigr]
  =\lambda^2m-2\lambda c+\mathbb E[x^2\mid\Phi]
  =m\Bigl(\lambda-\frac cm\Bigr)^{2}
   +\mathbb E[x^2\mid\Phi]-\frac{c^2}{m}.
  \label{eq:entrywise-completion}
\end{equation}
The conditional minimizer is $\lambda^\star=c/m$, which is Eq.~(\ref{eq:conditional-wiener-gate}) element-wise; summing over entries and taking expectations gives Eq.~(\ref{eq:conditional-excess}) with $\mathcal R(\boldsymbol\Lambda^\star)=\mathbb E\sum_{a,b}(\mathbb E[x^2\mid\Phi]-c^2/m)$. Under $\mathbb E[\mathbf S^\star\odot\mathbf N\mid\Phi]=\mathbf 0$ we get $c=\mathbf m_\star$ and $m=\mathbf m_\star+\mathbf m_N$, both non-negative with $m>0$, so the ratio lies in $[0,1]$. Conditional Cauchy--Schwarz gives $\mathbb E[m(\lambda^\star)^2]=\mathbb E[c^2/m]\le\mathbb E[x^2]<\infty$, so $\boldsymbol\Lambda^\star$ has finite weighted norm. \hfill$\square$

The noise-to-signal interpretation in Eq.~(\ref{eq:optimal-penalty}) requires conditional uncorrelatedness for the specified $\Phi$; unconditional orthogonality alone does not imply this condition.

\paragraph{Proof of Corollary~\ref{cor:sharing-cost}.}
Let $\mathcal H$ be the space of $\Phi$-measurable $d_v\times d_k$ random matrices with $\mathbb E\|\mathbf m_-^{1/2}\odot\boldsymbol\Lambda\|_F^2<\infty$, with inner product $\langle\mathbf A,\mathbf B\rangle_{\mathbf m}=\mathbb E\sum_{a,b}m_{-,ab}A_{ab}B_{ab}$ and induced norm $\|\cdot\|_{\mathbf m}$. By Lemma~\ref{lem:conditional-wiener}, $\mathcal R(\boldsymbol\Lambda)=\mathcal R(\boldsymbol\Lambda^\star)+\|\boldsymbol\Lambda-\boldsymbol\Lambda^\star\|_{\mathbf m}^2$, so minimizing $\mathcal R$ over a closed linear subspace is orthogonal projection of $\boldsymbol\Lambda^\star$ onto it. The constant families form nested finite-dimensional subspaces $\mathcal C_{\mathrm{scalar}}\subset\mathcal C_{\mathrm{key}}\subset\mathcal C_{\mathrm{elem}}\subset\mathcal H$, so their projections compose and
\begin{equation}
  \|\boldsymbol\Lambda^\star-\Pi_{\mathrm{scalar}}\boldsymbol\Lambda^\star\|_{\mathbf m}^2
  =\|\boldsymbol\Lambda^\star-\Pi_{\mathrm{elem}}\boldsymbol\Lambda^\star\|_{\mathbf m}^2
  +\|\Pi_{\mathrm{elem}}\boldsymbol\Lambda^\star-\Pi_{\mathrm{key}}\boldsymbol\Lambda^\star\|_{\mathbf m}^2
  +\|\Pi_{\mathrm{key}}\boldsymbol\Lambda^\star-\Pi_{\mathrm{scalar}}\boldsymbol\Lambda^\star\|_{\mathbf m}^2,
  \label{eq:nested-pythagoras}
\end{equation}
because the first difference is orthogonal to $\mathcal C_{\mathrm{elem}}$, which contains the other two, and the second is orthogonal to $\mathcal C_{\mathrm{key}}$, which contains the third. For a constant $\mathbf c$, $\|\boldsymbol\Lambda^\star-\mathbf c\|_{\mathbf m}^2=\sum_{a,b}\mathbb E[m_{-,ab}(\Lambda^\star_{ab}-c_{ab})^2]$ is minimized at
\begin{equation*}
  c_{ab}
  =\frac{\mathbb E[m_{-,ab}\Lambda^\star_{ab}]}{\bar m_{ab}}
  =\frac{\mathbb E\bigl[\mathbb E[S^\star_{ab}S^-_{ab}\mid\Phi]\bigr]}{\bar m_{ab}}
  =\bar\Lambda_{ab},
\end{equation*}
using $\mathbf m_-\odot\boldsymbol\Lambda^\star=\mathbb E[\mathbf S^\star\odot\mathbf S^-\mid\Phi]$ and the tower property. Restricted to constants the inner product is $\sum_{a,b}\bar m_{ab}A_{ab}B_{ab}$, so projecting $\bar{\boldsymbol\Lambda}$ onto row-constant matrices gives $\mu_b$ and onto multiples of $\mathbf 1$ gives $\mu$; the two expressions for $\mu$ agree because $\sum_b\bar m_{\cdot b}\mu_b=\sum_{a,b}\bar m_{ab}\bar\Lambda_{ab}$. Substituting into Eq.~(\ref{eq:nested-pythagoras}) gives Eq.~(\ref{eq:sharing-cost}). Each projection is an $\bar{\mathbf m}$-weighted average of values of $\boldsymbol\Lambda^\star$, so the box constraint is inactive whenever $\boldsymbol\Lambda^\star\in[0,1]$ almost surely. \hfill$\square$

\begin{corollary}[A second chain: the cost of adaptive sharing]
\label{cor:two-chains}
Let $\mathcal C^{\mathrm{adapt}}_{\mathrm{key}}=\{\mathbf 1_{d_v}\bm\lambda(\Phi)^\top:\bm\lambda\ \Phi\text{-measurable}\}\cap\mathcal H$. Then $\mathcal C_{\mathrm{scalar}}\subset\mathcal C_{\mathrm{key}}\subset\mathcal C^{\mathrm{adapt}}_{\mathrm{key}}\subset\mathcal C_{\mathrm{adapt}}$ is a second chain whose middle projection is
\begin{equation*}
  \lambda_b(\Phi)
  =\frac{\sum_a m_{-,ab}\Lambda^\star_{ab}}{\sum_a m_{-,ab}}
  =\frac{\sum_a\mathbb E[S^\star_{ab}S^-_{ab}\mid\Phi]}
        {\sum_a\mathbb E[(S^-_{ab})^2\mid\Phi]},
\end{equation*}
and
\begin{equation}
  \min_{\mathcal C_{\mathrm{scalar}}}\mathcal R-\min_{\mathcal C_{\mathrm{adapt}}}\mathcal R
  =\underbrace{\mathbb E\!\sum_{a,b}m_{-,ab}\bigl(\Lambda^\star_{ab}-\lambda_b(\Phi)\bigr)^2}
    _{\mathcal E^{\mathrm{cond}}_{\mathrm{value}}}
  +\underbrace{\mathbb E\!\sum_{a,b}m_{-,ab}\bigl(\lambda_b(\Phi)-\mu_b\bigr)^2}
    _{\mathcal E^{\mathrm{key}}_{\mathrm{state}}}
  +\mathcal E_{\mathrm{key}} .
  \label{eq:two-chains}
\end{equation}
\end{corollary}

\begin{proof}
$\mathcal C^{\mathrm{adapt}}_{\mathrm{key}}$ is a closed linear subspace of $\mathcal H$ containing $\mathcal C_{\mathrm{key}}$, so the proof of Corollary~\ref{cor:sharing-cost} applies with $\mathcal C_{\mathrm{elem}}$ replaced by $\mathcal C^{\mathrm{adapt}}_{\mathrm{key}}$. Minimizing $\mathbb E\sum_{a,b}m_{-,ab}(\Lambda^\star_{ab}-\lambda_b)^2$ pointwise in $\Phi$ and $b$ gives the stated $\lambda_b(\Phi)$, and the second form follows from $\mathbf m_-\odot\boldsymbol\Lambda^\star=\mathbb E[\mathbf S^\star\odot\mathbf S^-\mid\Phi]$. The last term is the same $\mathcal E_{\mathrm{key}}$ because both chains project onto $\mathcal C_{\mathrm{key}}$ and then onto $\mathcal C_{\mathrm{scalar}}$.
\end{proof}

The two chains share $\mathcal E_{\mathrm{key}}$ but are not comparable term by term, since $\mathcal C_{\mathrm{elem}}$ and $\mathcal C^{\mathrm{adapt}}_{\mathrm{key}}$ are not nested: one fixes the gate while keeping element-wise freedom, the other allows adaptation to $\Phi$ while sharing across value rows. These families provide idealized counterparts to the variants in Table~\ref{tab:ablation}: element-wise \sdg belongs to a restricted subclass of $\mathcal C_{\mathrm{adapt}}$, channel-wise \sdg to a restricted subclass of $\mathcal C^{\mathrm{adapt}}_{\mathrm{key}}$, the learned static gate to $\mathcal C_{\mathrm{elem}}$, and fixed boundary decay to $\mathcal C_{\mathrm{scalar}}$. The excess-risk formulas compare the optima of these families under one fixed state distribution, not the trained variants themselves.

\begin{remark}[Bayes optimality under state dependence]
\label{rem:bayes}
Suppose $\mathbf S^-$ is $\Phi$-measurable, as for the full boundary-state oracle. Then $\mathbf m_-=\mathbf S^-\odot\mathbf S^-$ and $\boldsymbol\Lambda^\star=\mathbb E[\mathbf S^\star\mid\Phi]\oslash\mathbf S^-$, so
\begin{equation}
  \boldsymbol\Lambda^\star\odot\mathbf S^-=\mathbb E[\mathbf S^\star\mid\Phi],
  \label{eq:bayes-gate}
\end{equation}
the minimum mean-squared-error estimator of $\mathbf S^\star$ among \emph{all} $\Phi$-measurable estimators, not only multiplicative ones; $\mathcal R(\boldsymbol\Lambda^\star)$ is thus the Bayes risk. Restricting the gate to $(0,1]$ attains Eq.~(\ref{eq:bayes-gate}) only where $0<\mathbb E[S^\star_{ab}\mid\Phi]/S^-_{ab}\le1$, that is, where the posterior mean shrinks the observed entry without changing sign, whereas a constant gate is confined to linear shrinkage however informative $\Phi$ is. These are conditions for the oracle family. The implemented \sdg gate uses only $\Delta\mathbf S_{[n]}$, which need not determine $\mathbf S^-$, and its finite-score parameterization yields gates strictly between zero and one.
\end{remark}

\subsection{Superposition memory and two shrinkage criteria}
\label{app:superposition-example}
\label{app:readout-derivation}

\paragraph{Example: superposition memory.}
Let $\mathbf S^\star=\sum_{i\in\mathcal F}\bm v_i\bm k_i^\top$ and $\mathbf N=\sum_{j\in\mathcal O}\bm u_j\bm g_j^\top$ collect useful and interfering associations. Take independent keys uniform on the unit sphere. Values are zero-mean, independent across associations and of the keys, with $\mathbb E[v_{i,a}^2]=\sigma_a^2$ and $\mathbb E[u_{j,a}^2]=\tau_a^2$. The index sets are independent of both keys and values.
For this example, let the oracle observe only the counts, $\Phi_{\mathrm{cnt}}=(|\mathcal F|,|\mathcal O|)$. Assume $|\mathcal F|>0$ almost surely, $\mathbb E|\mathcal F|<\infty$, $0<\mathbb E|\mathcal O|<\infty$, and $0<\sigma_a^2,\tau_a^2<\infty$ for every value channel. These conditions make the following conditional moments and ratios well-defined. The optimal counts-only gate is
\begin{equation}
  \Lambda^{\star,\mathrm{cnt}}_{ab}
  =\frac{|\mathcal F|\sigma_a^2}
        {|\mathcal F|\sigma_a^2+|\mathcal O|\tau_a^2}
  =\Bigl(1+\tfrac{|\mathcal O|}{|\mathcal F|}\cdot\tfrac{\tau_a^2}{\sigma_a^2}\Bigr)^{-1}.
  \label{eq:superposition-optimal-gate}
\end{equation}
The following calculation derives this gate and the associated risk terms.

\paragraph{Element-wise moments.}
With $\mathbf S^\star=\sum_{i\in\mathcal F}\bm v_i\bm k_i^\top$ and $\mathbf N=\sum_{j\in\mathcal O}\bm u_j\bm g_j^\top$ as defined above, entry $(a,b)$ satisfies $S^\star_{ab}=\sum_{i\in\mathcal F}v_{i,a}k_{i,b}$. The values are independent across associations, zero-mean, and independent of the keys, so all cross terms vanish and
\begin{equation}
  \mathbb E\bigl[(S^\star_{ab})^2\bigm|\mathcal F,\mathcal O\bigr]
  =\sum_{i\in\mathcal F}\mathbb E[v_{i,a}^2]\,\mathbb E[k_{i,b}^2]
  =\frac{|\mathcal F|\sigma_a^2}{d_k},
  \qquad
  \mathbb E\bigl[N_{ab}^2\bigm|\mathcal F,\mathcal O\bigr]
  =\frac{|\mathcal O|\tau_a^2}{d_k},
  \label{eq:superposition-moments}
\end{equation}
using $\mathbb E[\bm k\bm k^\top]=d_k^{-1}\mathbf I$ for keys uniform on the unit sphere. The same independence gives $\mathbb E[S^\star_{ab}N_{ab}\mid\mathcal F,\mathcal O]=0$. Because the index sets are independent of the keys and values, conditioning only on their cardinalities preserves these moments and the zero cross moment. Lemma~\ref{lem:conditional-wiener} therefore yields Eq.~(\ref{eq:superposition-optimal-gate}). Both $\Lambda^{\star,\mathrm{cnt}}_{ab}$ and $\bar m_{ab}$ are independent of $b$, so Corollary~\ref{cor:sharing-cost} gives $\mathcal E_{\mathrm{key}}=0$ and
\begin{equation}
  \mathcal E_{\mathrm{value}}
  =d_k\sum_{a}\bar m_{a}\bigl(\bar\Lambda_{a}-\mu\bigr)^2,
  \qquad
  \bar m_a=\frac{\mathbb E|\mathcal F|\,\sigma_a^2+\mathbb E|\mathcal O|\,\tau_a^2}{d_k},
  \qquad
  \bar\Lambda_a=\Bigl(1+\tfrac{\mathbb E|\mathcal O|}{\mathbb E|\mathcal F|}\cdot\tfrac{\tau_a^2}{\sigma_a^2}\Bigr)^{-1},
  \label{eq:superposition-value-cost}
\end{equation}
Under the stated positivity assumptions, Eq.~(\ref{eq:superposition-value-cost}) is strictly positive exactly when $\tau_a^2/\sigma_a^2$ is not constant across value channels. Moreover, for every channel $(1+r\tau_a^2/\sigma_a^2)^{-1}$ is strictly decreasing in $r=|\mathcal O|/|\mathcal F|$. Since the conditional weights are positive, $\mathcal E_{\mathrm{state}}(\Phi_{\mathrm{cnt}})>0$ exactly when this count ratio is not almost surely constant.

These conclusions apply to the counts-only oracle. Conditioning on the realized state may change both the optimal gate and the conditional signal--noise cross moment, and we assume no information inclusion between $\Phi_{\mathrm{cnt}}$ and either the boundary-state pair or $\Delta\mathbf S_{[n]}$. More generally, if $\sigma(\Phi_1)\subseteq\sigma(\Phi_2)$ under the same state distribution, then
\begin{equation*}
  \mathcal E_{\mathrm{state}}(\Phi_1)
  \leq \mathcal E_{\mathrm{state}}(\Phi_2),
\end{equation*}
because the larger admissible family cannot have greater minimum risk, whereas the best constant element-wise risk is unchanged. No such ordering follows without nested information.

\paragraph{Readout-level shrinkage and why the two criteria differ.}
A complementary calculation places the criterion at the readout rather than the state. Take $m>1$ independent keys uniform on the unit sphere, values independent, zero-mean, and independent of the keys with $\mathbb E[\bm v_j\bm v_j^\top]=\sigma_v^2\mathbf I$, and query with a stored key $\bm q=\bm k_i$:
\begin{equation}
  \bm y=\mathbf S\bm k_i=\bm x+\bm\varepsilon,
  \qquad \bm x=\bm v_i,
  \qquad \bm\varepsilon=\sum_{j\ne i}\bm v_j(\bm k_j^\top\bm k_i).
  \label{eq:vector-readout}
\end{equation}
Conditionally on $\bm k_i$, each $z_j=\bm k_j^\top\bm k_i$ has mean zero and second moment $1/d_k$, so
\begin{equation}
  \mathbb E\|\bm x\|_2^2=d_v\sigma_v^2,
  \qquad
  \mathbb E\|\bm\varepsilon\|_2^2=\frac{m-1}{d_k}d_v\sigma_v^2,
  \qquad
  \mathbb E[\bm x^\top\bm\varepsilon]=0,
  \label{eq:vector-readout-moments}
\end{equation}
giving $\mathrm{SNR}_{\mathrm{readout}}=d_k/(m-1)$ and, by the same completion of the square, the scalar minimizer
\begin{equation}
  \lambda^\star_{\mathrm{readout}}
  =\frac{d_k}{d_k+m-1}
  =\frac{\mathrm{SNR}_{\mathrm{readout}}}{1+\mathrm{SNR}_{\mathrm{readout}}}.
  \label{eq:scalar-wiener-readout}
\end{equation}
The corresponding state-level quantity for the same memory is different. With one target and $m-1$ interfering associations and $\sigma_a=\tau_a=\sigma_v$, Eq.~(\ref{eq:superposition-moments}) gives $\Lambda^\star_{ab}=1/m$ uniformly, whereas Eq.~(\ref{eq:scalar-wiener-readout}) is close to one for $m\ll d_k$. The discrepancy is entirely a difference of query model, not an inconsistency: Eq.~(\ref{eq:isotropic-query-risk}) weights all $d_k$ directions equally, while Eq.~(\ref{eq:vector-readout}) uses a query aligned with the target's key and therefore amplifies the target by a factor $d_k$ relative to the isotropic average. We report the state-level criterion in the main text because it is separable across entries, which is what makes element-wise gate families directly comparable; a general query covariance couples entries and destroys this separability. Neither calculation characterizes the capacity of learned linear attention.

\paragraph{Isotropic future queries.}
For Eq.~(\ref{eq:isotropic-query-risk}), condition on $\mathbf A=\widehat{\mathbf S}-\mathbf S^\star$. Independence and isotropy of $\bm q$ give
\begin{equation}
  \mathbb E_{\bm q}\|\mathbf A\bm q\|_2^2
  =\operatorname{tr}\bigl(\mathbf A^\top\mathbf A\,\mathbb E[\bm q\bm q^\top]\bigr)
  =\frac1{d_k}\|\mathbf A\|_F^2 .
  \label{eq:query-trace-identity}
\end{equation}
Taking the remaining expectation proves the identity.

\section{State propagation and value-row structure}
\label{app:operator-view}

\paragraph{Token-wise gates and shared transitions.}
A direct extension of GDN or KDA replaces the token-wise scalar or diagonal decay with a full matrix gate:
\begin{equation}
  \mathbf S_t
  =(\boldsymbol\Lambda_t\odot\mathbf S_{t-1})
  (\mathbf I-\beta_t\bm k_t\bm k_t^\top)
  +\beta_t\bm v_t\bm k_t^\top,
  \qquad
  \boldsymbol\Lambda_t\in(0,1]^{d_v\times d_k}.
  \label{eq:tokenwise-matrix-gate}
\end{equation}
This recurrence is well-defined, but its chunkwise computation differs from that of the base layer. In KDA, every value row shares the same key-space transition. At position $r$ within a chunk, its homogeneous transition is
\begin{equation}
  \mathbf P_r
  =\prod_{i=1}^{r}\mathbf D_i(\mathbf I-\beta_i\bm k_i\bm k_i^\top),
  \label{eq:shared-chunk-transition}
\end{equation}
with earlier factors on the left. The chunk index is suppressed; $\mathbf P_C$ is $\mathbf P_{[n]}$ in Eq.~(\ref{eq:preliminary-affine-chunk}). One set of WY factors therefore serves all value channels~\citep{yang2024gdn,kimi2025linear}.

For a general element-wise gate, let $\bm\lambda_i^{(a)}$ denote row $a$ of $\boldsymbol\Lambda_i$. Given the realized token gates, row $a$ has homogeneous transition
\begin{equation}
  \mathbf P_r^{(a)}
  =\prod_{i=1}^{r}
  \operatorname{Diag}(\bm\lambda_i^{(a)})
  (\mathbf I-\beta_i\bm k_i\bm k_i^\top).
  \label{eq:value-dependent-chunk-transition}
\end{equation}
These transitions generally differ across value rows, requiring up to $d_v$ sets of factors instead of the shared set used by existing kernels. If the gates also read the running state, the factors cannot be prepared from token inputs alone. Equation~(\ref{eq:value-dependent-chunk-transition}) describes propagation along the realized gate path, not the Jacobian of the full state-dependent recurrence. Boundary placement retains the native shared transitions within each chunk.

\paragraph{Boundary-gated state propagation.}
This section unrolls the boundary-gated recurrence into a temporal kernel over past writes and shows that this kernel is diagonal: it scales each value coordinate separately rather than mixing them.
\begin{equation}
  \mathbf S_{[n+1]}^-
  =
  \mathbf S_{[n]}^+\mathbf P_{[n]}+\boldsymbol\Omega_{[n]},
  \label{eq:affine-chunk-transition}
\end{equation}
where $\mathbf P_{[n]}\in\mathbb R^{d_k\times d_k}$ summarizes the within-chunk transition and $\boldsymbol\Omega_{[n]}$ collects writes. Vectorize the state column-wise and let
\begin{equation}
  \mathbf D_{[n]}
  =
  \operatorname{Diag}(\operatorname{vec}\boldsymbol\Lambda_{[n]}),
  \qquad
  \mathbf T_{[n]}
  =
  \mathbf D_{[n]}
  (\mathbf P_{[n]}^\top\otimes\mathbf I_{d_v}).
  \label{eq:vectorized-transition}
\end{equation}
Conditional on the realized gate trajectory,
\begin{equation}
  \operatorname{vec}(\mathbf S_{[n+1]}^+)
  =
  \mathbf T_{[n]}\operatorname{vec}(\mathbf S_{[n]}^+)
  +
  \mathbf D_{[n]}\operatorname{vec}(\boldsymbol\Omega_{[n]}),
  \label{eq:vectorized-sdg}
\end{equation}
and unrolling gives
\begin{equation}
  \operatorname{vec}(\mathbf S_{[n+1]}^+)
  =
  \left(\prod_{\ell=1}^{n}\mathbf T_{[\ell]}\right)
  \operatorname{vec}(\mathbf S_{[1]}^+)
  +
  \sum_{r=1}^{n}
  \left(\prod_{\ell=r+1}^{n}\mathbf T_{[\ell]}\right)
  \mathbf D_{[r]}\operatorname{vec}(\boldsymbol\Omega_{[r]}),
  \label{eq:conditional-unroll}
\end{equation}
with later transitions ordered on the left. With the readout $\bm o_t=(\bm q_t^\top\otimes\mathbf I_{d_v})\operatorname{vec}(\mathbf S_t)$, a zero initial state, and the realized transition and gate factors held fixed, Eq.~(\ref{eq:conditional-unroll}) expresses the output as a temporal kernel applied to past writes,
\begin{equation}
  \bm o_t=\sum_{j\leq t}\boldsymbol{\mathcal K}_{t,j}\bm z_j,
  \qquad
  \boldsymbol{\mathcal K}_{t,j}\in\mathbb R^{d_v\times d_v},
  \label{eq:operator-valued-kernel}
\end{equation}
where $\bm z_j$ is the write vector of token $j$: $\bm w_j\odot\bm v_j$ for GDN-2, $\beta_j\bm v_j$ for GDN and KDA, and a trapezoidal combination of $\bm v_{j-1}$ and $\bm v_j$ for Mamba-3. This kernel is not an arbitrary operator.

\begin{lemma}[Value-row decoupling]
\label{lem:row-decoupling}
Suppose the base recurrence has the form
\begin{equation*}
  \mathbf S_t=\mathbf S_{t-1}\mathbf A_t
  +\sum_{\ell=1}^{L}\bm z_t^{(\ell)}\bigl(\bm k_t^{(\ell)}\bigr)^\top,
\end{equation*}
with $\mathbf A_t\in\mathbb R^{d_k\times d_k}$ and all $\bm k_t^{(\ell)}$ determined by token-local quantities. This covers Eqs.~(\ref{eq:delta-preliminary})--(\ref{eq:gdn2-preliminary}) with $L=1$ and $\mathbf A_t=\mathbf D_t(\mathbf I-\bm e_t\bm k_t^\top)$, and Eq.~(\ref{eq:mamba3-trapezoid-recurrence}) with $L=2$ and $\mathbf A_t=\alpha_t\mathbf I$. Add the boundary update of Eq.~(\ref{eq:backbone-sdg-insertion}). Then for every $a$, row $a$ of the state follows a recurrence involving only row $a$ of the initial state, the shared key-side factors $\{\mathbf A_t,\bm k_t^{(\ell)}\}$, the scalar sources $\{z^{(\ell)}_{t,a}\}$, and row $a$ of $\mathbf G$. Consequently the kernel of Eq.~(\ref{eq:operator-valued-kernel}) is diagonal,
\begin{equation}
  \boldsymbol{\mathcal K}_{t,j}
  =\operatorname{Diag}\bigl(c^{(1)}_{t,j},\ldots,c^{(d_v)}_{t,j}\bigr).
  \label{eq:diagonal-kernel}
\end{equation}
\end{lemma}

\begin{proof}
Row $a$ of the update is $\bm s_{t-1,a}^\top\mathbf A_t+\sum_\ell z^{(\ell)}_{t,a}(\bm k^{(\ell)}_t)^\top$, because right multiplication acts on the key index only and each outer product contributes to row $a$ alone. At a boundary, row $a$ is multiplied element-wise by row $a$ of $\boldsymbol\Lambda_{[n]}$, and row $a$ of $\mathbf Z_{[n]}=\mathbf G\odot(\Delta\mathbf S_{[n]}\mathbf W_S^\top)$ equals $\bm g_a\odot(\mathbf W_S\Delta\bm s_a)$, which depends only on row $a$ of $\Delta\mathbf S_{[n]}$. Both steps preserve the property, so by induction over tokens and boundaries row $a$ never depends on row $a'\ne a$. Since $\bm o_t=\mathbf S_t\bm q_t$, coordinate $a$ of $\bm o_t$ is a linear functional of row $a$ alone, which is Eq.~(\ref{eq:diagonal-kernel}).
\end{proof}

In the base models $c^{(a)}_{t,j}$ does not depend on $a$, so the kernel is $c_{t,j}\mathbf I_{d_v}$: one weight shared by all value coordinates. An element-wise boundary gate makes it a diagonal operator with value-channel-specific entries. Two scopes differ here. Diagonality is \emph{structural} and holds for every realization, because row $a$ of the gate reads only row $a$ of the state change. The \emph{linear} representation in Eq.~(\ref{eq:operator-valued-kernel}) is conditional on the realized gate path: $\mathbf D_{[n]}$ depends on the evolving state, so the coefficients are themselves functions of past values and cannot be precomputed from positions or reduced to one input-independent attention matrix.

Lemma~\ref{lem:row-decoupling} also justifies two implementation choices. Within a chunk $\mathbf A_t$ is shared across value rows, since \sdg acts only at boundaries, so one set of WY factors still serves all value channels---the requirement identified in Eqs.~(\ref{eq:shared-chunk-transition})--(\ref{eq:value-dependent-chunk-transition}). Across value rows the augmented recurrence is embarrassingly parallel, so the state scan may be tiled along $d_v$ with no cross-tile communication, while each tile holds full rows along $d_k$ because $\mathbf W_S$ mixes key channels.

\section{Kernel and backward details}
\label{app:kernel-details}

\paragraph{Boundary gate and implementation.}
For a fixed base rate $\rho_0>0$, define
\begin{equation*}
  \operatorname{squareplus}(z;\rho_0^2)
  =\tfrac12\bigl(z+\sqrt{z^2+\rho_0^2}\bigr),
  \qquad
  \rho(z)=2\operatorname{squareplus}(z;\rho_0^2).
\end{equation*}
This is the Squareplus parameterization~\citep{barron2021squareplus} in Eq.~(\ref{eq:boundary-rate}). The rate is smooth and increasing, with value $\rho_0$ and derivative one at zero. If the state is unchanged, then $\Delta\mathbf S_{[n]}=\mathbf0$ and the gate is $e^{-\rho_0}\mathbf1$.

\paragraph{Forward pass.}
We implement \sdg in Triton~\citep{tillet2019triton}, building on Flash Linear Attention~\citep{yang2024fla}. The boundary update is fused into the inter-chunk state scan; native factor preparation and output kernels are retained. On supported fused paths, each program holds a state tile in SRAM or registers, forms the difference in FP32, and projects it using Tensor Cores. The projected difference, score, rate, and gate remain on chip, avoiding HBM writes for these forward intermediates. Appendix~\ref{app:backbone-details} gives backbone-specific details and supported configurations.

\paragraph{Backward pass.}
The backward pass traverses the state scan in reverse order~\citep{hatamizadeh2026gdn2}. Differentiating the gate produces gradients for both the raw state and the incoming state. The former enters the native chunk backward computation; the latter is added directly to its incoming-state gradient. We save or recompute boundary states, recompute gate intermediates on chip, and accumulate the gain gradient during the scan. A separate reduction computes the projection gradient from saved gradients of $\mathbf X_{[n]}$. The following subsections give the gate VJP and additive reference forward and backward algorithms.

\subsection{Gate-function derivatives}

For one boundary, write $\mathbf S^{\src}$ for the state at the preceding boundary and $\mathbf S^{\mathrm{out}}$ for the gated result, and define
\begin{equation}
  \begin{aligned}
    \Delta\mathbf S&=\mathbf S^--\mathbf S^{\src},
    &\mathbf X&=\Delta\mathbf S\mathbf W_S^\top,\\
    \mathbf Z&=\mathbf G\odot\mathbf X,
    &\boldsymbol\rho&=\mathbf Z+
       \sqrt{\mathbf Z\odot\mathbf Z+\rho_0^2\mathbf 1},\\
    \boldsymbol\Lambda&=\exp(-\boldsymbol\rho),
    &\mathbf S^{\mathrm{out}}&=\boldsymbol\Lambda\odot\mathbf S^-.
  \end{aligned}
  \label{eq:gate-vjp-setup}
\end{equation}
For a tensor $\mathbf T$, write $\mathrm{d}\mathbf T=\partial\mathcal L/\partial\mathbf T$. Given $\mathrm{d}\mathbf S^{\mathrm{out}}$, the gate vector--Jacobian product computes
\begin{equation}
  \mathrm{d}\boldsymbol\rho
  =-\mathrm{d}\mathbf S^{\mathrm{out}}\odot\mathbf S^{\mathrm{out}},
  \qquad
  \mathrm{d}\mathbf Z
  =\mathrm{d}\boldsymbol\rho\odot
    \frac{\boldsymbol\rho}
    {\sqrt{\mathbf Z\odot\mathbf Z+\rho_0^2\mathbf 1}},
  \qquad
  \mathrm{d}\mathbf X=\mathrm{d}\mathbf Z\odot\mathbf G,
  \label{eq:gate-vjp-rate}
\end{equation}
with element-wise division, and then
\begin{equation}
  \mathrm{d}\mathbf G
  =\mathrm{d}\mathbf Z\odot\mathbf X,
  \qquad
  \mathrm{d}\mathbf W_S
  =\mathrm{d}\mathbf X^{\top}\Delta\mathbf S,
  \label{eq:gate-vjp-parameters}
\end{equation}
summed over chunks and batch examples. The state gradients are
\begin{align}
  \mathrm{d}\mathbf S^-
  &=
  \mathrm{d}\mathbf S^{\mathrm{out}}\odot\boldsymbol\Lambda
  +\mathrm{d}\mathbf X\mathbf W_S,
  \label{eq:gate-vjp-result}\\
  \mathrm{d}\mathbf S^{\src}
  &=
  -\mathrm{d}\mathbf X\mathbf W_S,
  \label{eq:gate-vjp-source}
\end{align}
the two gate-path terms having opposite signs because of the subtraction in $\Delta\mathbf S$, and $\mathrm{d}\mathbf S^{\src}$ is added to the gradient returned by the base chunk recurrence. This gives a consistency check used in our tests: for standard linear attention $\Delta\mathbf S_{[n]}=\mathbf V_{[n]}^\top\mathbf K_{[n]}$ is independent of the incoming state, so the two terms must cancel exactly once accumulated into $\mathrm{d}\mathbf S^+_{[n]}$.

For stable evaluation, the kernel computes $\sqrt{z^2+\rho_0^2}$ with a scaled norm. When $z<0$, it evaluates the equivalent expression $\rho_0^2/(\sqrt{z^2+\rho_0^2}-z)$ to avoid cancellation. The derivative is $\rho(z)/\sqrt{z^2+\rho_0^2}$, where $\rho(z)=z+\sqrt{z^2+\rho_0^2}$. This expression is positive on both sides of zero and equals one at zero.

\subsection{Additive reference forward pass}
\label{app:sdg-additive-forward}

Algorithm~\ref{alg:sdg-additive-forward} expands Algorithm~\ref{alg:sdg-forward} for unnormalized additive linear attention. Its on-chip steps illustrate the arithmetic rather than a common kernel launch schedule for all backbones. In this special case, $\Delta\mathbf S_{[n]}=\mathbf V_{[n]}^\top\mathbf K_{[n]}$ is independent of the incoming state for fixed chunk inputs. The gated recurrent layers evaluated here retain their native state transitions, so their state differences also depend on the incoming memory. This example assumes complete chunks, $L=NC$; arbitrary lengths and cached continuations follow Appendix~\ref{app:reference-boundary-scan}, where incomplete final chunks do not trigger boundary decay.

\begin{minipage}{\linewidth}
  \vspace{0.2em}

  \refstepcounter{algorithm}
  \label{alg:sdg-additive-forward}
  \small
  \hrule height 1.1pt
  \vspace{0.25em}
  \noindent\textbf{Algorithm \thealgorithm}\quad \textsc{SDG}: Illustrative Forward Pass for Additive Linear Attention
  \vspace{0.2em}
  \hrule
  \vspace{0.3em}

  \noindent\textbf{Require:} $L=NC$ and $N=L/C$ complete chunks; $\mathbf Q,\mathbf K\in\mathbb R^{L\times d_k}$, $\mathbf V\in\mathbb R^{L\times d_v}$, initial state $\mathbf S_{[1]}^+\in\mathbb R^{d_v\times d_k}$, chunk size $C$, and gate parameters $\mathbf W_S,\mathbf G$ with fixed base rate $\rho_0$.

  \setcounter{sdgalgorithmline}{0}
  \sdgalgline{Partition $\mathbf Q,\mathbf K,\mathbf V$ into $N$ complete chunks of $C$ tokens, $\{\mathbf Q_{[n]},\mathbf K_{[n]},\mathbf V_{[n]}\}_{n=1}^{N}$.}
  \sdgalgline{Load $\mathbf S_{[1]}^+$ from HBM to SRAM and construct the causal mask $\mathbf M_C$ on chip.}
  \sdgalgline{\bfseries Load $\mathbf W_S,\mathbf G$ from HBM to SRAM.}
  \sdgalgline{\textbf{for} $n=1,\ldots,N$ \textbf{do}}
  \sdgalgline[1.2em]{Load $\mathbf Q_{[n]},\mathbf K_{[n]},\mathbf V_{[n]}$ from HBM to SRAM.}
  \sdgalgline[1.2em]{On chip, compute $\mathbf O_{[n]}= \mathbf Q_{[n]}(\mathbf S_{[n]}^+)^\top +(\mathbf Q_{[n]}\mathbf K_{[n]}^\top\odot\mathbf M_C)\mathbf V_{[n]}$.}
  \sdgalgline[1.2em]{On chip, compute $\mathbf S_{[n+1]}^- =\mathbf S_{[n]}^++\mathbf V_{[n]}^\top\mathbf K_{[n]}$.}
  \sdgalgline[1.2em]{\bfseries On chip, compute $\Delta\mathbf S_{[n]}= \mathbf S_{[n+1]}^--\mathbf S_{[n]}^+$.}
  \sdgalgline[1.2em]{\bfseries On chip, compute $\mathbf X_{[n]}=\Delta\mathbf S_{[n]}\mathbf W_S^\top, \quad\mathbf Z_{[n]}=\mathbf G\odot\mathbf X_{[n]}$.}
  \sdgalgline[1.2em]{\bfseries On chip, compute $\boldsymbol\Lambda_{[n]}= \exp\!\left[-2\operatorname{squareplus}(\mathbf Z_{[n]};\rho_0^2) \right]$.}
  \sdgalgline[1.2em]{\bfseries On chip, compute $\mathbf S_{[n+1]}^+= \boldsymbol\Lambda_{[n]}\odot\mathbf S_{[n+1]}^-$.}
  \sdgalgline[1.2em]{Store $\mathbf O_{[n]}$ from SRAM to HBM.}
  \sdgalgline{\textbf{end for}}
  \sdgalgline{Store $\mathbf S_{[N+1]}^+$ from SRAM to HBM.}
  \sdgalgline{\textbf{return} $\mathbf O=\operatorname{Concat} (\mathbf O_{[1]},\ldots,\mathbf O_{[N]})$ and $\mathbf S_{[N+1]}^+$.}
  \vspace{0.3em}
  \hrule height 1.1pt
\end{minipage}

\subsection{Additive reference backward pass}
\label{app:sdg-backward-algorithm}

Algorithm~\ref{alg:sdg-backward} gives the reverse chunk scan for the additive example in Algorithm~\ref{alg:sdg-additive-forward}. The prefix $\mathrm{d}$ denotes the gradient of the loss with respect to a tensor. The quantities $\mathrm{d}\mathbf S_{[n+1]}^-$ and $\mathrm{d}\mathbf S_{[n]}^{\src}$ are the gradients returned by the boundary update with respect to $\mathbf S_{[n+1]}^-$ and $\mathbf S_{[n]}^+$. They follow Eqs.~(\ref{eq:gate-vjp-rate})--(\ref{eq:gate-vjp-source}). The scan saves $\mathrm{d}\mathbf X_{[n]}$ for a separate tiled reduction of the projection gradient.

\begin{minipage}{\linewidth}
  \refstepcounter{algorithm}
  \label{alg:sdg-backward}
  \small
  \hrule height 1.1pt
  \vspace{0.25em}
  \noindent\textbf{Algorithm \thealgorithm}\quad \textsc{SDG}: Fused Backward Pass for Standard Linear Attention
  \vspace{0.2em}
  \hrule
  \vspace{0.3em}

  \noindent\textbf{Require:} $\mathbf Q,\mathbf K,\mathbf V$, output gradients $\mathrm{d}\mathbf O$, saved or recomputed boundary states, and $\mathrm{d}\mathbf S_{[N+1]}^+$, with $\mathbf W_S,\mathbf G,\rho_0$.

  \setcounter{sdgalgorithmline}{0}
  \sdgalgline{Initialize parameter gradients to zero and construct $\mathbf M_C$ on chip.}
  \sdgalgline{\bfseries Load $\mathbf W_S,\mathbf G$ from HBM to SRAM.}
  \sdgalgline{\textbf{for} $n=N,\ldots,1$ \textbf{do}}
  \sdgalgline[1.2em]{Load $\mathbf Q_{[n]},\mathbf K_{[n]},\mathbf V_{[n]}$, $\mathrm{d}\mathbf O_{[n]}$, $\mathbf S_{[n]}^+$, and $\mathbf S_{[n+1]}^-$ from HBM to SRAM.}
  \sdgalgline[1.2em]{\bfseries On chip, compute $(\mathrm{d}\mathbf S_{[n+1]}^-, \mathrm{d}\mathbf S_{[n]}^{\src}) = \operatorname{VJP}_{\mathrm{Eqs.}\,(\ref{eq:projected-state-change})\text{--} (\ref{eq:canonical-sdg})} (\mathrm{d}\mathbf S_{[n+1]}^+)$.}
  \sdgalgline[1.2em]{\bfseries On chip, accumulate $\mathrm{d}\mathbf G\mathrel{+}= \mathrm{d}\mathbf Z_{[n]}\odot\mathbf X_{[n]}$.}
  \sdgalgline[1.2em]{\bfseries Store $\mathrm{d}\mathbf X_{[n]}$ from SRAM to HBM.}
  \sdgalgline[1.2em]{On chip, compute $\mathbf A_{qk,[n]}= \mathbf Q_{[n]}\mathbf K_{[n]}^\top\odot\mathbf M_C$ and $\mathrm{d}\mathbf A_{qk,[n]}= \mathrm{d}\mathbf O_{[n]}\mathbf V_{[n]}^\top$.}
  \sdgalgline[1.2em]{On chip, compute $\mathrm{d}\mathbf Q_{[n]}= \mathrm{d}\mathbf O_{[n]}\mathbf S_{[n]}^+ +(\mathrm{d}\mathbf A_{qk,[n]}\odot\mathbf M_C)\mathbf K_{[n]}$.}
  \sdgalgline[1.2em]{On chip, compute $\mathrm{d}\mathbf K_{[n]}= (\mathrm{d}\mathbf A_{qk,[n]}\odot\mathbf M_C)^\top\mathbf Q_{[n]} +\mathbf V_{[n]}\mathrm{d}\mathbf S_{[n+1]}^-$.}
  \sdgalgline[1.2em]{On chip, compute $\mathrm{d}\mathbf V_{[n]}= \mathbf A_{qk,[n]}^\top\mathrm{d}\mathbf O_{[n]} +\mathbf K_{[n]}(\mathrm{d}\mathbf S_{[n+1]}^-)^\top$.}
  \sdgalgline[1.2em]{On chip, accumulate $\mathrm{d}\mathbf S_{[n]}^+= \mathrm{d}\mathbf S_{[n+1]}^- +\mathrm{d}\mathbf O_{[n]}^\top\mathbf Q_{[n]} +\mathrm{d}\mathbf S_{[n]}^{\src}$.}
  \sdgalgline[1.2em]{Store $\mathrm{d}\mathbf Q_{[n]}$, $\mathrm{d}\mathbf K_{[n]}$, and $\mathrm{d}\mathbf V_{[n]}$ to HBM.}
  \sdgalgline{\textbf{end for}}
  \sdgalgline{\bfseries Load tiles of $\mathrm{d}\mathbf X_{[n]}$, $\mathbf S_{[n]}^+$, and $\mathbf S_{[n+1]}^-$ from HBM to SRAM.}
  \sdgalgline{\bfseries On chip, compute $\mathrm{d}\mathbf W_S= \sum_{n=1}^N\mathrm{d}\mathbf X_{[n]}^\top (\mathbf S_{[n+1]}^--\mathbf S_{[n]}^+)$ by tiled reduction.}
  \sdgalgline{\textbf{return} input, initial-state, and gate-parameter gradients.}
  \vspace{0.3em}
  \hrule height 1.1pt
\end{minipage}

\subsection{Fused kernel implementation}

\paragraph{GDN.}
The adapter retains GDN's native projections, token gates, WY preparation, intra-chunk interaction, and output path. The fused state scan applies the boundary gate after the raw state of a complete chunk has been formed. Its backward pass either reads saved boundary states or recomputes them according to the configured path.

\paragraph{GDN-2.}
The native channel-wise WY preparation, intra-chunk interaction, and output kernel are retained. A persistent state program consumes the native transition factors, holds one key/value state tile, computes the boundary branch after a complete chunk, and writes the gated state. Its custom backward mirrors the state scan and can use saved boundary states or recomputation according to the configured path.

\paragraph{KDA.}
The adapter retains KDA's gate, intra-chunk, output, and local backward kernels. It reuses the boundary state scan where the WY state interface matches GDN-2, while preserving the KDA-specific token transition.

\paragraph{Mamba-3.}
The boundary branch is integrated into the native SSM matrix-state scan and corresponding backward pass. Rotary phase, trapezoidal quantities, output gates, and auxiliary state recurrences remain unchanged. The one-token decode path advances a device-resident counter. Inactive rows perform no matrix-sized boundary work, while rows completing a $C$-token interval update the matrix and source cache.

\subsection{Complexity and saved state}

For sequence lengths $L_n$, the exact number of newly completed boundaries depends on incoming tail counters $r_n$ and is
\begin{equation}
  N_{\mathrm{bnd}}
  =
  \sum_n
  \left\lfloor
    \frac{r_n+L_n}{C}
  \right\rfloor.
  \label{eq:number-boundaries}
\end{equation}
For linear attention layers, forward gate-function work is $\Theta(N_{\mathrm{bnd}}H d_vd_k^2)$ for the projection and $\Theta(N_{\mathrm{bnd}}H d_vd_k)$ for the element-wise operations. For Mamba-3, replace $d_k$ by $d_s$. A decode cache adds one source matrix and one integer counter per enabled layer and sequence. Training stores or recomputes boundary states. The backward scan also writes $\mathrm{d}\mathbf X_{[n]}$ to HBM for the projection-gradient reduction. This reduction has the same asymptotic arithmetic cost as the forward projection.

\begingroup
\setlength{\parskip}{3pt plus 1pt}
\setlength{\abovedisplayskip}{7pt plus 2pt minus 2pt}
\setlength{\belowdisplayskip}{7pt plus 2pt minus 2pt}
\setlength{\abovedisplayshortskip}{3pt plus 1pt}
\setlength{\belowdisplayshortskip}{4pt plus 1pt}
\titlespacing*{\section}{0pt}{1.5ex plus .5ex minus .2ex}{1ex plus .2ex}
\titlespacing*{\paragraph}{0pt}{1.5ex plus .3ex minus .2ex}{.8em}
\section{FLOPs analysis}
\label{app:flops}

\paragraph{Counting convention.}
We count a multiply--add as two floating-point operations (FLOPs), so a dense $(m\times k)(k\times n)$ product costs $2mkn$ FLOPs. Consider a sequence of length $L$, $H$ recurrent heads, state dimensions $d_v\times d_k$, and chunk size $C$. The counts below describe the additional gate computation in one \sdg-enabled layer; they exclude the unchanged base recurrence, input/output projections, and feed-forward network. We report the leading matrix-multiplication cost explicitly and group element-wise arithmetic into lower-order terms. Square roots and exponentials are evaluated once per state element in the forward gate; their hardware-dependent costs are not equated to those of an addition or multiplication.

\paragraph{Forward computation.}
At each completed boundary, the product $\Delta\mathbf S\mathbf W_S^\top$ costs $2Hd_vd_k^2$ FLOPs. Forming the state difference, applying the gain, evaluating the rate and gate, and scaling the state require $O(Hd_vd_k)$ additional element-wise work. For an aligned sequence with $C\mid L$, there are $L/C$ boundaries, giving
\begin{equation}
  F_{\mathrm{SDG,fwd}}
  =\frac{2LHd_vd_k^2}{C}
   +O\!\left(\frac{LHd_vd_k}{C}\right),
  \qquad
  \frac{F_{\mathrm{SDG,fwd}}}{L}
  =\frac{2Hd_vd_k^2}{C}
   +O\!\left(\frac{Hd_vd_k}{C}\right).
  \label{eq:sdg-forward-flops}
\end{equation}
For arbitrary lengths or cached continuations, replace $L/C$ by the number of completed boundaries: $\lfloor L/C\rfloor$ for a fresh sequence, or $N_{\mathrm{bnd}}$ from Eq.~(\ref{eq:number-boundaries}) for a batch with incoming tail counters. A partial final chunk does not trigger a boundary gate. Ordinary decoding steps perform no gate matrix multiplication; a step completing a boundary pays the full $2Hd_vd_k^2$ projection cost. Thus, amortized cost and individual boundary-step cost differ by a factor of $C$ for complete intervals. For the Mamba-3 matrix state, replace $d_k$ by $d_s$.

\paragraph{Backward computation and recomputation.}
The gate derivatives require two matrix products: $\mathrm{d}\mathbf X\mathbf W_S$ for the state gradients and $\mathrm{d}\mathbf X^\top\Delta\mathbf S$ for the projection gradient. Each costs $2Hd_vd_k^2$ FLOPs per boundary. The first product is shared by both state-gradient paths in Eqs.~(\ref{eq:gate-vjp-result}) and~(\ref{eq:gate-vjp-source}); it need not be evaluated twice. With saved gate intermediates, a forward and backward evaluation therefore costs $6Hd_vd_k^2$ projection FLOPs per boundary. Recomputing $\mathbf X$ once during backward, as in the fused gate path, adds another $2Hd_vd_k^2$. Element-wise derivatives and gain-gradient accumulation add lower-order work. Replaying the base recurrence to recover boundary states, further activation checkpointing, and the optimizer update are separate costs and are excluded from these projection counts.

\paragraph{Batch and model scaling.}
For a batch of $B$ sequences of length $L$ and $M$ \sdg-enabled layers with the same dimensions and chunk size, the leading added forward cost is $2BMLHd_vd_k^2/C$. The corresponding forward-plus-backward projection cost is $6BMLHd_vd_k^2/C$ with saved intermediates, or $8BMLHd_vd_k^2/C$ with one gate recomputation. For heterogeneous layers, sum each layer's count instead of multiplying by $M$. If $d_k=d_v=d$, the forward cost simplifies to
\begin{equation}
  F_{\mathrm{SDG,fwd}}^{\mathrm{model}}
  =\frac{2BMLHd^3}{C}
   +O\!\left(\frac{BMLHd^2}{C}\right).
  \label{eq:sdg-model-flops}
\end{equation}
Only layers using \sdg contribute to this additional cost, including in hybrid models.

\paragraph{Scaling and measured efficiency.}
For fixed head dimensions and $C$, the additional FLOPs grow linearly with the number of tokens, and their amortized value decreases as $1/C$. This does not make $C$ a cost-only tuning parameter: changing it also changes when the model applies decay. The relative arithmetic overhead is $F_{\mathrm{SDG}}/F_{\mathrm{base}}$ under the same counting scope, and cannot be inferred from the added parameter count alone because gate parameters are used once per boundary. Nor does a FLOP ratio predict a latency ratio. The gate runs in the sequential state scan; matrix size, Tensor Core utilization, register pressure, and memory traffic affect its runtime. Fusion avoids materializing forward gate intermediates in HBM, whereas backward recomputation and the projection-gradient reduction introduce additional work and traffic. The end-to-end throughput and latency measurements in Section~\ref{sec:experiments} complement these arithmetic counts.

\endgroup
\end{document}

%% file: math_commands.tex
\usepackage{amsmath,amsfonts,bm}

\def\eqref#1{equation~\ref{#1}}

\def\1{\bm{1}}

\DeclareMathAlphabet{\mathsfit}{\encodingdefault}{\sfdefault}{m}{sl}
\SetMathAlphabet{\mathsfit}{bold}{\encodingdefault}{\sfdefault}{bx}{n}

\DeclareMathOperator*{\argmin}{arg\,min}

%% file: main.bbl
\begin{thebibliography}{69}
\providecommand{\natexlab}[1]{#1}
\providecommand{\url}[1]{\texttt{#1}}
\expandafter\ifx\csname urlstyle\endcsname\relax
  \providecommand{\doi}[1]{doi: #1}\else
  \providecommand{\doi}{doi: \begingroup \urlstyle{rm}\Url}\fi

\bibitem[Arora et~al.(2023)Arora, Yang, Eyuboglu, Narayan, Hojel, Trummer, and
  R{\'e}]{arora2023datalakes}
Arora, S., Yang, B., Eyuboglu, S., Narayan, A., Hojel, A., Trummer, I., and
  R{\'e}, C.
\newblock Language models enable simple systems for generating structured views
  of heterogeneous data lakes.
\newblock \emph{arXiv preprint arXiv:2304.09433}, 2023.

\bibitem[Arora et~al.(2024{\natexlab{a}})Arora, Eyuboglu, Timalsina, Johnson,
  Poli, Zou, Rudra, and R{\'e}]{arora2023zoology}
Arora, S., Eyuboglu, S., Timalsina, A., Johnson, I., Poli, M., Zou, J.~Y.,
  Rudra, A., and R{\'e}, C.
\newblock Zoology: Measuring and improving recall in efficient language models.
\newblock In \emph{International conference on learning representations},
  volume 2024, pp.\  15664--15730, 2024{\natexlab{a}}.

\bibitem[Arora et~al.(2024{\natexlab{b}})Arora, Timalsina, Singhal, Spector,
  Eyuboglu, Zhao, Rao, Rudra, and R{\'e}]{arora2024jrt}
Arora, S., Timalsina, A., Singhal, A., Spector, B., Eyuboglu, S., Zhao, X.,
  Rao, A., Rudra, A., and R{\'e}, C.
\newblock Just read twice: closing the recall gap for recurrent language
  models.
\newblock \emph{arXiv preprint arXiv:2407.05483}, 2024{\natexlab{b}}.

\bibitem[Ba et~al.(2016)Ba, Hinton, Mnih, Leibo, and
  Ionescu]{ba2016fastweights}
Ba, J., Hinton, G.~E., Mnih, V., Leibo, J.~Z., and Ionescu, C.
\newblock Using fast weights to attend to the recent past.
\newblock \emph{Advances in neural information processing systems}, 29, 2016.

\bibitem[Bai et~al.(2024)Bai, Lv, Zhang, Lyu, Tang, Huang, Du, Liu, Zeng, Hou,
  et~al.]{bai2023longbench}
Bai, Y., Lv, X., Zhang, J., Lyu, H., Tang, J., Huang, Z., Du, Z., Liu, X.,
  Zeng, A., Hou, L., et~al.
\newblock Longbench: A bilingual, multitask benchmark for long context
  understanding.
\newblock In \emph{Proceedings of the 62nd annual meeting of the association
  for computational linguistics (volume 1: Long papers)}, pp.\  3119--3137,
  2024.

\bibitem[Barron(2021)]{barron2021squareplus}
Barron, J.
\newblock Squareplus: A softplus-like algebraic rectifier.
\newblock \emph{arXiv preprint arXiv:2112.11687}, 2021.

\bibitem[Beck et~al.(2024)Beck, P{\"o}ppel, Spanring, Auer, Prudnikova, Kopp,
  Klambauer, Brandstetter, and Hochreiter]{beck2024xlstm}
Beck, M., P{\"o}ppel, K., Spanring, M., Auer, A., Prudnikova, O., Kopp, M.,
  Klambauer, G., Brandstetter, J., and Hochreiter, S.
\newblock xlstm: Extended long short-term memory.
\newblock \emph{Advances in Neural Information Processing Systems},
  37:\penalty0 107547--107603, 2024.

\bibitem[Behrouz et~al.(2026{\natexlab{a}})Behrouz, Razaviyayn, Zhong, and
  Mirrokni]{behrouz2026miras}
Behrouz, A., Razaviyayn, M., Zhong, P., and Mirrokni, V.
\newblock It's all connected: A journey through test-time memorization,
  attentional bias, retention, and online optimization.
\newblock In \emph{International Conference on Learning Representations},
  volume 2026, pp.\  131306--131333, 2026{\natexlab{a}}.

\bibitem[Behrouz et~al.(2026{\natexlab{b}})Behrouz, Zhong, and
  Mirrokni]{behrouz2025titans}
Behrouz, A., Zhong, P., and Mirrokni, V.
\newblock Titans: Learning to memorize at test time.
\newblock \emph{Advances in Neural Information Processing Systems},
  38:\penalty0 113506--113543, 2026{\natexlab{b}}.

\bibitem[Bischof \& Van~Loan(1987)Bischof and Van~Loan]{bischof1987wy}
Bischof, C. and Van~Loan, C.
\newblock The wy representation for products of householder matrices.
\newblock \emph{SIAM Journal on Scientific and Statistical Computing},
  8\penalty0 (1):\penalty0 s2--s13, 1987.

\bibitem[Bisk et~al.(2020)Bisk, Zellers, Gao, Choi, et~al.]{bisk2020piqa}
Bisk, Y., Zellers, R., Gao, J., Choi, Y., et~al.
\newblock Piqa: Reasoning about physical commonsense in natural language.
\newblock In \emph{Proceedings of the AAAI conference on artificial
  intelligence}, volume~34, pp.\  7432--7439, 2020.

\bibitem[Cabannes et~al.(2026)Cabannes, Mazar{\'e}, Szilvasy, Douze, Lomeli,
  Auzina, Carpentier, Synnaeve, and J{\'e}gou]{cabannes2026sdm}
Cabannes, L., Mazar{\'e}, P.-E., Szilvasy, G., Douze, M., Lomeli, M., Auzina,
  I.~A., Carpentier, J., Synnaeve, G., and J{\'e}gou, H.
\newblock Sparse delta memory: Scaling the state of linear rnns through
  sparsity.
\newblock \emph{arXiv preprint arXiv:2607.07386}, 2026.

\bibitem[Choromanski et~al.(2020)Choromanski, Likhosherstov, Dohan, Song, Gane,
  Sarlos, Hawkins, Davis, Mohiuddin, Kaiser, et~al.]{choromanski2021performers}
Choromanski, K., Likhosherstov, V., Dohan, D., Song, X., Gane, A., Sarlos, T.,
  Hawkins, P., Davis, J., Mohiuddin, A., Kaiser, L., et~al.
\newblock Rethinking attention with performers.
\newblock \emph{arXiv preprint arXiv:2009.14794}, 2020.

\bibitem[Clark et~al.(2019)Clark, Lee, Chang, Kwiatkowski, Collins, and
  Toutanova]{clark2019boolq}
Clark, C., Lee, K., Chang, M.-W., Kwiatkowski, T., Collins, M., and Toutanova,
  K.
\newblock Boolq: Exploring the surprising difficulty of natural yes/no
  questions.
\newblock In \emph{Proceedings of the 2019 Conference of the North American
  Chapter of the Association for Computational Linguistics: Human Language
  Technologies, Volume 1 (Long and Short Papers)}, pp.\  2924--2936, 2019.

\bibitem[Clark et~al.(2018)Clark, Cowhey, Etzioni, Khot, Sabharwal, Schoenick,
  and Tafjord]{clark2018arc}
Clark, P., Cowhey, I., Etzioni, O., Khot, T., Sabharwal, A., Schoenick, C., and
  Tafjord, O.
\newblock Think you have solved question answering? try arc, the ai2 reasoning
  challenge.
\newblock \emph{arXiv preprint arXiv:1803.05457}, 2018.

\bibitem[Cui(2026)]{cui2026hola}
Cui, W.
\newblock A hippocampus for linear attention: An exact memory for what the
  recurrent state forgets.
\newblock \emph{arXiv preprint arXiv:2607.02303}, 2026.

\bibitem[Dao \& Gu(2024)Dao and Gu]{dao2024mamba2}
Dao, T. and Gu, A.
\newblock Transformers are ssms: Generalized models and efficient algorithms
  through structured state space duality.
\newblock \emph{arXiv preprint arXiv:2405.21060}, 2024.

\bibitem[Dua et~al.(2019)Dua, Wang, Dasigi, Stanovsky, Singh, and
  Gardner]{dua2019drop}
Dua, D., Wang, Y., Dasigi, P., Stanovsky, G., Singh, S., and Gardner, M.
\newblock Drop: A reading comprehension benchmark requiring discrete reasoning
  over paragraphs.
\newblock In \emph{Proceedings of the 2019 Conference of the North American
  Chapter of the Association for Computational Linguistics: Human Language
  Technologies, Volume 1 (Long and Short Papers)}, pp.\  2368--2378, 2019.

\bibitem[Duchi \& Singer(2009)Duchi and Singer]{duchi2009forwardbackward}
Duchi, J. and Singer, Y.
\newblock Efficient online and batch learning using forward backward splitting.
\newblock \emph{The Journal of Machine Learning Research}, 10:\penalty0
  2899--2934, 2009.

\bibitem[Duchi et~al.(2010)Duchi, Shalev-Shwartz, Singer, and
  Tewari]{duchi2010composite}
Duchi, J.~C., Shalev-Shwartz, S., Singer, Y., and Tewari, A.
\newblock Composite objective mirror descent.
\newblock In \emph{Colt}, volume~10, pp.\  14--26, 2010.

\bibitem[Gao et~al.(2024)Gao, Tow, Abbasi, Biderman, Black, DiPofi, Foster,
  Golding, Hsu, Le~Noac'h, Li, McDonell, Muennighoff, Ociepa, Phang, Reynolds,
  Schoelkopf, Skowron, Sutawika, Tang, Thite, Wang, Wang, and
  Zou]{eval-harness}
Gao, L., Tow, J., Abbasi, B., Biderman, S., Black, S., DiPofi, A., Foster, C.,
  Golding, L., Hsu, J., Le~Noac'h, A., Li, H., McDonell, K., Muennighoff, N.,
  Ociepa, C., Phang, J., Reynolds, L., Schoelkopf, H., Skowron, A., Sutawika,
  L., Tang, E., Thite, A., Wang, B., Wang, K., and Zou, A.
\newblock The language model evaluation harness, 07 2024.
\newblock URL \url{https://zenodo.org/records/12608602}.

\bibitem[Gu \& Dao(2023)Gu and Dao]{gu2023mamba}
Gu, A. and Dao, T.
\newblock Mamba: Linear-time sequence modeling with selective state spaces.
\newblock \emph{arXiv preprint arXiv:2312.00752}, 2023.

\bibitem[Gu et~al.(2021)Gu, Goel, and R{\'e}]{gu2022s4}
Gu, A., Goel, K., and R{\'e}, C.
\newblock Efficiently modeling long sequences with structured state spaces.
\newblock \emph{arXiv preprint arXiv:2111.00396}, 2021.

\bibitem[Hao et~al.(2011)Hao, Cai, Pang, and Zhang]{hao2011swde}
Hao, Q., Cai, R., Pang, Y., and Zhang, L.
\newblock From one tree to a forest: a unified solution for structured web data
  extraction.
\newblock In \emph{Proceedings of the 34th international ACM SIGIR conference
  on Research and development in Information Retrieval}, pp.\  775--784, 2011.

\bibitem[Hatamizadeh et~al.(2026)Hatamizadeh, Choi, and
  Kautz]{hatamizadeh2026gdn2}
Hatamizadeh, A., Choi, Y., and Kautz, J.
\newblock Gated deltanet-2: Decoupling erase and write in linear attention.
\newblock \emph{arXiv preprint arXiv:2605.22791}, 2026.

\bibitem[Hsieh et~al.(2024)Hsieh, Sun, Kriman, Acharya, Rekesh, Jia, Zhang, and
  Ginsburg]{hsieh2024ruler}
Hsieh, C.-P., Sun, S., Kriman, S., Acharya, S., Rekesh, D., Jia, F., Zhang, Y.,
  and Ginsburg, B.
\newblock Ruler: What's the real context size of your long-context language
  models?
\newblock \emph{arXiv preprint arXiv:2404.06654}, 2024.

\bibitem[Hu et~al.(2026)Hu, Pan, Du, Lan, Tang, Wen, Liang, and
  Sun]{hu2025comba}
Hu, J., Pan, Y., Du, J., Lan, D., Tang, T., Wen, Q., Liang, Y., and Sun, W.
\newblock Improving bilinear rnn with closed-loop control.
\newblock \emph{Advances in Neural Information Processing Systems},
  38:\penalty0 107480--107506, 2026.

\bibitem[Hutchins et~al.(2022)Hutchins, Schlag, Wu, Dyer, and
  Neyshabur]{hutchins2022block}
Hutchins, D., Schlag, I., Wu, Y., Dyer, E., and Neyshabur, B.
\newblock Block-recurrent transformers.
\newblock \emph{Advances in neural information processing systems},
  35:\penalty0 33248--33261, 2022.

\bibitem[Joshi et~al.(2017)Joshi, Choi, Weld, and
  Zettlemoyer]{joshi2017triviaqa}
Joshi, M., Choi, E., Weld, D.~S., and Zettlemoyer, L.
\newblock Triviaqa: A large scale distantly supervised challenge dataset for
  reading comprehension.
\newblock In \emph{Proceedings of the 55th Annual Meeting of the Association
  for Computational Linguistics (Volume 1: Long Papers)}, pp.\  1601--1611,
  2017.

\bibitem[Katharopoulos et~al.(2020)Katharopoulos, Vyas, Pappas, and
  Fleuret]{katharopoulos2020transformers}
Katharopoulos, A., Vyas, A., Pappas, N., and Fleuret, F.
\newblock Transformers are rnns: Fast autoregressive transformers with linear
  attention.
\newblock In \emph{International conference on machine learning}, pp.\
  5156--5165. PMLR, 2020.

\bibitem[Kohonen(1972)]{kohonen1972correlation}
Kohonen, T.
\newblock Correlation matrix memories.
\newblock \emph{IEEE transactions on computers}, 100\penalty0 (4):\penalty0
  353--359, 1972.

\bibitem[Kwiatkowski et~al.(2019)Kwiatkowski, Palomaki, Redfield, Collins,
  Parikh, Alberti, Epstein, Polosukhin, Devlin, Lee,
  et~al.]{kwiatkowski2019naturalquestions}
Kwiatkowski, T., Palomaki, J., Redfield, O., Collins, M., Parikh, A., Alberti,
  C., Epstein, D., Polosukhin, I., Devlin, J., Lee, K., et~al.
\newblock Natural questions: a benchmark for question answering research.
\newblock \emph{Transactions of the Association for Computational Linguistics},
  7:\penalty0 453--466, 2019.

\bibitem[Lahoti et~al.(2026)Lahoti, Li, Chen, Wang, Bick, Kolter, Dao, and
  Gu]{lahoti2026mamba3}
Lahoti, A.~S., Li, K., Chen, B., Wang, C., Bick, A., Kolter, Z., Dao, T., and
  Gu, A.
\newblock Mamba-3: Improved sequence modeling using state space principles.
\newblock In \emph{International Conference on Learning Representations},
  volume 2026, pp.\  86173--86202, 2026.

\bibitem[Lin et~al.(2025)Lin, Nikishin, He, and Courville]{lin2025forgetting}
Lin, Z., Nikishin, E., He, X., and Courville, A.
\newblock Forgetting transformer: Softmax attention with a forget gate.
\newblock In \emph{International Conference on Learning Representations},
  volume 2025, pp.\  69704--69738, 2025.

\bibitem[Liu et~al.(2025)Liu, Wang, Wu, Feng, Stone, and Liu]{liu2024longhorn}
Liu, B., Wang, R., Wu, L., Feng, Y., Stone, P., and Liu, Q.
\newblock Longhorn: State space models are amortized online learners.
\newblock In \emph{International Conference on Learning Representations},
  volume 2025, pp.\  95419--95434, 2025.

\bibitem[Lockard et~al.(2019)Lockard, Shiralkar, and
  Dong]{lockard2019openceres}
Lockard, C., Shiralkar, P., and Dong, X.~L.
\newblock Openceres: When open information extraction meets the semi-structured
  web.
\newblock In \emph{Proceedings of the 2019 Conference of the North American
  Chapter of the Association for Computational Linguistics: Human Language
  Technologies, Volume 1 (Long and Short Papers)}, pp.\  3047--3056, 2019.

\bibitem[Lozhkov et~al.(2024)Lozhkov, Ben~Allal, von Werra, and
  Wolf]{lozhkov2024finewebedu}
Lozhkov, A., Ben~Allal, L., von Werra, L., and Wolf, T.
\newblock Fineweb-edu: the finest collection of educational content, 2024.
\newblock URL \url{https://huggingface.co/datasets/HuggingFaceFW/fineweb-edu}.

\bibitem[Merity et~al.(2016)Merity, Xiong, Bradbury, and
  Socher]{merity2017pointer}
Merity, S., Xiong, C., Bradbury, J., and Socher, R.
\newblock Pointer sentinel mixture models.
\newblock \emph{arXiv preprint arXiv:1609.07843}, 2016.

\bibitem[Mihaylov et~al.(2018)Mihaylov, Clark, Khot, and
  Sabharwal]{mihaylov2018openbookqa}
Mihaylov, T., Clark, P., Khot, T., and Sabharwal, A.
\newblock Can a suit of armor conduct electricity? a new dataset for open book
  question answering.
\newblock In \emph{Proceedings of the 2018 conference on empirical methods in
  natural language processing}, pp.\  2381--2391, 2018.

\bibitem[Paperno et~al.(2016)Paperno, Kruszewski, Lazaridou, Pham, Bernardi,
  Pezzelle, Baroni, Boleda, and Fern{\'a}ndez]{paperno2016lambada}
Paperno, D., Kruszewski, G., Lazaridou, A., Pham, N.-Q., Bernardi, R.,
  Pezzelle, S., Baroni, M., Boleda, G., and Fern{\'a}ndez, R.
\newblock The lambada dataset: Word prediction requiring a broad discourse
  context.
\newblock In \emph{Proceedings of the 54th annual meeting of the association
  for computational linguistics (volume 1: Long papers)}, pp.\  1525--1534,
  2016.

\bibitem[Parikh \& Boyd(2014)Parikh and Boyd]{parikh2014proximal}
Parikh, N. and Boyd, S.
\newblock Proximal algorithms.
\newblock \emph{Foundations and Trends in optimization}, 1\penalty0
  (3):\penalty0 127--239, 2014.

\bibitem[Peng et~al.(2025)Peng, Zhang, Goldstein, Alcaide, Du, Hou, Lin, Liu,
  Lu, Merrill, et~al.]{peng2025rwkv7}
Peng, B., Zhang, R., Goldstein, D., Alcaide, E., Du, X., Hou, H., Lin, J., Liu,
  J., Lu, J., Merrill, W., et~al.
\newblock Rwkv-7" goose" with expressive dynamic state evolution.
\newblock \emph{arXiv preprint arXiv:2503.14456}, 2025.

\bibitem[Peng et~al.(2021)Peng, Pappas, Yogatama, Schwartz, Smith, and
  Kong]{peng2021rfa}
Peng, H., Pappas, N., Yogatama, D., Schwartz, R., Smith, N.~A., and Kong, L.
\newblock Random feature attention.
\newblock \emph{arXiv preprint arXiv:2103.02143}, 2021.

\bibitem[Qin et~al.(2024{\natexlab{a}})Qin, Sun, Li, Shen, Sun, and
  Zhong]{qin2024lightning}
Qin, Z., Sun, W., Li, D., Shen, X., Sun, W., and Zhong, Y.
\newblock Lightning attention-2: A free lunch for handling unlimited sequence
  lengths in large language models.
\newblock \emph{arXiv preprint arXiv:2401.04658}, 2024{\natexlab{a}}.

\bibitem[Qin et~al.(2024{\natexlab{b}})Qin, Yang, Sun, Shen, Li, Sun, and
  Zhong]{qin2024hgrn2}
Qin, Z., Yang, S., Sun, W., Shen, X., Li, D., Sun, W., and Zhong, Y.
\newblock Hgrn2: Gated linear rnns with state expansion.
\newblock \emph{arXiv preprint arXiv:2404.07904}, 2024{\natexlab{b}}.

\bibitem[Rae et~al.(2019)Rae, Potapenko, Jayakumar, and
  Lillicrap]{rae2019compressive}
Rae, J.~W., Potapenko, A., Jayakumar, S.~M., and Lillicrap, T.~P.
\newblock Compressive transformers for long-range sequence modelling.
\newblock \emph{arXiv preprint arXiv:1911.05507}, 2019.

\bibitem[Rajpurkar et~al.(2018)Rajpurkar, Jia, and Liang]{rajpurkar2018squad}
Rajpurkar, P., Jia, R., and Liang, P.
\newblock Know what you don't know: Unanswerable questions for squad.
\newblock In \emph{Proceedings of the 56th Annual Meeting of the Association
  for Computational Linguistics (Volume 2: Short Papers)}, pp.\  784--789,
  2018.

\bibitem[Sakaguchi et~al.(2021)Sakaguchi, Bras, Bhagavatula, and
  Choi]{sakaguchi2020winogrande}
Sakaguchi, K., Bras, R.~L., Bhagavatula, C., and Choi, Y.
\newblock Winogrande: An adversarial winograd schema challenge at scale.
\newblock \emph{Communications of the ACM}, 64\penalty0 (9):\penalty0 99--106,
  2021.

\bibitem[Sap et~al.(2019)Sap, Rashkin, Chen, Le~Bras, and
  Choi]{sap2019socialiqa}
Sap, M., Rashkin, H., Chen, D., Le~Bras, R., and Choi, Y.
\newblock Social iqa: Commonsense reasoning about social interactions.
\newblock In \emph{Proceedings of the 2019 conference on empirical methods in
  natural language processing and the 9th international joint conference on
  natural language processing (EMNLP-IJCNLP)}, pp.\  4463--4473, 2019.

\bibitem[Schlag et~al.(2021)Schlag, Irie, and Schmidhuber]{schlag2021linear}
Schlag, I., Irie, K., and Schmidhuber, J.
\newblock Linear transformers are secretly fast weight programmers.
\newblock In \emph{International conference on machine learning}, pp.\
  9355--9366. PMLR, 2021.

\bibitem[Schmidhuber(1992)]{schmidhuber1992fastweights}
Schmidhuber, J.
\newblock Learning to control fast-weight memories: An alternative to dynamic
  recurrent networks.
\newblock \emph{Neural Computation}, 4\penalty0 (1):\penalty0 131--139, 1992.

\bibitem[Siems et~al.(2026{\natexlab{a}})Siems, Carstensen, Zela, Hutter,
  Pontil, and Grazzi]{siems2025deltaproduct}
Siems, J., Carstensen, T., Zela, A., Hutter, F., Pontil, M., and Grazzi, R.
\newblock Deltaproduct: Improving state-tracking in linear rnns via householder
  products.
\newblock \emph{Advances in Neural Information Processing Systems},
  38:\penalty0 153738--153782, 2026{\natexlab{a}}.

\bibitem[Siems et~al.(2026{\natexlab{b}})Siems, Grazzi, P{\"o}ppel, Singh,
  Zela, Carstensen, Jitsev, Hutter, Cevher, Orvieto,
  et~al.]{siems2026complexkda}
Siems, J., Grazzi, R., P{\"o}ppel, K., Singh, J., Zela, A., Carstensen, T.,
  Jitsev, J., Hutter, F., Cevher, V., Orvieto, A., et~al.
\newblock Complex kda: Understanding and enhancing the expressivity of kimi
  delta attention.
\newblock \emph{arXiv preprint arXiv:2609.24797}, 2026{\natexlab{b}}.

\bibitem[Sun et~al.(2023)Sun, Dong, Huang, Ma, Xia, Xue, Wang, and
  Wei]{sun2023retnet}
Sun, Y., Dong, L., Huang, S., Ma, S., Xia, Y., Xue, J., Wang, J., and Wei, F.
\newblock Retentive network: A successor to transformer for large language
  models.
\newblock \emph{arXiv preprint arXiv:2307.08621}, 2023.

\bibitem[Sun et~al.(2024)Sun, Li, Dalal, Xu, Vikram, Zhang, Dubois, Chen, Wang,
  Koyejo, et~al.]{sun2024ttt}
Sun, Y., Li, X., Dalal, K., Xu, J., Vikram, A., Zhang, G., Dubois, Y., Chen,
  X., Wang, X., Koyejo, S., et~al.
\newblock Learning to (learn at test time): Rnns with expressive hidden states.
\newblock \emph{arXiv preprint arXiv:2407.04620}, 2024.

\bibitem[Team et~al.(2025)Team, Zhang, Lin, Yao, Hu, Meng, Liu, Men, Yang, Li,
  et~al.]{kimi2025linear}
Team, K., Zhang, Y., Lin, Z., Yao, X., Hu, J., Meng, F., Liu, C., Men, X.,
  Yang, S., Li, Z., et~al.
\newblock Kimi linear: An expressive, efficient attention architecture.
\newblock \emph{arXiv preprint arXiv:2510.26692}, 2025.

\bibitem[Tillet et~al.(2019)Tillet, Kung, and Cox]{tillet2019triton}
Tillet, P., Kung, H.-T., and Cox, D.
\newblock Triton: an intermediate language and compiler for tiled neural
  network computations.
\newblock In \emph{Proceedings of the 3rd ACM SIGPLAN International Workshop on
  Machine Learning and Programming Languages}, pp.\  10--19, 2019.

\bibitem[Tumma et~al.(2026)Tumma, Loo, and Rus]{tumma2026preconditioned}
Tumma, N., Loo, N., and Rus, D.
\newblock Preconditioned deltanet: Curvature-aware sequence modeling for linear
  recurrences.
\newblock \emph{arXiv preprint arXiv:2604.21100}, 2026.

\bibitem[von Oswald et~al.(2026)von Oswald, Scherrer, Kobayashi, Versari, Yang,
  Schlegel, Maile, Schimpf, Sieberling, Meulemans,
  et~al.]{vonoswald2026mesanet}
von Oswald, J., Scherrer, N., Kobayashi, S., Versari, L., Yang, S., Schlegel,
  M., Maile, K., Schimpf, Y., Sieberling, O., Meulemans, A., et~al.
\newblock Mesanet: Sequence modeling by locally optimal test-time training.
\newblock In \emph{International Conference on Learning Representations},
  volume 2026, pp.\  26397--26447, 2026.

\bibitem[Wang et~al.(2025)Wang, Shi, and Fox]{wang2026ttr}
Wang, K.~A., Shi, J., and Fox, E.~B.
\newblock Test-time regression: a unifying framework for designing sequence
  models with associative memory.
\newblock \emph{arXiv preprint arXiv:2501.12352}, 2025.
\newblock URL \url{https://arxiv.org/abs/2501.12352}.

\bibitem[Wen et~al.(2025)Wen, Dang, and Lyu]{wen2025rnns}
Wen, K., Dang, X., and Lyu, K.
\newblock Rnns are not transformers (yet): The key bottleneck on in-context
  retrieval.
\newblock In \emph{International Conference on Learning Representations},
  volume 2025, pp.\  48813--48856, 2025.

\bibitem[Widrow \& Hoff(1960)Widrow and Hoff]{widrow1960adaptive}
Widrow, B. and Hoff, Jr., M.~E.
\newblock Adaptive switching circuits.
\newblock In \emph{IRE WESCON Convention Record}, volume~4, pp.\  96--104,
  1960.

\bibitem[Wiener(1949)]{wiener1949extrapolation}
Wiener, N.
\newblock \emph{Extrapolation, interpolation, and smoothing of stationary time
  series: with engineering applications}.
\newblock The MIT press, 1949.

\bibitem[Yang \& Zhang(2024)Yang and Zhang]{yang2024fla}
Yang, S. and Zhang, Y.
\newblock Fla: A triton-based library for hardware-efficient implementations of
  linear attention mechanism, January 2024.
\newblock URL \url{https://github.com/fla-org/flash-linear-attention}.

\bibitem[Yang et~al.(2023)Yang, Wang, Shen, Panda, and Kim]{yang2024gla}
Yang, S., Wang, B., Shen, Y., Panda, R., and Kim, Y.
\newblock Gated linear attention transformers with hardware-efficient training.
\newblock \emph{arXiv preprint arXiv:2312.06635}, 2023.

\bibitem[Yang et~al.(2024)Yang, Wang, Zhang, Shen, and Kim]{yang2024deltanet}
Yang, S., Wang, B., Zhang, Y., Shen, Y., and Kim, Y.
\newblock Parallelizing linear transformers with the delta rule over sequence
  length.
\newblock \emph{Advances in neural information processing systems},
  37:\penalty0 115491--115522, 2024.

\bibitem[Yang et~al.(2025)Yang, Kautz, and Hatamizadeh]{yang2024gdn}
Yang, S., Kautz, J., and Hatamizadeh, A.
\newblock Gated delta networks: Improving mamba2 with delta rule.
\newblock In \emph{International Conference on Learning Representations},
  volume 2025, pp.\  29687--29707, 2025.

\bibitem[Zellers et~al.(2019)Zellers, Holtzman, Bisk, Farhadi, and
  Choi]{zellers2019hellaswag}
Zellers, R., Holtzman, A., Bisk, Y., Farhadi, A., and Choi, Y.
\newblock Hellaswag: Can a machine really finish your sentence?
\newblock In \emph{Proceedings of the 57th annual meeting of the association
  for computational linguistics}, pp.\  4791--4800, 2019.

\bibitem[Zhang et~al.(2024)Zhang, Yang, Zhu, Zhang, Cui, Wang, Wang, Shi, Wang,
  Bi, et~al.]{zhang2024gsa}
Zhang, Y., Yang, S., Zhu, R., Zhang, Y., Cui, L., Wang, Y., Wang, B., Shi, F.,
  Wang, B., Bi, W., et~al.
\newblock Gated slot attention for efficient linear-time sequence modeling.
\newblock \emph{arXiv preprint arXiv:2409.07146}, 2024.

\end{thebibliography}
